\documentclass[times,twocolumn,final]{elsarticle}

\usepackage{medima}
\usepackage{framed,multirow}

\usepackage{amssymb}
\usepackage{latexsym}

\usepackage{url}
\usepackage[dvipsnames]{xcolor}
\usepackage{hyperref}
\usepackage{amsmath,amssymb,amsfonts}
\usepackage{breqn}
\usepackage{algorithm}
\usepackage{algpseudocode}
\usepackage{graphicx}
\usepackage{textcomp}
\usepackage{thmtools, thm-restate}
\usepackage{epsfig}
\usepackage{booktabs}
\usepackage{verbatim}
\usepackage{xcolor}
\usepackage{amstext}
\usepackage{upgreek}
\usepackage{multirow}
\usepackage{bbm}
\usepackage{rotating}
\usepackage{pifont}
\usepackage{colortbl}
\usepackage{arydshln}

\newcommand{\real}{\mathbb{R}}

\usepackage[switch]{lineno}
\usepackage{subfig}
\usepackage{amsfonts}

\definecolor{mypink1}{rgb}{0.858, 0.188, 0.478}
\definecolor{myorange}{RGB}{255,165,0}
\definecolor{gray}{cmyk}{0.86,0.86,0.86,0.86}
\usepackage{color, colortbl}
\definecolor{Gray}{gray}{0.9}

\usepackage{threeparttable, tablefootnote}

\definecolor{newcolor}{rgb}{.8,.349,.1}
\def\rmX{{\mathbf{X}}}

\def\rmt{{\mathbf{T}}}
\def\real{{\mathbb{R}}}

\journal{Medical Image Analysis}

\begin{document}

\verso{Silva-Rodr\'iguez \textit{et~al.}}

\begin{frontmatter}

\title{A Foundation Language-Image Model of the Retina (FLAIR): Encoding Expert Knowledge in Text Supervision}

\author[1]{Julio Silva-Rodr\'iguez$^\ast$}
\cortext[cor1]{Corresponding author: julio-jose.silva-rodriguez@etsmtl.ca}
\author[2]{Hadi Chakor}
\author[2]{Riadh Kobbi}
\author[1,3]{Jose Dolz }
\author[1,3]{Ismail Ben Ayed}

\address[1]{ÉTS Montréal, Québec, Canada}
\address[2]{DIAGNOS Inc., Québec, Canada}
\address[3]{Centre de Recherche du Centre Hospitalier de l’Universit\'e de Montr\'eal (CR-CHUM), Québec, Canada}

\begin{abstract}

    Foundation vision-language models are currently transforming computer vision, and are on the rise in medical imaging fueled by their very promising generalization capabilities. However, the initial attempts to transfer this new paradigm to medical imaging have shown less impressive performances than those observed in other domains, due to the significant domain shift and the complex, expert domain knowledge inherent to medical-imaging tasks. Motivated by the need for domain-expert foundation models, we present FLAIR, a pre-trained vision-language model for universal retinal fundus image understanding. To this end, we compiled $38$ open-access, mostly categorical fundus imaging datasets from various sources, with up to $101$ different target conditions and $288,307$ images. We integrate the expert's domain knowledge in the form of descriptive textual prompts, during both pre-training and zero-shot inference, enhancing the less-informative categorical supervision of the data. Such a textual expert's knowledge, which we compiled from the relevant clinical literature and community standards, describes the fine-grained features of the pathologies as well as the hierarchies and dependencies between them. We report comprehensive evaluations, which illustrate the benefit of integrating expert knowledge and the strong generalization capabilities of FLAIR under difficult scenarios with domain shifts or unseen categories. When adapted with a lightweight linear probe, FLAIR outperforms fully-trained, dataset-focused models, more so in the few-shot regimes. Interestingly, FLAIR outperforms by a wide margin larger-scale generalist image-language models and retina domain-specific self-supervised networks, which emphasizes the potential of embedding experts' domain knowledge and the limitations of generalist models in medical imaging. The pre-trained model is available at: \url{https://github.com/jusiro/FLAIR}. \\


\end{abstract}

\end{frontmatter}



\section{Introduction}
\label{sec:intro}

At least 1 billion people have a vision impairment that could have been prevented or is yet to be addressed \citep{who}. In this context, color fundus images combined with computer vision systems present a promising, cost-effective solution for population-based screenings and early detection of ophthalmologic diseases \citep{Balyen2019, Bellemo2019}. 

Driven by public datasets, deep learning has reached remarkable performances in a breadth of fundus image analysis problems, such as diabetic retinopathy grading \citep{Liu2022}, glaucoma detection \citep{Orlando2019}, lesion segmentation \citep{Porwal2020} or multi-disease detection \citep{1000x39}. Nevertheless, several limitations impede the widespread adoption of these methods. In particular, current deep learning solutions for fundus image analysis may not generalize well whenever there are shifts in the imaging data or in the task at hand (e.g., new or rare classes) \citep{Li2021, Sengupta2020}. In retinal imaging, and in the much broader field of medical imaging, the current dominant deep learning paradigm is to supervise models on very specific tasks, e.g., diabetic retinopathy classification into a few grades \citep{Liu2022}. Learning representations that might be too specialized for the task and training images at hand, such task-focused models may have difficulty in (i) dealing with the high variability in real clinical scenarios \citep{Finlayson2021}, due to the high variations in image acquisition and patient demographics; and (ii) capturing rare conditions that are not well represented in the training data.

There is currently a paradigm shift in artificial intelligence algorithms, driven by the growing prevalence of models trained on large and diverse datasets, which could be adapted to a broad span of downstream tasks. These models, commonly referred to as \textit{foundation} models, have gained increasing popularity and showed significant success in computer vision and natural language processing tasks \citep{brown2020language,Radford2021}. In particular, vision-language models such as CLIP \citep{Radford2021} or ALIGN \citep{Jia2021} have shown impressive generalization capabilities when fine-tuned on various downstream computer-vision tasks, emerging as powerful alternatives to narrowly-supervised, task-focused models. Learning from large-scale amounts of image-text pairs, such models leverage the rich semantic knowledge in the language-based supervision, thereby yielding visual features that are more descriptive than their task-specific counterparts. 

\begin{figure}[h!]
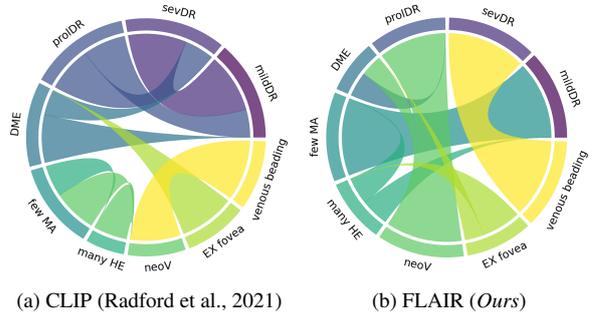

    \begin{center}

          \subfloat[CLIP \citep{Radford2021}\label{fig:chord_a}]{\includegraphics[width=0.49\linewidth]{images/chord_plot_clip.pdf}}
          \hspace*{\fill}
          \subfloat[FLAIR (\textit{Ours})\label{fig:chord_b}]{\includegraphics[width=0.49\linewidth]{images/chord_plot.pdf}}
          \hspace*{\fill}

        \caption{\textbf{CLIP limitations on medical domains.} The figure depicts the cosine similarities of the text embeddings for common retinal diseases and lesions observed on fundus images. While CLIP mostly focuses on general medical relations (e.g., “\textit{diabetic}”, or “\textit{neovascularization}”- “\textit{venous}”), the proposed domain-specific model (\textit{i.e.}, FLAIR) is able to capture the hierarchical dependencies between concepts (e.g., the fundus images of “\textit{mildDR}” contain “\textit{only a few microaneurysms}”, and “\textit{neovascularization}” is the differential sign for “\textit{prolDR}” diagnosis).}
        \label{fig:chords}
    \end{center}
\end{figure}

In computer vision tasks, such as image classification, this new {\em pretrain-and-finetune} paradigm enhanced robustness to image-data shifts and showed promising zero-shot and few-shot transferability. Nonetheless, initial attempts to directly apply these foundation models to the medical domain yielded less convincing performances \citep{Wang2022}. Indeed, generalist models like CLIP may not capture the fine-grained image features and class dependencies/hierarchies, which might be complex, highly specialized concepts inherent to the expert's domain knowledge; see Figure \ref{fig:chords} for an illustration in the case of retinal fundus images. This has recently motivated the development of foundation models specialized for medical imaging applications \citep{Malwina2022, Moor2023}.

Vision-language models are currently emerging in medical image analysis. Several recent studies investigated foundation models specialized in radiology \citep{Zhang2020,GLoRIA,Wang2022}, focusing mostly on chest-radiography data. These were motivated by the prevalence of diagnostic text reports in radiology, and the availability of large domain resources to mine such textual information \citep{Bodenreider2004,Jain2021}. However, this may not be the case in other modalities. In retinal imaging, for instance, text information is scarce and most datasets are categorically labeled (see Table \ref{datasets_assembly}), \textit{i.e.}, the label of each training image is a single category (or class), e.g., “\textit{mild diabetic retinopathy}” (mildDR). 

\begin{figure}[ht!]
\begin{center}
\includegraphics[width=.48\textwidth]{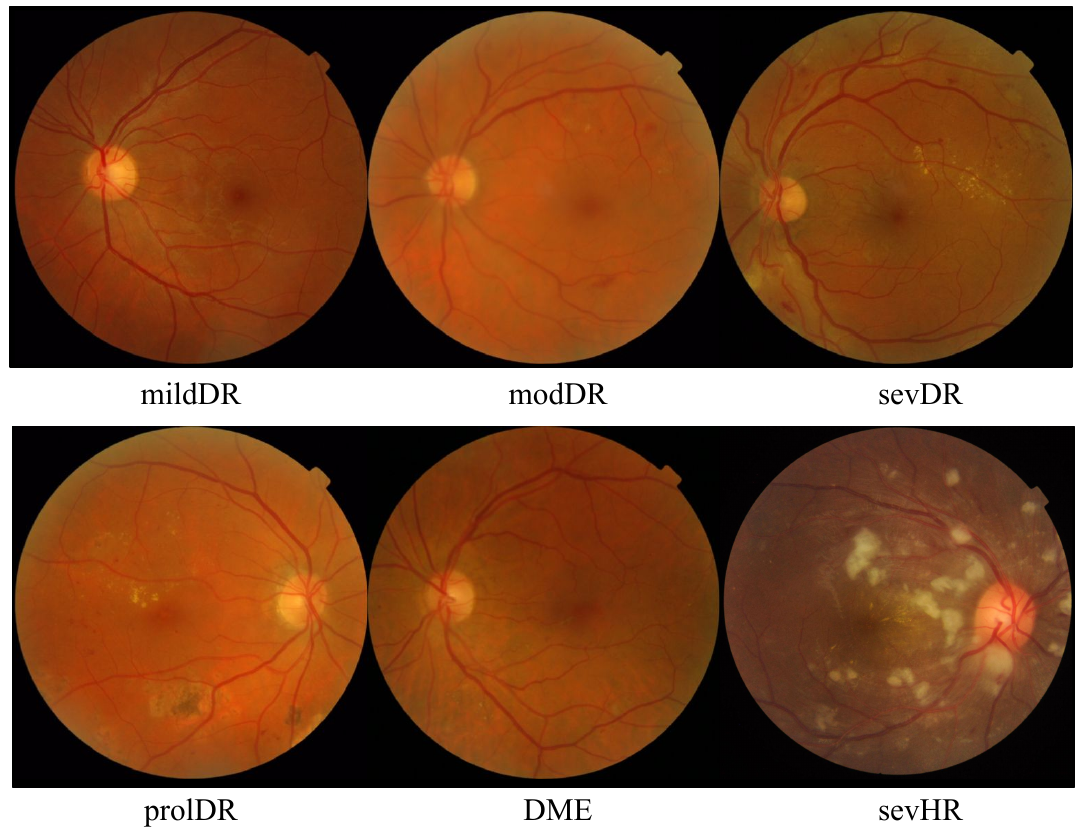}
\caption{\textbf{Expert knowledge descriptors.} The analysis of fundus images by ophthalmologists is driven by hierarchical features. According to the American Academy of Ophthalmology \citep{Wilkinson2003}, mildDR is characterized by “\textit{only few microaneurysms present}”, modDR includes “\textit{retinal haemorrhages in few quadrants}”, “\textit{many haemorrhages}” or “\textit{cotton wool spots}”, and sevDR and prolDR are distinguished by “\textit{venous beading}”/“\textit{intraretinal microvascular abnormalities}” and “\textit{neovascularization}”, respectively. DME is also usually featured by “\textit{hard exudates involving the center of the macula}”. Furthermore, according to \citep{sevHR}, hypertensive retinopathy is generally described as “\textit{flame-shaped hemorrhages in the superficial layers of the retina and cotton-wool patches}”. Going deeper into the hierarchies between concepts, exudates are “\textit{small white or yellowish deposits}”, and microaneurysms are “\textit{small red dots}”.}
\label{fig:diseases_images}
\end{center}
\end{figure}

We argue that, even for categorically-labeled images, vision-language pre-training is an appealing solution to integrate domain-specific, fine-grained knowledge, such as the dependencies between the categories, into visual representations. The analysis of medical images by clinical experts is a process of searching for differential features of candidate conditions. In this process, there are, for instance, hierarchical dependencies between the presence of local lesions and the differential diagnosis at the global level. Such expert's domain knowledge is usually overlooked in conventional training but could be integrated in the form of text descriptions, to build powerful image-language models. To illustrate this, we provide in Figure \ref{fig:diseases_images} a few retinal-imaging examples with categorical labels along with the corresponding text descriptions encoding domain knowledge. For instance, the text description “\textit{only a few microaneurysms are present}” informs on local conditions known to point to the category mildDR \citep{Wilkinson2003}. In Table \ref{dk_description}, we provide a comprehensive list of the correspondences between the categorical labels and textual domain-knowledge descriptions, which we compiled from the relevant clinical literature \citep{sevHR} and from community standards \citep{Wilkinson2003}, to build our foundation model of the retina. 

In this work, we introduce FLAIR, a Foundation LAnguage-Image model of the Retina, for color fundus image analysis. FLAIR is trained and validated on a large assembly of $38$ datasets, with $288,307$ images and $101$ different target categories, which we compiled from different publicly available sources. We integrate the expert's domain knowledge in the form of text supervision during both pre-training and zero-shot prediction, thereby enhancing the categorical information of the data. Such a textual expert's knowledge describes fine-grained features of the pathologies as well as the hierarchies and dependencies between them. We report comprehensive evaluations, comparisons and ablation studies, which show the substantial effect of embedding expert knowledge and the strong generalization and transferability capabilities of FLAIR under challenging scenarios with domain shifts or novel (unseen) categories. When adapted with a lightweight linear probe classifier, FLAIR outperforms models that are fully trained on the target dataset, more so under low-data (few-shot) settings. Furthermore, FLAIR outperforms by a large margin more generalist, larger-scale image-language models such as CLIP or BiomedCLIP. Our results point to the potential of embedding expert domain knowledge and to the limitations of generalist models.


\section{Related Work}
\label{sec:rw}

\subsection{Transfer learning in medical image analysis}

Training robust deep-learning models from scratch requires large datasets and huge computational resources \citep{Erhan2009}. These conditions are rarely met in medical imaging. The high variability in image acquisition, the low prevalence of certain conditions, and the limited resources of institutions make it difficult for standard supervised-learning models to capture the substantial variability in real clinical contexts. This is due to the fact that supervised models are typically trained on and specialized for relatively small data sets and tasks, due to the prohibitive costs and resources of labeling the data. To mitigate this to some extent, transfer learning from natural images, whereby a deep model is pre-trained on a large labeled dataset such as ImageNet and then used as initialization for adaptation to a target task, has become the \textit{de-facto} solution \citep{raghu2019transfusion, Kanavati2021, Matsoukas2022}. In particular, fine-tuning the whole network, or just the last layers, demonstrated promising performances in a breadth of medical-imaging tasks, across various domains such as radiology, cardiology, and ophthalmology \citep{Tajbakhsh2017, Abramoff2016, Fauw2018}. Nonetheless, several in-depth empirical studies have exposed the limited performance gains of such transfer-learning solutions in certain scenarios in medical-image classification \citep{raghu2019transfusion, Neyshabur2020, Matsoukas2022}. Larger-scale pre-training, which leverages unlabeled data via self-supervision \citep{Chen2022, Huang2023}, is a promising alternative. However, supervised, domain-specific pre-training remains the prevalent solution for optimal transfer learning \citep{Zhang2020,Liu2023}. Under this standard supervised-learning paradigm, top-competing solutions on the DeepDRiD challenge for diabetic retinopathy grading on fundus images \citep{Liu2022} performed an exhaustive, task-specific pre-training using public datasets. 

\subsection{From supervised, task-specific models to large-scale vision-language pre-training}

As discussed above, supervised, task-specific models are currently prevalent in medical imaging. Nevertheless, the generalization of such task-focused models across the various conditions encountered in clinical scenarios remains a major challenge for wider adoption \citep{Finlayson2021}. Task-specific models, e.g., diabetic retinopathy classification into a few grades, might yield features that are too specialized for the task and training data at hand. As we will show in the empirical validation of this work, such models generalize poorly whenever there are shifts in the task (e.g., new classes) or in the imaging data. Such shifts occur frequently in medical imaging due to the high variability in image acquisition and patient demographics, and/or to the low prevalence of certain conditions. 

Very recently, Vision-Language Pre-training (VLP) has made substantial progress in computer vision and machine learning, emerging as a powerful solution to improve the generalization of deep models. Large-scale VLP models leverage paired image-language data through image-text contrastive learning \citep{Radford2021, Jia2021, Yang2022}, yielding robust and generic feature extractors. Such pre-training models have shown impressive transfer-learning capabilities when fine-tuned on downstream tasks \citep{Radford2021}. On natural images, this widely emerging pre-train-and-finetune paradigm yielded excellent robustness to data shifts and strong generalization to new tasks, with no (\textit{i.e.}, zero-shot \citep{shu2022test,zhao2022exploiting}) or only a few (\textit{i.e.}, few-shot \citep{hu2022pushing}) labeled samples in the new task. For instance, CLIP \citep{Radford2021} provides prompt-based (zero-shot) classifications by capturing the similarity between the image and a textual description of the target class, via the jointly trained vision and language encoders. In addition, in few-shot regimes (\textit{i.e.}, when few labeled samples in the target task are available), the visual representations show strong transferability by updating a linear-classifier layer on top of the frozen network. Such a fast fine-tuning procedure is commonly referred to as Linear Probing (LP). These observations have motivated an increasing interest in efficient forms of adapting CLIP to downstream tasks and domains, using lightweight multi-modal modules, known as Adapters \citep{clipAdapter, tipAdapter}. Although these efficient transferability properties are of huge interest in medical imaging analysis, applying CLIP directly to medical imaging data yields sub-optimal results \citep{Wang2022}. Along with the domain shifts occurring in the imaging modality, this might be due, in part, to the complex, highly specialized terminology encountered in medical imaging. 

\subsection{Vision-language models in medical imaging}

The rise of VLP models is at its beginning in medical imaging. Several recent works investigated contrastive image-language models tailored to medical data \cite{Zhang2020,GLoRIA,Wang2022,Lu_2023_CVPR}, but mostly in the application area of chest radiographs. VLP models are particularly appealing in the field of radiology \citep{Zhang2020}, as the diagnostic text reports associated with the images are common in everyday radiology practices. Thus, public, large-scale, and multi-modal datasets with paired images and language descriptions started to appear recently, such as MIMIC-CXR \citep{Johnson2019}, PadChest \citep{Bustos2019} or ROCO \citep{Pelka2018}. In addition, there exist large domain resources, such as UMLS \citep{Bodenreider2004}, BioClinicalBERT or RadGraph \citep{Jain2021}, which favor the processing and knowledge extraction of structured clinical information from free-text radiology reports. 

Fueled by the existence of this domain knowledge, several recent works have developed strategies to overcome the limitations of CLIP in the medical field. For example, methods using CLIP for inference have integrated modality prompts \citep{Liu2023}, or attribute descriptions \citep{Menon2023} for prompt-based inference, using pre-trained question-answering models to describe the shape and color of the target conditions \citep{Qin2023}. Other works have focused on the pre-training stage, generating domain-specialized VLP models such as ConVirt \citep{Zhang2020}, PubMedCLIP \citep{PubMedCLIP}, GLorIA \citep{GLoRIA}, MedCLIP \citep{Wang2022} or MedKLIP \citep{Wu2023}, among others \citep{Windsor2023, Wang2022_Multi-Granularity, RueckertJoint, Chen2022GraphDK}. One of the main challenges of such pre-training lies in the low prevalence of text-based supervision on publicly available datasets. To alleviate this issue, MedCLIP incorporated categorically-labeled samples through label-space alignment \citep{Wang2022}. Other methods have taken profit from well-established domain tools in radiology such as UMLS and RadGraph to augment the available text reports \citep{Chen2022GraphDK,Wu2023}. 

Despite these recent advances in the development of vision-language pre-training strategies in medical imaging, the use of categorically-labeled datasets has been overlooked. In this work, we argue and show that such supervision could still be exploited to train powerful vision-language representations, by encoding expert's domain knowledge into text supervision.

\subsection{Expert knowledge-driven models of fundus images}

The idea of integrating domain knowledge into deep learning for medical image analysis is not new, and has triggered interest in the recent literature \citep{Xie2021}. In particular, domain-specific, expert knowledge (EK) from clinicians could be retrieved to highlight areas of interest, relevant features, anatomical priors, or inter-disease dependencies and hierarchies. In retinal imaging, the expert's knowledge has been integrated in various ways. For instance, \cite{Giancardo2012} first segmented the exudates, which served as a proxy for macular edema detection. Similarly, several other strategies train attention modules to enhance local lesions, which act as surrogates for disease classification. Closely related to our work, we have identified several categories, which include: using pixel-level annotated lesions for AMD staging \citep{Fang2019}, weakly-supervised strategies based on the relationships between diabetic retinopathy and diabetic macular edema \citep{Xiaomeng2020}, or disentangling disease-specific saliency maps for diabetic retinopathy grading \citep{Sun2021}. In addition, expert knowledge for glaucoma detection in fundus images is usually integrated by cropping the optic-disk area as an initial step before classification \citep{acrima,AIROGS}. Unlike this existing literature, we study the use of well-established expert knowledge on retinal image analysis via vision-language pre-training, which has been largely overlooked in the context of foundation models. Concretely, we propose a contrastive image-text pre-training, which incorporates relevant features, hierarchies, and relationships between the classes as well as information on the regions of interest characterizing the target diseases, in the form of descriptive textual prompts, paired with the corresponding images. 


\section{Methodology}
\label{sec:methods}

\begin{figure*}[ht!]
\begin{center}

\includegraphics[width=1.0\textwidth]{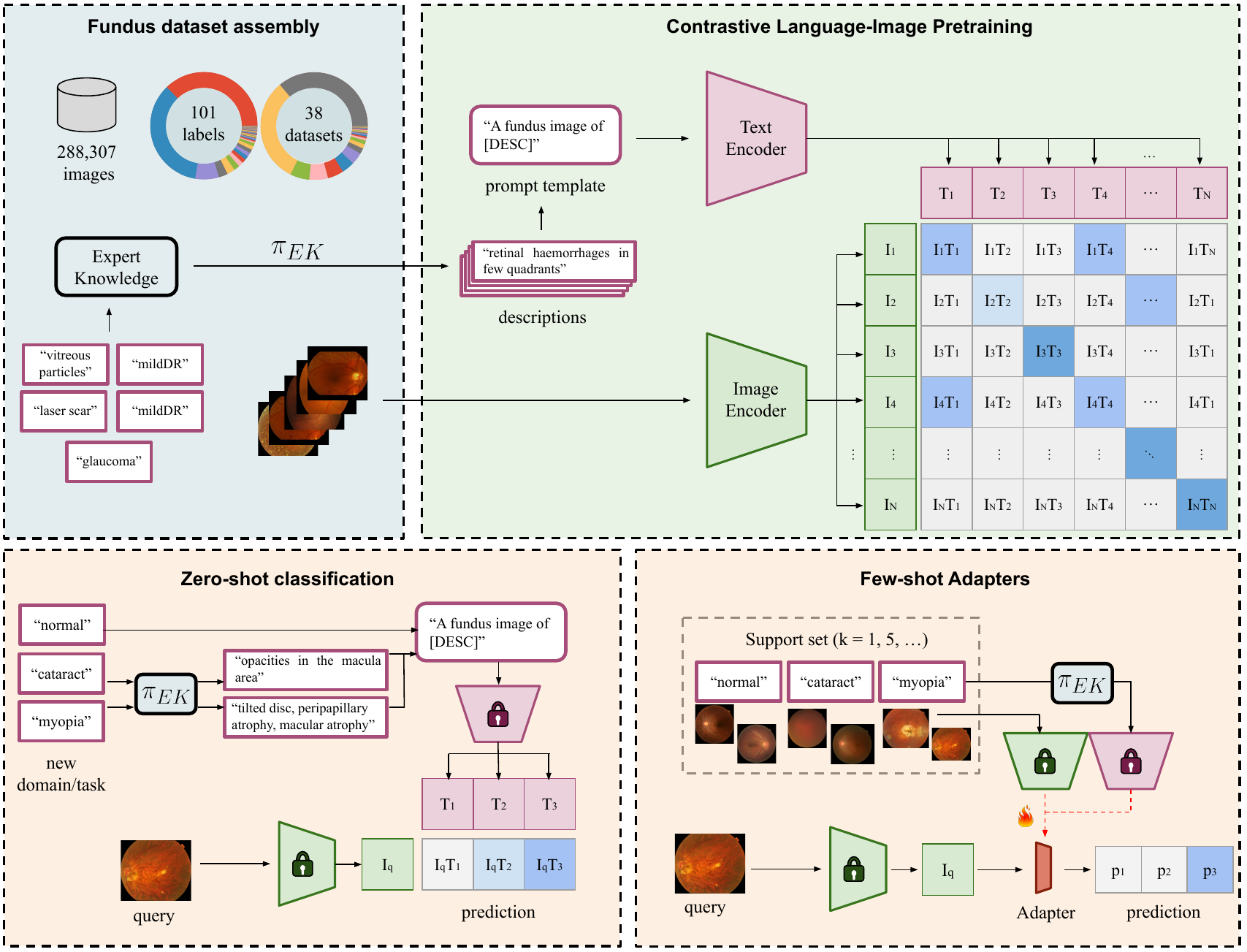}
\caption{\textbf{Framework overview}. We have developed a knowledge-based universal model of the retina from an assembly of 38 public datasets, which contains 288,307 color fundus images and 101 different categories (\textit{see top-left}). The foundation model consists of vision and language encoders, which are trained in a contrastive fashion on paired images and textual descriptors. To mitigate the scarcity of text-based supervision in publicly available retinal fundus imaging datasets, we propose to augment the categorical image labels by using well-established domain knowledge (\textit{see top-right}). The ensuing pre-training model enables to prediction of new categories in a zero-shot fashion, using well-designed descriptors based on domain knowledge and local features of the novel diseases; \textit{see bottom-left}. In addition, the model could adapt to downstream tasks and domains by tuning a lightweight Adapter on top of the image and vision encoders, by using only a few labeled samples (the support set); \textit{see bottom-right}.}
\label{fig:summary}
\end{center}
\end{figure*}

Fig. \ref{fig:summary} depicts an overview of our framework. We introduce each methodological component formally in the following.

\paragraph{\textbf{Problem setup}} Let us define an assembly dataset, $\mathcal{D}_{T}$, which contains $N$ samples gathered from different publicly available fundus image datasets, including heterogeneous sources and findings. For each sample, we build a multi-modal triplet including an image, a categorical label and a text description: $\mathcal{D}_{T}=\{(\rmX_n, y_n, \rmt_{n})\}_{n=1}^{N}$. $\rmX_n \in \real^{\Omega_n}$ denotes a fundus 2D image, with $\Omega_n$ its spatial domain, $y_n \in \{1, ... , C\}$ is a label 
among the $C$ unique categories in the assembly dataset, and $\rmt_{n}\in \mathcal{T}$ is a text description associated with the label. Figure \ref{fig:diseases_images} provides a few examples of categorical labels, such as DME, and the associated text descriptions encoding domain knowledge, e.g., “\textit{hard exudates involving the center of the macula}”. Such textual domain knowledge could be derived from the relevant clinical literature \citep{sevHR} and/or from community standards \citep{Wilkinson2003}. 
Table \ref{dk_description} provides a comprehensive list of the correspondences between the categorical labels and textual domain-knowledge descriptions, which we compiled from the relevant clinical literature, to build our foundation model of the retina. Note that a single categorical label may correspond to several text descriptions, each describing a different finding or feature in the image. The objective of our vision-language pre-training is to provide a powerful multi-modal model capable of learning a feature representation space where samples are aligned across the three modalities: images, categories, and text.

\subsection{Aligning images, labels and domain-knowledge text}
\label{ssec:vlp}

Our multi-modal pre-training integrates vision and language encoders. Let $\theta = \{\theta_f(\cdot), \theta_p(\cdot)\}$ denotes the vision encoder, with $\theta_f(\cdot)$ a feature extractor and $\theta_p(\cdot)$ a projection head. The feature extractor $\theta_f(\cdot)$ yields a feature representation $\tilde{{\mathbf u}} \in \real^{D_{{\mathbf u}}}: \tilde{{\mathbf u}}_i=\theta_f(\rmX_i)$ of an input image $\rmX_i$, with $D_{{\mathbf u}}$ the dimension of the visual feature space. Analogously, let $\phi = \{\phi_f(\cdot), \phi_p(\cdot)\}$ denotes the text encoder, $\phi_f(\cdot)$ being a feature extractor and $\phi_p(\cdot)$ a projection head. The feature extractor $\phi_f(\cdot)$ provides an embedding $\tilde{{\mathbf v}} \in \real^{D_{{\mathbf v}}}: \tilde{{\mathbf v}}_j=\phi_f(\rmt_j)$ of an input text $\rmt_j$, with $D_{{\mathbf v}}$ denoting the dimension of the space of text features. Each of the projection heads, $\theta_p(\cdot)$ and $\phi_p(\cdot)$, maps the independent modality representations into a joint unit hyper-sphere space: ${\mathbf u} = \frac{\theta_p(\tilde{{\mathbf u}})}{||\theta_p(\tilde{{\mathbf u}})||}$ and ${\mathbf v} = \frac{\phi_p(\tilde{{\mathbf v}})}{||\phi_p(\tilde{{\mathbf v}})||}$. In this normalized space, the similarity between image $\rmX_i$ and text description $\rmt_j$ is evaluated by the cosine similarity: 
${\mathbf u}^{T}_i {\mathbf v}_j$, where $T$ denotes the transpose operator.

The objective consists of learning feature representations that minimize the distances between paired image and text descriptions while maximizing the distances between unpaired samples. We build image-text pairs from the available categorical label information, thereby encouraging samples belonging to the same category to have close feature representations, in both the image and text domains. More formally, let $\mathcal{B}$ denote a batch containing a set of images $\{\rmX_i\}_{i \in \mathcal{X}_{B}}$ and a set of text descriptions 
$\{\rmt_j\}_{j \in \mathcal{T}_{B}}$, where $\mathcal{X}_{B} \subset \{1, \dots, N\}$ denotes the set of indices of the images in $\mathcal{B}$, and $\mathcal{T}_B \subset \{1, \dots, N\}$ the set of indices of the text descriptions in $\mathcal{B}$. We minimize category-aware image-to-text ($\mathcal{L}_{i2t}$) and text-to-image ($\mathcal{L}_{t2i}$) contrastive objectives, defined as follows:
\begin{align}
\label{eq:i2t}
\mathcal{L}_{i2t} (\theta,\phi,\tau | \mathcal{B}) = - \sum_{i \in \mathcal{X}_B} \frac{1}{|P_{\mathcal{T}_B}(i)|} \sum_{i' \in P_{\mathcal{T}_B}(i)} \text{log} \frac{\text{exp}({\mathbf u}_{i}^{T} {\mathbf v}_{i'} / \tau)}{\sum_{j\in\mathcal{T}_B}\text{exp}({\mathbf u}^{T}_i {\mathbf v}_j / \tau)}
\end{align}
 \begin{align}
\label{eq:t2i}
\mathcal{L}_{t2i} (\theta,\phi,\tau | \mathcal{B}) = - \sum_{j \in \mathcal{T}_B} \frac{1}{|P_{\mathcal{X}_B}(j)|} \sum_{j' \in P_{\mathcal{X}_B}(j)} \text{log} \frac{\text{exp}({\mathbf u}_{j'}^{T} {\mathbf v}_{j} / \tau)}{\sum_{i\in\mathcal{X}_B}\text{exp}({\mathbf u}^{T}_i {\mathbf v}_j / \tau)}
\end{align}
where $\tau \in \real_{++}$ is a trainable scaling parameter, $|\cdot|$ denotes the cardinality of a set and $P_{\mathcal{T}_B}(i)$ and $P_{\mathcal{X}_B}(j)$ contain indices of similar-category subsets of batch $\mathcal{B}$:
\[P_{\mathcal{T}_B}(i) = \{ i'|i' \in \mathcal{T}_B, y_{i'} = y_i \} \mbox{~~and~~} P_{\mathcal{X}_B}(j) = \{ j'|j' \in \mathcal{X}_B, y_{j'} = y_j \}\]
Thus, the vision and language encoders of the foundation fundus model are trained to optimize a bidirectional image and language alignment, using gradient descent and randomly batched samples: 
 \begin{align}
\label{eq:loss}
\min_{\theta,\phi,\tau} \quad & \sum_{\mathcal{B}} \mathcal{L}_{i2t} (\theta,\phi,\tau | \mathcal{B}) + \mathcal{L}_{t2i} (\theta,\phi,\tau | \mathcal{B})
\end{align}

\subsection{Expert knowledge as additional text supervision during pre-training.}
\label{ssec:dk}

Vision-language pre-training for medical imaging has been mostly investigated in the context of radiology images \citep{Zhang2020}, where diagnostic text reports are common. Thus, large multi-modal datasets with paired radiology images and text descriptions are available \citep{Johnson2019}. Nevertheless, this is not the case in other medical imaging modalities, such as retinal fundus images.
In the vast majority of the datasets in our assembly data $\mathcal{D}_{T}=\{(\rmX_n, y_n, \rmt_{n})\}_{n=1}^{N}$, the text representation $\rmt_{n}$ is not available (see Table \ref{datasets_assembly}). Therefore, we introduce a mapping function, $\pi(\cdot):\mathcal{Y} \rightarrow \mathcal{T}$, which generates text descriptions from the categorical labels, thereby building our multi-modal dataset as $\mathcal{D}_{T}=\{(\rmX_n, y_n, \pi(y_{n}))\}_{n=1}^{N}$. 

A typical solution would be to use a {\em naive} transfer function, $\pi_{\textit{naive}}(\cdot)$, which only brings information on the imaging modality used (e.g. \citep{Liu2023} for CT volumes). In the case of fundus images, the modality prompt template would thereby be “\textit{A fundus photograph of [CLS]}”, where “\textit{[CLS]}” indicates the category name. Although this solution might integrate semantic relations of similarly-named categories, it fails to capture domain-specific hierarchies and thus may become \textit{uninformative} \citep{Menon2023}. In this work, we propose to exploit well-established \textit{domain expert knowledge} (EK) descriptions, which we denote as $\pi_{\textit{EK}}(\cdot)$. This transformation maps each category to text descriptions that contain relevant findings for each disease, as well as inter-category relationships. Given a category $y^*$, the mapping produces an ensemble of $P$ text descriptions such that $\{\rmt^*\}_1^P = \pi_{\textit{EK}}(y^*)$. For example, a text description of category “\textit{no diabetic retinopathy}” would be “{\em no relevant haemorrhages, microaneurysms or exudates”}, while the category “\textit{exudates}” could be described as “\textit{small white or yellowish-white deposits with sharp margins}". It is worth mentioning that $P$, the number of textual expert knowledge descriptions, might differ from one category to another. For additional examples, we refer the reader to Table \ref{dk_description} in Appendix. Thus, during the stochastic gradient descent optimization of the loss function in \ref{eq:loss}, and for a given sample of a categorically-labeled dataset in a batch $\mathcal{B}$, a text description is uniformly sampled from a set containing the naive prompt and the corresponding expert knowledge prompts. By encoding this domain knowledge during training, vision-language pre-training can capture stronger inter-category relations, potentially leading to richer representations. 

\subsection{Expert knowledge to enhance zero-shot inference.}
\label{ssec:dk_2}

Vision-language pre-trained models might serve as a powerful tool for zero-shot classification. This involves the generalization to \textit{unseen datasets}, which might present novel target tasks. Formally, let us define a target image, $x^{*}$, and a given set of novel categories within the target dataset, $y' \in \{C+1, ... , C'\}$. In this case, the inference is driven by the observed cosine similarity between the image representation produced by the image encoder, $\mathbf u^*$, and a representation of each category produced by the text encoder, $\mathbf v_{c'}$, using a language description of each category, $\pi(\{c'\}_{C+1}^{C'})$. Specifically, the predicted category corresponds to the maximum cosine similarity:
 \begin{align}
\label{eq:inference}
\underset{c'}{\mathrm{argmax}} 
\frac{\text{exp}({\mathbf u^*}^{T} {\mathbf v}_{c'} / \tau)}{\sum_{c\in \{C+1, ... , C'\}}\text{exp}({\mathbf u^*}^{T} {\mathbf v}_{c} / \tau)}
\end{align}

Regarding the text representation, the first works using CLIP-alike models for inference resorted to a naive prompting strategy based on category name \cite{Radford2021}. Nevertheless, recent reports indicate that the category names might overlook the full value of the additional information provided by the language modality \citep{Menon2023}. For instance, \citep{Menon2023} investigated how to enhance class representations using large language models in the context of natural images, and \citep{Wang2022, Wu2023} used domain-knowledge prompts in the context of radiology images. In line with these recent developments, we take advantage of the expert-knowledge prompts during the inference phase, in addition to their use during training, which we introduced in the previous sub-sections. This could potentially enhance discrimination in pathologies that are not seen during training, based on the description of their underlying characteristics. In this setting, for each novel category $c$, we compute its textual representation ${\mathbf v}_{c}$ in Eq. \ref{eq:inference} as the centroid of the $P$ text embeddings of the expert-knowledge prompts corresponding to category $c$.


\section{Experimental setting}
\label{sec:exp}

\subsection{Datasets} 

\paragraph{\textbf{Assemblying the dataset $\mathcal{D}_T$}} A total of 38 public datasets are assembled for training and evaluating the proposed universal model. A summary of the datasets is presented in Table \ref{datasets_assembly}. The assembled dataset combines the main tasks explored for fundus image analysis, which include: diabetic retinopathy grading \citep{Etienne2014, Porwal2020, PARAGUAY, Takahashi2017, SYSU, DDR, BRSET}, glaucoma detection \citep{LAG_Li2019, Kovalyk2022, acrima, chaksu, AIROGS, Orlando2019}, myopic maculopathy grading \citep{mmac}, and lesion segmentation \citep{Pires2014, SYSU, DDR, DiaRetDB1, Giancardo2012}. Furthermore, we included datasets that target the classification of other, less-prevalent diseases \citep{Pachade2021, 1000x39, BRSET, FUND-OCT1}. While most of the datasets contain categorical labels, we also included three datasets that contain text-based descriptions of the images: DeepEyeNet (DEN) \citep{Huang2021}, ODIR-5K, and STARE \citep{Hoover2000, Hoover2003}. For the datasets that contain pixel-level annotations of fundus images, these are converted to image-level labels. The assembly dataset $\mathcal{D}_T$ is thus composed of 288,307 images, which include 101 different categories. For further details, we refer the reader to Appendix \ref{sec:supMaterials}.

\begin{table*}[htb]
\centering
\caption{\textbf{Assembly of retinal fundus images datasets from various open-access sources.} We have developed and validated a vision-language universal model for color fundus image understanding by compiling 38 publicly available, mostly categorical datasets. The ensuing dataset assembly consists of a total of 288,307 images corresponding to 101 different categories. Among the 38 datasets, only 3 include text-based supervision of fundus images. For more detailed information on category abbreviations, we refer the reader to Appendix \ref{sec:supMaterials}.}
\label{datasets_assembly}
\scriptsize
\begin{tabular}{lcrll}
\hline
Datasets                         & \#Targets & \#Images & Labels                                            & Annotations \\ \hline
01.EYEPACS\citep{eyepacs}                     & 5         & 88,702   & noDR, mildDR, modDR, sevDR, prolDR.                & Categorical    \\
\begin{tabular}[c]{@{}l@{}}02.MESSIDOR2 \citep{Etienne2014,Krause2018} \\  \vspace{0pt}  \end{tabular}   &  \begin{tabular}[c]{@{}l@{}}9 \\  \vspace{0pt}  \end{tabular}           & \begin{tabular}[c]{@{}l@{}}1,748 \\  \vspace{0pt}  \end{tabular}    & \begin{tabular}[c]{@{}l@{}}noDR, mildDR, modDR, sevDR, prolDR, noisy, clean,\\ DME, noDME, hEX. \end{tabular} & \begin{tabular}[c]{@{}l@{}}Categorical \\  \vspace{0pt}  \end{tabular}    \\
\begin{tabular}[c]{@{}l@{}}03.IDRID \citep{Porwal2020}  \\  \vspace{0pt}    \end{tabular}               & \begin{tabular}[c]{@{}l@{}}10  \\  \vspace{0pt}   \end{tabular}           & \begin{tabular}[c]{@{}l@{}}597  \\  \vspace{0pt}   \end{tabular}    & \begin{tabular}[c]{@{}l@{}}MA, HE, hEX, sEX, noDR, mildDR, modDR, sevDR,
 \\  prolDR, noDME, nonCSDME, DME.  \end{tabular}   &              \begin{tabular}[c]{@{}l@{}}Categorical   \\  \vspace{0pt}   \end{tabular}    \\ 
\begin{tabular}[c]{@{}l@{}}04.RFMid \citep{Pachade2021}  \\  \vspace{0pt} \\  \vspace{0pt} \\  \vspace{0pt} \\  \vspace{0pt}  \end{tabular}                                 & \begin{tabular}[c]{@{}l@{}}46   \\  \vspace{0pt} \\  \vspace{0pt} \\  \vspace{0pt} \\  \vspace{0pt} \end{tabular}           &  \begin{tabular}[c]{@{}l@{}}3,200   \\  \vspace{0pt} \\  \vspace{0pt} \\  \vspace{0pt} \\  \vspace{0pt} \end{tabular}         &  \begin{tabular}[c]{@{}l@{}}DR, ARMD, MH, DN, MYA, BRVO, TSLN, ERM, LS, MS  \\ CSR, ODC, CRVO, TV, AH, ODP, ODE, ST, AION, PT, RT \\ RS, CRS, EX, RPEC, RPEC, MHL, RP, CWS, CB, ODM, \\ PRH, MNF, HR, CRAO, TD, CME, PTCR, CF, VH, MCA \\ VS, BRAO, PLQ, HPED, CL.  \vspace{0pt}  \end{tabular} & \begin{tabular}[c]{@{}l@{}}Categorical  \\  \vspace{0pt} \\  \vspace{0pt} \\  \vspace{0pt} \\  \vspace{0pt} \end{tabular}  \\
\begin{tabular}[c]{@{}l@{}}05.1000x39 \citep{1000x39}   \\  \vspace{0pt} \\  \vspace{0pt} \\  \vspace{0pt}  \end{tabular}                                 & \begin{tabular}[c]{@{}l@{}}39  \\  \vspace{0pt} \\  \vspace{0pt} \\  \vspace{0pt} \end{tabular}           &  \begin{tabular}[c]{@{}l@{}}1,000   \\  \vspace{0pt} \\  \vspace{0pt}  \\  \vspace{0pt} \end{tabular}         &  \begin{tabular}[c]{@{}l@{}}N, TSLN, LOC, mildDR, modDR, sevDR, BRVO, CRVO, G,  \\ CRAO, RD, CSR, VKH, M, ERM, MHL, MYA, HE, OA, NP, \\ sevHR, DSE, DD, CDA, RP, BCD, PRDB, MNF, VH, F, \\ hEX, YWSF, CWS, TV, CB, LS, noisy, noProlDR, prolDR.   \vspace{1pt}  \end{tabular} & \begin{tabular}[c]{@{}l@{}}Categorical  \\  \vspace{0pt} \\  \vspace{0pt} \\  \vspace{0pt} \end{tabular}  \\
06.DEN \citep{Huang2021}                  & -       & 15,708   & --                & Text    \\
07.LAG \citep{LAG_Li2019}                   & 2       & 4,854   & G, noG.                & Categorical    \\
08.ODIR-5K \citep{odir}                    & $\geq$7        & 8,000   & N, DR, G, CAT, ARMD, HR, MYA.          & Text    \\
09.PAPILA \citep{Kovalyk2022}                   & 2       & 488   &  G, N.         & Categorical    \\
10.PARAGUAY \citep{PARAGUAY} & 7  &  1,437  & noDR, mildDR, modDR, sevDR, prolDR. & Categorical  \\
11.STARE \citep{Hoover2000, Hoover2003}                    & -       & 397   & --          & Text    \\
12.ARIA \citep{Farnell2008}                                & 3          & 143         &  N, ARMD, DR.                                                 &  Categorical      \\
13.FIVES \citep{Jin2022}                                 & 6          & 800         &  noisy, clean, ARMD, DR, G, N.                                              &  Categorical \\
14.AGAR300 \citep{agar300} & 2 & 28 & DR, MA. & Categorical \\ 
15.APTOS\citep{aptos} & 5 & 5,590 & noDR, mildDR, modDR, sevDR, prolDR. & Categorical \\ 
16.FUND-OCT \citep{FUND-OCT1, FUND-OCT2} & 7 & 200 & G, N, CME, neovARMD, geoARMD, acCSR, chCSR. & Categorical \\
17.DiaRetDB1 \citep{DiaRetDB1} & 9 & 89 & IrMA, neoV, ReSD, hEX, HE, sEX, MA. & Categorical \\ 
18.DRIONS-DB \citep{Carmona2008} & 1 & 110 & noCAT, Dis. & Categorical\\
19.Drishti-GS1 \citep{Sivaswamy2014} & 2 & 100 & N, G. & Categorical \\
20.E-ophta \citep{eophta} & 2 & 463 & EX, MA.,  & Categorical  \\ 
21.G1020 \citep{Bajwa2020} & 2 & 1,020 & G, N. & Categorical \\
22.HEI-MED \citep{Giancardo2012} & 3 & 169 & EX, CWS, DN. & Categorical  \\ 
23.HRF \citep{Budai2013} & 4 & 81 & N, G, DR, noisy. & Categorical\\
24.ORIGA \citep{Zhang2010} & 2 & 650 & G, noG. & Categorical\\
25.REFUGE \citep{Orlando2019, Li2020} & 2 & 1200 & G, noG. & Categorical\\ 
26.ROC \citep{Niemeijer2010} & 1 & 100 & MA. & Categorical \\ 
\begin{tabular}[c]{@{}l@{}}27.BRSET \citep{BRSET, Physionet}  \\  \vspace{0pt}   \\  \vspace{0pt}  \end{tabular}               & \begin{tabular}[c]{@{}l@{}}24  \\  \vspace{0pt}  \\  \vspace{0pt}  \end{tabular}           & \begin{tabular}[c]{@{}l@{}}16,266  \\  \vspace{0pt}  \\  \vspace{0pt}  \end{tabular}    & \begin{tabular}[c]{@{}l@{}}noDR, mildDR, modDR, sevDR, prolDR, HE, hEX, sEX, MA,
 \\  AOD, AV, AM, noisy, clean, ME, S, NE, ARMD, BRVO, HR, \\ DN, HE, RD, MYA, ICD. \end{tabular}   &              \begin{tabular}[c]{@{}l@{}}Categorical  \\  \vspace{0pt}  \\  \vspace{0pt}  \end{tabular}    \\
28.OIA-DDR \citep{DDR} & 9 & 13,673 & noDR, mildDR, modDR, sevDR, prolDR, HE, hEX, sEX, MA.  & Categorical \\ 
29.AIROGS \citep{AIROGS} & 2 & 101,442  & G, noG  & Categorical \\
30.SYSU \citep{SYSU} & 8 & 1,220  & noDR, mildDR, modDR, sevDR, prolDR, HE, hEX, sEX. & Categorical \\ 
31.JICHI \citep{Takahashi2017} & 5 & 9,940  & noDR, mildDR, modDR, sevDR, prolDR  & Categorical \\
32.CHAKSU \citep{chaksu} & 2 & 1,345  & G, noG  & Categorical \\
33.DR1-2 \citep{Pires2014} & 7 & 1,597  & N, ReSD, hEX, DN, CWS, supHE, deepHE  & Categorical \\
34.Cataract\citep{cataract_dataset} & 4 & 601  & N, G, CAT, RS  & Categorical \\
35.ScarDat \citep{accv2018-lsd} & 2 & 997  & LS, noLS  & Categorical \\
36.ACRIMA \citep{acrima} & 2 & 705  & G, noG  & Categorical \\
37.DeepDRiD \citep{Liu2022} & 5 & 2,256  & noDR, mildDR, modDR, sevDR, prolDR  & Categorical \\
38.MMAC \citep{mmac} & 5 & 1,391  & MMg0, MMg1, MMg2, MMg3, MMg4  & Categorical \\
\hdashline
                                & $\geq$101          & 288,307         &                                                  &   \\ \hline                           

\end{tabular}
\end{table*}

\paragraph{\textbf{Standarization and augmentations}} All images are resized to a canvas of size $512\times512$, and zero-padding is applied to rectangular-shaped images to avoid distortions. Furthermore, all images are intensity-normalized to be in the $[0, 1]$ range. During training, random image augmentations are applied using horizontal flips, random rotations of $[-5, 5]$ degrees, zoom scaling in the range $[0.9, 1.1]$, and color jitter. The image resolution employed aims to preserve fine-grained lesions such as potential microaneurysms, and the particular image transformations are selected following extensive evaluations on open-community challenges \citep{Liu2022}. 

\subsection{Evaluation protocol}

The proposed foundation model is validated under two different scenarios, with regard to the target task: \textit{domain shift} (\textit{i.e.}, a new dataset consisting of categories that are used for training) and \textit{unseen categories} (\textit{i.e.} novel, unseen diseases that are not used during training). For these purposes, we omitted several datasets and categories during training. 

\textbf{\textit{Domain shift}}. We used three datasets, consisting of the main fundus image analysis tasks, for testing. Specifically, these are removed from the training data, and are: The \textbf{MESSIDOR} dataset for the task of diabetic retinopathy grading, the \textbf{REFUGE} dataset for glaucoma detection, and the \textbf{FIVES} dataset for the classification of heterogeneous diseases. 

\textbf{\textit{Unseen categories}}. To evaluate the zero-shot generalization and transferability of our method to novel diseases, we selected four categories and removed them from the training data: retinitis pigmentosa (RP), macular hole (MHL), cataract (CAT) and pathologic myopia (MYA) (see Appendix, Figure \ref{fig:incremental_images}, for the visualization of these conditions). Thus, two subsets are created: \textbf{20x3}, which contains 20 samples for normal, RP, and MHL retrieved from the 1000x39 dataset, and \textbf{ODIR200x3}, which contains 200 images for normal, CAT, and MYA retrieved from the ODIR-5K dataset. It is worth mentioning that any sample corresponding to the novel categories was not used during the training of the foundation model. We also evaluated the proposed foundation model in an unseen task and dataset: five-class myopic maculopathy grading, using \textbf{MMAC-A} dataset \citep{mmac}. We only employed samples from clinical center A and kept the challenge validation partition for testing. Table \ref{datasets_validation} provides a summary of the main datasets used for evaluation.

\textbf{\textit{Generalization after adaptation}}. We used several additional datasets to evaluate the generalization performance of the foundation model after tuning. The datasets employed for domain generalization after adaptation consist of samples that were not employed during the pre-training of FLAIR, nor during its adaptation. We replicate the same tasks in adaptation, but using samples from different datasets and centers. In particular, we used \textbf{ACRIMA} for glaucoma detection, \textbf{DeepDRiD} for diabetic retinopathy grading, and \textbf{MMAC-B}, \textit{i.e.} samples from center B, for myopic maculopathy grading, and the \textbf{RP-MHL-2} and \textbf{CAT-MYA-2} subsets for the novel categories. A detailed description of these additional subsets is presented in Appendix \ref{sec:supMaterials} and Table \ref{datasets_validation_supplemental}. 

\begin{table}[h!]
\centering
\caption{\textbf{Dataset distribution for evaluating the foundation model}. To evaluate the generalization capabilities of FLAIR, we removed several datasets from the training phase. The evaluation scenarios are: \textit{i)} domain (image-data) shifts on classes that are seen during training, and \textit{ii)} novel, unseen categories. For the latter scenario, the samples corresponding to the new target categories (\textit{i.e.}, RP, MHL, CAT, MYA, and MM) were not used during training.}
\label{datasets_validation}
\scriptsize
\begin{tabular}{lrl}
\hline
Dataset                & \multicolumn{1}{l}{\#Images} & Labels                               \\ \hline
\multicolumn{3}{l}{\textit{Domain shift}}                                    \\ \hdashline
MESSIDOR               & 1748                         & noDR, mildDR, modDR, sevDR, prolDR.  \\
FIVES                  & 800                          & N, DR, G, ARMD.                      \\
REFUGE                 & 1200                         & G, noG                               \\ \hline
\multicolumn{3}{l}{\textit{Unseen categories}}                                        \\ \hdashline
20x3                   & 60                           & N, RP, MHL                           \\ 
ODIR200x3              & 600                          & N, CAT, MYA                          \\ \hline
\multicolumn{3}{l}{\textit{Unseen categories + Domain shift}}                                    \\ \hdashline
MMAC                   & 1,391                          & MMg0, MMg1, MMg2, MMg3, MMg4                          \\ \hline
\end{tabular}
\end{table}

\subsection{Foundation model pre-training}

We designed the vision and language encoders following previous relevant literature on vision-language pre-training for medical images. In particular, we used ResNet-50 \citep{He2016} pre-trained on ImageNet \citep{Deng2009} as a vision encoder, $\theta_f$, following ConVIRT \citep{Zhang2020}, GloRIA \citep{GLoRIA} and MedCLIP \citep{Wang2022}. The text encoder $\phi_f$ is BioClinicalBERT \citep{bioclinicalbert}, similarly to MedCLIP \citep{Wang2022}. For both encoders, linear layers with $512$ output features are used as projection heads, $\theta_p$ and $\phi_p$. The proposed vision-language pre-training is performed on the assembly $\mathcal{D}_T$ dataset by minimizing Eq. \ref{eq:loss} during $15$ epochs, using a batch size of $128$ images. We use AdamW with a base learning rate of $1e^{-4}$, and a warm-up cosine scheduler during the first epoch. We employ domain knowledge descriptors for categorically-labeled data to map labels into text. For details on the used descriptors, please refer to Table \ref{dk_description}. Hereafter, we refer to the proposed foundation model as FLAIR-$\pi_{\textit{EK}}$. The training is carried out using mixed precision, on a single RTX A6000 card, and it takes about 16 hours.

\subsection{Baselines}
\label{ssec:baselines}

In the following, we describe the different baselines and methods used to assess the performance of the proposed foundation model. In particular, we benchmark the generalization and transferability capabilities of FLAIR-$\pi_{\textit{EK}}$ against other relevant strategies. Concretely, we compare to \textit{i)} vision-language models for zero-shot generalization, \textit{ii)} vision encoder pre-training for efficient transfer learning, and \textit{iii)} fully model training for each target dataset, instead of following an efficient adaptation strategy.

\subsubsection{Language-driven zero-shot classification}
\label{ssec:languageZeroShotBaselines}

\paragraph{\textbf{Vision-language pre-training (VLP)}} First, we define the baselines for the task of language-driven (\textit{i.e.}, zero-shot) classification. We use CLIP \citep{Radford2021} with its original weights, pre-trained on 400M image-text pairs from heterogeneous sources. It is worth mentioning that CLIP pre-training included several medical imaging datasets. In addition, we include BiomedCLIP \citep{BiomedCLIP}, a recently published generalist vision-language model for understanding biomedical images. This model follows a strategy that is currently gaining increasing popularity, in which the vision-language pre-training is carried out with large biomedical imaging datasets from highly diverse sets of 
modalities and tasks, e.g. radiographs, CT volumes, histology, dermatology, ultrasound images, etc., in order to obtain a common foundation model for all biomedical fields. Thus, BiomedCLIP is pre-trained using more than 15 million image-text pairs from heterogeneous medical domains, extracted from PubMed. Also, we use a basic version of the proposed foundation model, which does not integrate domain-knowledge descriptors in the training. We refer to this as FLAIR-$\pi_{\textit{naive}}$. Note that, for this basic version, the training implementations are the same as those used for the proposed knowledge-driven foundation model.

\subsubsection{\textit{Pre-train-and-adapt} baselines}
\label{ssec:pretBsaelines}

\paragraph{\textbf{Task-specific models (TSMs)}} To evaluate the benefits of incorporating language supervision during pre-training across multiple tasks, we set as baselines task-specific models, trained from the assembly of datasets. A different model is trained for each of the main evaluated tasks: diabetic retinopathy grading (TSM$_{DR}$ model), glaucoma detection (TSM$_{Glaucoma}$ model), and multiple disease classification (TSM$_{Diseases}$ model) - see Table \ref{datasets_validation} for additional details on each task. For each task, all samples categorically labeled with the corresponding task-target categories are retrieved from the assembly dataset. For example, TSM$_{DR}$ model is pre-trained using all images labeled as noDR, mildDR, modDR, sevDR, and prolDR, from the dataset assembly. The resulting pre-training sub-datasets, per task, are composed of nearly $\sim100,000$ images for TSM$_{DR}$ and TSM$_{Glaucoma}$ tasks and nearly $\sim10,000$ samples for TSM$_{Diseases}$. The task-specific models are trained using a standard multi-class cross-entropy loss. 

\paragraph{\textbf{Other pre-training baselines}} We examined additional baselines for learning pre-trained representations. Concretely, we used the features extracted from pre-trained ResNet-50 on ImageNet \citep{He2016}, which we refer to as \textit{ImageNet}. In addition, we evaluated contrastive unsupervised learning using SimCLR \citep{Chen2020} as a pre-training strategy in our scenario. Here, SimCLR is trained on the whole assembly dataset, using a batch size of $64$ images, during $15$ epochs. The same transformations used for our foundation model are applied in this setting for positive-pair augmentations.

\subsubsection{Fully-training upper-bound}
\label{ssec:supBsaelines}

\paragraph{\textbf{Dataset-specific models (Fine-tuning)}} As an upper bound, we train dataset-specific models on the target datasets. This means, instead of zero-shot generalization or lightweight adaptation of a pre-trained model, all model weights are tuned on the target dataset. More concretely, the backbone initialization is the same as the foundation model, using ResNet-50 with initialization weights pre-trained on ImageNet \citep{He2016}, as it is a popular approach for transfer learning to medical images \citep{raghu2019transfusion}. A batch size of $8$ images is used, and the model is trained via standard stochastic gradient descent during $50$ epochs using a linear classifier and multi-class cross-entropy loss. Other hyperparameters are ADAM optimizer and a learning rate of $1e^{-4}$. We refer to this strategy as \textit{Fine-tuning}. 

\subsection{Evaluation metrics}

The main figure of merit used for evaluation is accuracy per class, and the balanced average for each dataset (ACA) \citep{biranet}. In addition, task-specific metrics are incorporated into the evaluation for comparison with previous literature. In particular, DR grading is evaluated using the quadratic Cohen kappa ($\kappa$), the most popular choice in previous relevant literature \citep{Galdran2020}. In the case of Glaucoma detection, the area under receiving-operative-curve (AUC) is used as a figure of merit, following previous challenges on this topic \citep{Orlando2019,AIROGS}. For all the transferability experiments that require training or adaptation, metrics are averaged across $5$ cross-validation folds.

\section{Results}
\label{sec:results}

\subsection{Generalization}

In this section, we evaluate the generalization capabilities of the proposed model by direct prediction (\textit{i.e.} no adaptation of the trainable parameters) on several target datasets under two common scenarios: \textit{i)} domain shift, where the set of classes remains the same, but images present domain drifts, and \textit{ii)} novel classes, where the domain remains the same, but unseen classes are expected to be identified.

\subsubsection{Presence of domain shift}

First, we assess the performance of the proposed pre-trained vision-language approach, FLAIR, under domain distributional shifts, which is benchmarked against task-specific models. To achieve this, we consider two inference possibilities: using a naive mapping prompt consisting of the category names ($\pi_{\textit{naive}}(\cdot)$), and domain-knowledge descriptions of the target diseases ($\pi_{\textit{EK}}(\cdot)$). These scenarios are referred to as \textit{VLP-inference w/}$\pi_{naive}$ and \textit{VLP-inference w/}$\pi_{EK}$, respectively. Furthermore, to further understand the impact of integrating domain-knowledge descriptions during training, we evaluate our model when both naive and the proposed mapping are used, whose models are referred to as FLAIR-$\pi_{naive}$ and FLAIR-$\pi_{EK}$.

\begin{table}[h!]
\setlength{\tabcolsep}{3pt}
\centering
\caption{\textbf{Generalization under domain shifts.} The results obtained by direct prediction, \textit{i.e.}, no adaptation, of the pre-trained language-driven models under the presence of domain shifts. The proposed approach, FLAIR, is compared to task-specific models (pre-trained following the standard supervised paradigm), and the existing literature on the corresponding specific tasks: diabetic retinopathy (MESSIDOR), disease classification (FIVES) and glaucoma detection (REFUGE). For each task, we provide representative figures of merit in the literature. The proposed method, FLAIR-$\pi_{EK}$ is shadowed, whereas the best results are highlighted in bold.}
\label{generalization}
\scriptsize
\begin{tabular}{lccccc}
\hline
\multicolumn{1}{c}{Method}                                & \multicolumn{4}{c}{Datasets}       \\ \cline{2-5}
                                      & MESSIDOR      & FIVES & REFUGE & \textbf{Avg.}    \\ 
                                      & (ACA/$\kappa$)& (ACA) & (AUC) & (ACA)     \\ \hline
\textit{Prior literature}             &               &       & &           \\ \hdashline
{DR$_{graduate}$} \citep{DRgraduate}   & 0.596/0.710   & -     & - & -         \\ 
{AST} \citep{Galdran2020}              & 0.634/0.797   & -     & - & -         \\
{AIROGS$_{lb}$} \citep{AIROGS}       & -             & -       & $[0.88, 0.94]$ & -    \\\hline
\textit{Task-specific models (TSMs)}                &               &       &           \\ \hdashline
TSM$_{DR}$                                    & 0.550/\textbf{0.772}   & -     & -  & -       \\ 
TSM$_{Diseases}$                              & -             & 0.381 & -  & -       \\ 
TSM$_{Glaucoma}$                              & -             & -     & 0.904 & -    \\ \hline
\textit{VLP - inference w/ $\pi_{\textit{naive}}$}                          &               &       &           \\ \hdashline
CLIP                                              & 0.237/0.140 & 0.250 & 0.470 & 0.313  \\
BiomedCLIP                                        & 0.224/0.201 & 0.416 & 0.540 & 0.392  \\
FLAIR-$\pi_{\textit{naive}}$                      & 0.545/0.662 & 0.732 & 0.899 & 0.697 \\ 
\cellcolor{Gray}FLAIR-$\pi_{\textit{EK}}$         & \cellcolor{Gray}0.602/0.711   & \cellcolor{Gray}0.719 & \cellcolor{Gray}0.918 & \cellcolor{Gray}\textbf{0.734} \\ \hline
\textit{VLP - inference w/ $\pi_{\textit{EK}}$}                          &               &       &           \\ \hdashline
CLIP                                              & 0.200/0.000 & 0.256 & 0.433 & 0.318 \\
BiomedCLIP                                        & 0.207/0.188 & 0.415 & 0.624 & 0.368 \\
FLAIR-$\pi_{\textit{naive}}$                      & 0.442/0.694 & \textbf{0.744} & 0.871 & 0.637 \\ 
\cellcolor{Gray}FLAIR-$\pi_{\textit{EK}}$         & \cellcolor{Gray}\textbf{0.604}/\textbf{0.772}   & \cellcolor{Gray}0.735 & \cellcolor{Gray}\textbf{0.920} & \cellcolor{Gray}\textbf{0.740} \\ \hline
\end{tabular}
\end{table}

The results from this experiment are presented in Table \ref{generalization}. First, we can observe that standard vision-language pre-training (VLP), \textit{i.e.}, FLAIR-$\pi_{\textit{naive}}$, provides comparable results to those obtained by traditional supervision models on DR and Glaucoma tasks. In contrast, vision-language pre-training methods outperform by a large margin ($+34\%$) its standard supervised counterpart, TSM$_{Diseases}$, on the FIVES dataset. It is important to note that the task-specific models TSM$_{DR}$ and TSM$_{Glaucoma}$ were trained with a large number of labeled target task samples, whereas the size of the available dataset for TSM$_{Diseases}$ was much smaller. This may explain the significant discrepancies in performance differences between task-specific and VLP models across tasks. Furthermore, these results showcase a clear benefit of pre-trained vision-language models, as more datasets covering a wide variability can be used during training, circumventing the problem of small labeled datasets for a specific task. The second important observation from Table \ref{generalization} is that introducing domain-knowledge descriptions in the proposed foundation model (FLAIR-$\pi_{\textit{EK}}$) typically yields significant improvements on MESSIDOR (diabetic retinopathy grading) and REFUGE (glaucoma detection), whereas the performance on FIVES (diseases classification) is slightly degraded. We advocate that the large improvement gains observed are due to introducing hierarchical relationships between local findings in DR grades via integrating the domain-knowledge prompts in FLAIR-$\pi_{EK}$. Finally, the proposed approach obtains results that are on par with task-specific solutions in the literature, such as cost-sensitive optimization for DR grading \citep{Galdran2020} and optic-disk segmentation for Glaucoma detection (AIROGS leaderboard \citep{AIROGS}). In contrast, the proposed model is universal, providing a general representation of fundus images, which results in important performance gains even compared to standard supervised methods in targeted-task scenarios. We stress that, due to specific implementation differences, it might be difficult to establish direct comparisons with prior works in the standard targeted-task setting.

\subsubsection{Performance on novel classes}

We now present empirical evaluations of FLAIR in the zero-shot scenario, \textit{i.e.} there is no adaptation of the foundation model to novel, unseen categories. In this context, we study three different strategies for text-prompt design. First, we present the prompt generation as an anomaly detection task, in which all unseen diseases are treated as anomalies. In this case, the prompts used as input of the text encoder are simply either "\textit{normal}" or "\textit{disease}". Then, we introduce the notion of each novel disease when generating the text prompts. The first strategy involves using directly the names of the unseen diseases as prompts, in a \textit{naive} way. Last, and to further exploit the learning power of vision-language models, we propose to \textit{design} domain knowledge prompts, which briefly describe the differential findings on fundus images corresponding to each condition. For example, instead of employing the category name \textit{"cataract"}, as in the naive manner, we use the following finding as input text prompt: \textit{"opacities in the macular area"}. In Appendix, Fig. \ref{fig:incremental_images}, we provide additional prompts designed for the unseen categories. 

\begin{table}[h!]
\centering
\caption{\textbf{Generalization to unseen categories (zero-shot classification).} Transferability of the proposed foundation model to new tasks via prompt-based inference. We evaluated three different strategies for generating text prompts: anomaly detection (\textit{i.e.}, “\textit{normal}”/“\textit{disease}”), classification via naive prompt (\textit{i.e.}, the new disease name) and designed prompts (\textit{i.e.}, domain-knowledge descriptors). The metric presented is the accuracy for each category. The proposed method, FLAIR-$\pi_{EK}$, is shadowed, whereas the best results are in bold.}
\label{prompt_classification}
\scriptsize
\begin{tabular}{lcccc}
\hline
\multicolumn{1}{c}{Method} & \multicolumn{4}{c}{Datasets} \\ \cline{2-5}
                                             & 20x3   & ODIR200x3   & MMAC & \textbf{Avg.}      \\ \hline 
\multicolumn{5}{l}{\textit{Anomaly Detection Inference (i.e. "normal/disease")}} \\ \hdashline
CLIP                                         & 0.600                  & 0.591                  & 0.668 & 0.619      \\
BiomedCLIP                                   & 0.538                  & \textbf{0.785}         & 0.679 & 0.667     \\
FLAIR-$\pi_{\textit{naive}}$                 & 0.550                  & 0.551                  & 0.565 & 0.555     \\
\cellcolor{Gray}FLAIR-$\pi_{\textit{EK}}$    & \cellcolor{Gray}\textbf{0.812}  & \cellcolor{Gray}0.668  & \cellcolor{Gray}\textbf{0.712} & \cellcolor{Gray}\textbf{0.730}      \\ \hline
\multicolumn{5}{l}{\textit{Inference with Naive Prompts - $\pi_{\textit{naive}}$ (e.g. "cataract")}} \\ \hdashline 
CLIP                                         & 0.367                  & 0.445                  & \textbf{0.228} & 0.346                  \\
BiomedCLIP                                   & 0.750                  & \textbf{0.727}         & 0.027 & 0.501                  \\
FLAIR-$\pi_{\textit{naive}}$                 & 0.567                  & 0.447                  & 0.187 & 0.400                  \\
\cellcolor{Gray}FLAIR-$\pi_{\textit{EK}}$    & \cellcolor{Gray}\textbf{0.850}  & \cellcolor{Gray}0.483  & \cellcolor{Gray}0.199 & \cellcolor{Gray}\textbf{0.510} \\ \hline
\multicolumn{5}{l}{\textit{Inference with Expert Knowledge Prompts - $\pi_{\textit{EK}}$}} \\
\multicolumn{5}{l}{\textit{(e.g. "opacity in the macular area")}} \\ \hdashline
CLIP                                         & 0.333                  & 0.480                  & 0.183 & 0.332                 \\
BiomedCLIP                                   & 0.617                  & 0.583                  & 0.274 & 0.491                 \\
FLAIR-$\pi_{\textit{naive}}$                 & 0.650                  & 0.470                  & 0.214 & 0.444                \\
\cellcolor{Gray}FLAIR-$\pi_{\textit{EK}}$    & \cellcolor{Gray}\textbf{0.983}  & \cellcolor{Gray}\textbf{0.667} & \cellcolor{Gray}\textbf{0.400} & \cellcolor{Gray}\textbf{0.683}  \\ \hline
\end{tabular}
\end{table}

Table \ref{prompt_classification} reports the results. One may observe that training CLIP on the assembly dataset with categorical prompts (\textit{i.e.}, FLAIR-$\pi_{\textit{naive}}$) typically yields large performance gains in comparison to the standard CLIP model, particularly in the multi-class classification scenario (\textit{middle and bottom sections of the Table}). However, FLAIR-$\pi_{\textit{naive}}$ largely fails to differentiate between the target diseases. In contrast, integrating domain-specific knowledge brings substantial performance gains. First, using specific domain descriptors during training results in average improvements of $+11\%$ over the naive FLAIR version (see Table \ref{prompt_classification}, \textit{Prompt naive} section). The observed gains may be explained by the fact that the text prompts used in the proposed $\pi_{\textit{EK}}$ favor a richer text embedding, with hierarchical and domain-specific knowledge. Interestingly, further leveraging these better-designed prompts during inference enhances the different models by around $13\%$ (20x3), $18\%$ (ODIR200x3), and $20\%$ (MMAC), resulting in a gap with FLAIR-$\pi_{\textit{naive}}$ of nearly $33\%$ in 20x3 dataset, $20\%$ in ODIR200x3, and $20\%$ for MMAC (see Table \ref{prompt_classification}, section \textit{Expert Knowledge Prompts}). More significantly for the medical domain, the proposed model substantially outperforms CLIP across all the scenarios, with differences going up to $65\%$. This shows that universal vision-language models trained on general computer vision tasks yield suboptimal results in specialized medical imaging fields. The obtained results demonstrate that, in the absence of large datasets with text-based supervision, samples with categorical labels could still be exploited to train powerful vision-language representations, by encoding domain expert knowledge into text supervision. Furthermore, FLAIR reaches promising performances on the anomaly detection task, which does not require defining the diseases on the target dataset. Finally, it is worth mentioning the limitations observed using a generalist model, such as BiomedCLIP, for zero-shot generalization on retinal fundus images. First, even though this model improves CLIP consistently across all the tasks (see Tables \ref{generalization} and \ref{prompt_classification}), the results obtained in comparison to the specialized model (\textit{i.e.} FLAIR-$\pi_{\textit{EK}}$) are considerably lower, especially in the tasks reported in Table \ref{generalization}. Furthermore, while BiomedCLIP presents a competitive performance on one dataset for generalizing to unseen categories (\textit{i.e.}, ODIR200x3), its results degrade when using descriptive prompts at inference, which highlights its limited ability to encode specialized hierarchical expert knowledge. These results showcase that, despite the focus of the recent literature on generalist models, the use of domain-specialized models such as FLAIR shows more promising results in the context of fundus imaging. We anticipate that this might be case in other medical-imaging domains. It should be noted, however, that it is difficult to establish comparisons with BiomedCLIP, since the lack of transparency in the description of these massive databases makes it difficult to know whether the model has been trained on tasks used for testing or not.

\subsection{Transferability}

We now evaluate the capabilities of FLAIR-$\pi_{\textit{EK}}$ to transfer the learned representations to downstream domains and tasks using minimal, efficient tuning. We cover the scenario in which the trained feature extractor is frozen, and only a lightweight module with trainable parameters (\textit{i.e.} Adapter) is added on top. Concretely, two strategies are evaluated: \textit{i)} Linear Probing using only the image modality, and \textit{ii)} vision-language Adapters, which combine both modalities during adaptation.

\subsubsection{Linear Probing}
\label{sec:lp}

\textit{\textbf{Adaptation strategy}}. We use the features extracted from the vision encoder $\theta_f(\cdot)$ as input of an additional linear classifier, whose parameters are fine-tuned during adaptation. Concretely, we employed the same multi-class logistic regression optimizer as in CLIP paper \citep{Radford2021}, \textit{\textit{i.e.}}, L-BFGS \citep{optim}. This strategy, which is commonly referred to as Linear Probe (LP) in the literature, is further validated empirically in the ablation experiments presented in Section \ref{ablation_experiments}.

\textit{\textbf{Adaptation resources}}. To evaluate common scenarios, two data regimes are studied: \textit{i)} \textit{low data regime}, in which only a few support (\textit{i.e.}, labeled) images for each category in the target dataset are retrieved for adaptation ($k=\{1, 5, 10\})$, and \textit{ii)}: \textit{large data regime}, where a large percentage of the dataset is used during adaptation, \textit{i.e.} $\{20\%, 40\%, 60\%, 80\%\}$. 

\textit{\textbf{Training/testing splits}}. The test subset remains the same across all data regimes, using $20\%$ of the target dataset, with the exemption of MMAC, for which we employed the official challenge validation partition for testing. From the training subset, only a number of samples corresponding to the target data regime are randomly retrieved for adapting the pre-trained model. This process is averaged across 5 cross-validation folds. 

\textit{\textbf{Baselines}}. To validate the benefits of the proposed foundation model, other common transfer learning strategies are evaluated: \textit{i)}: LP of task-specific models (TSMs), in which the model is pre-trained using only all samples that contain a subset of classes included in the assembly dataset, in a standard supervised way, \textit{ii)}: LP with unsupervised pre-training, using SimCLR, \textit{iii)}: LP with transfer learning from natural images, using the backbone pre-trained on ImageNet, and \textit{iv)}: dataset-specific models via fully fine-tuning all network parameters on the target dataset, with weights pre-trained on ImageNet. For more details on the different baselines, we refer the reader to Sections \ref{ssec:pretBsaelines} and \ref{ssec:supBsaelines}.

\begin{figure}[h!]
    \begin{center}

           \subfloat{\includegraphics[width=0.92\linewidth]{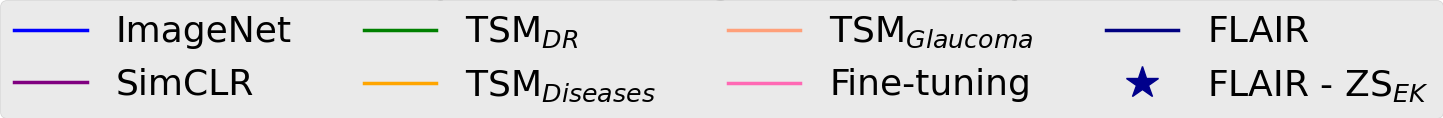}} 
          
          \subfloat{\includegraphics[height=0.31\linewidth]{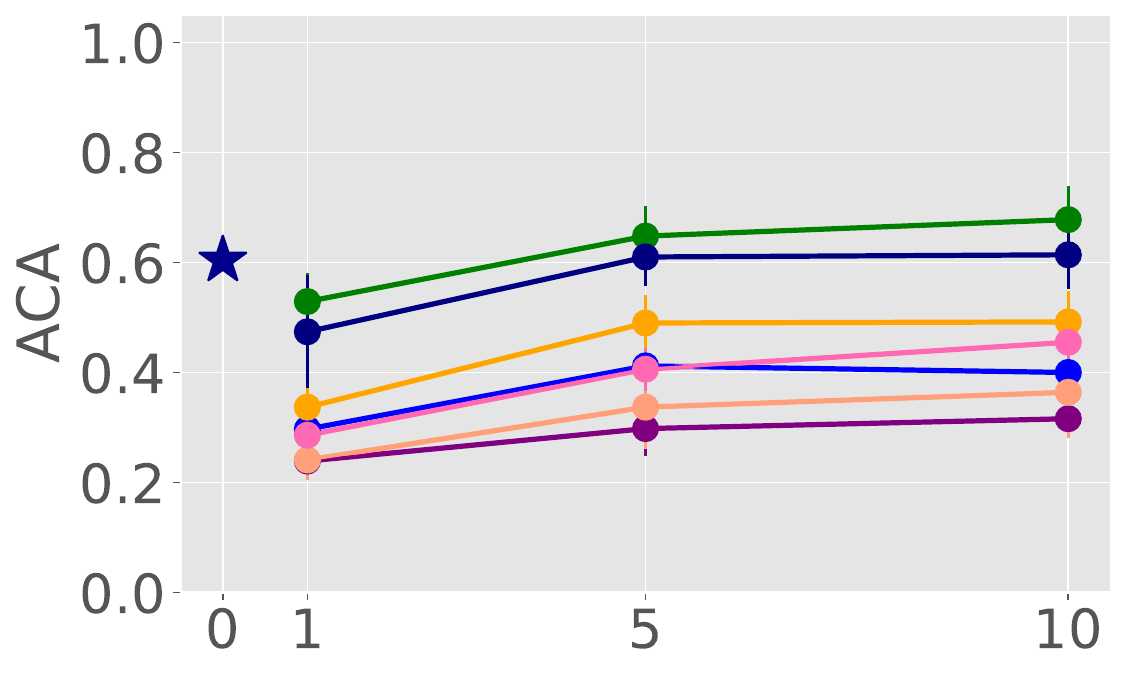}}
          \subfloat{\includegraphics[height=0.31\linewidth]{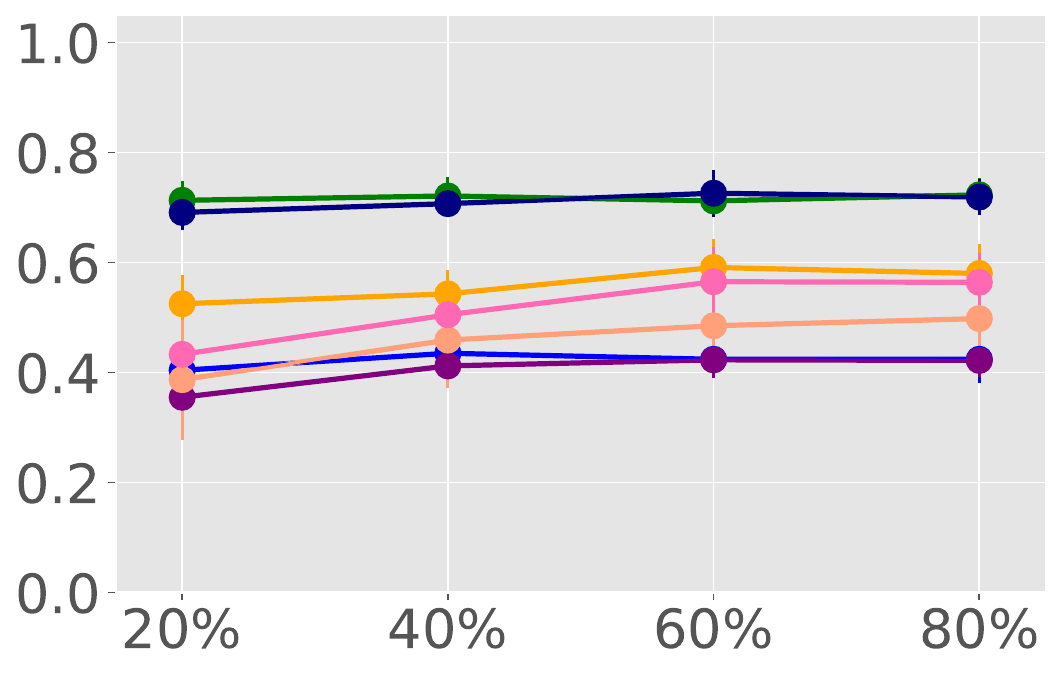}}

        \scriptsize{MESSIDOR - \textit{Domain shift - DR grading}}

          \subfloat{\includegraphics[height=0.31\linewidth]{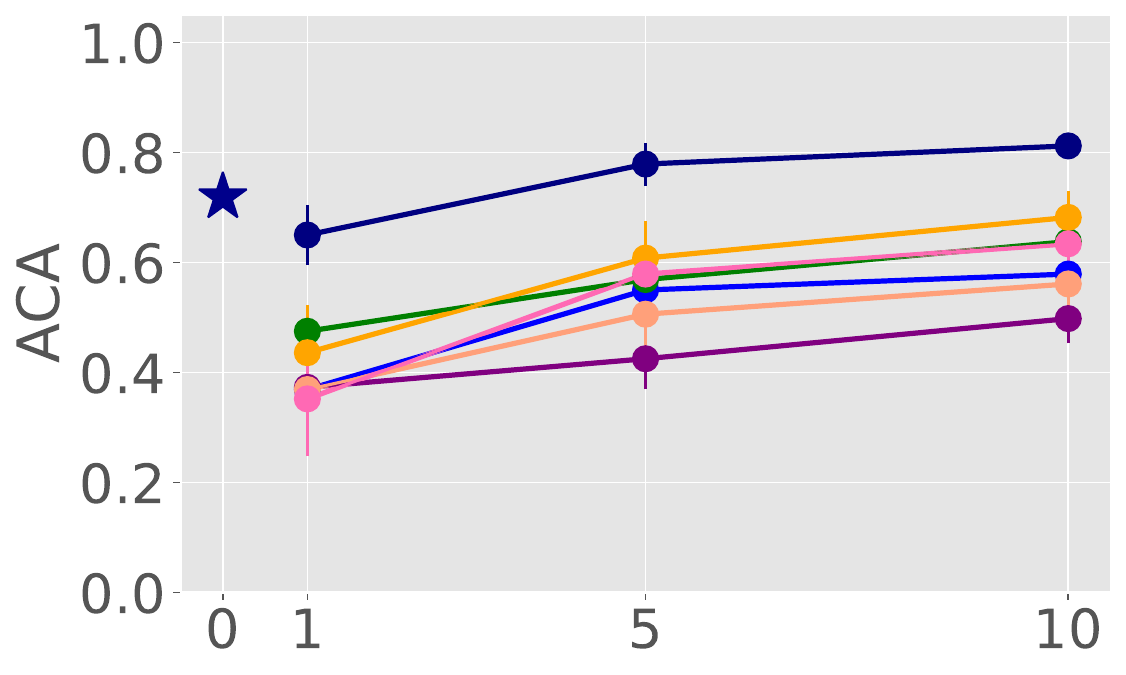}}
          \subfloat{\includegraphics[height=0.31\linewidth]{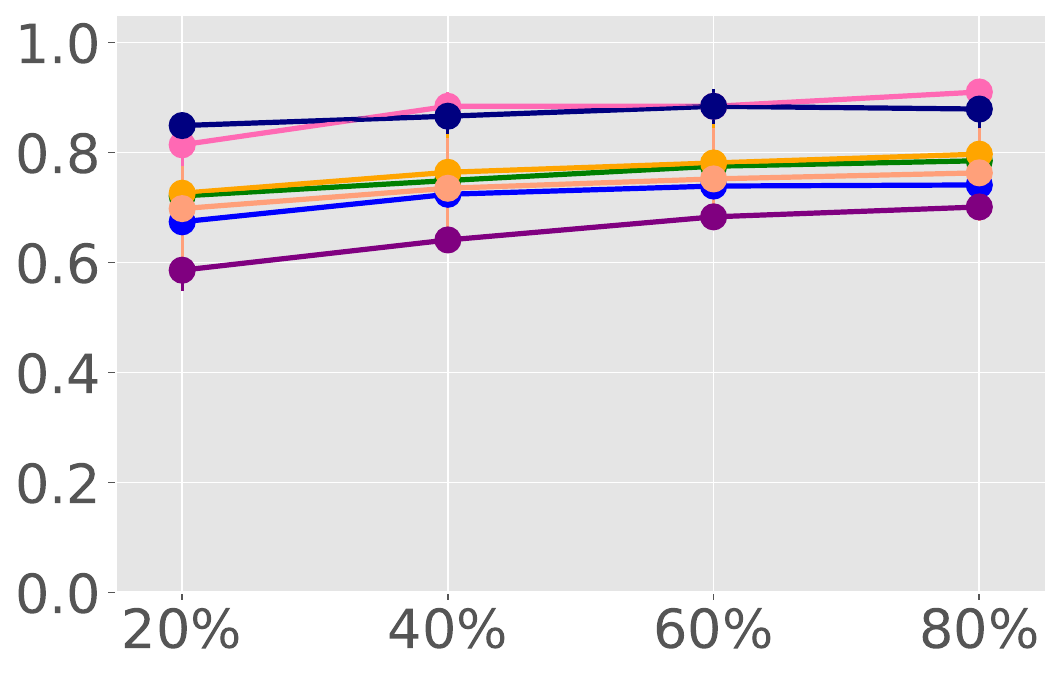}}

         \scriptsize{FIVES - \textit{Domain shift - Diseases}}

          \subfloat{\includegraphics[height=0.31\linewidth]{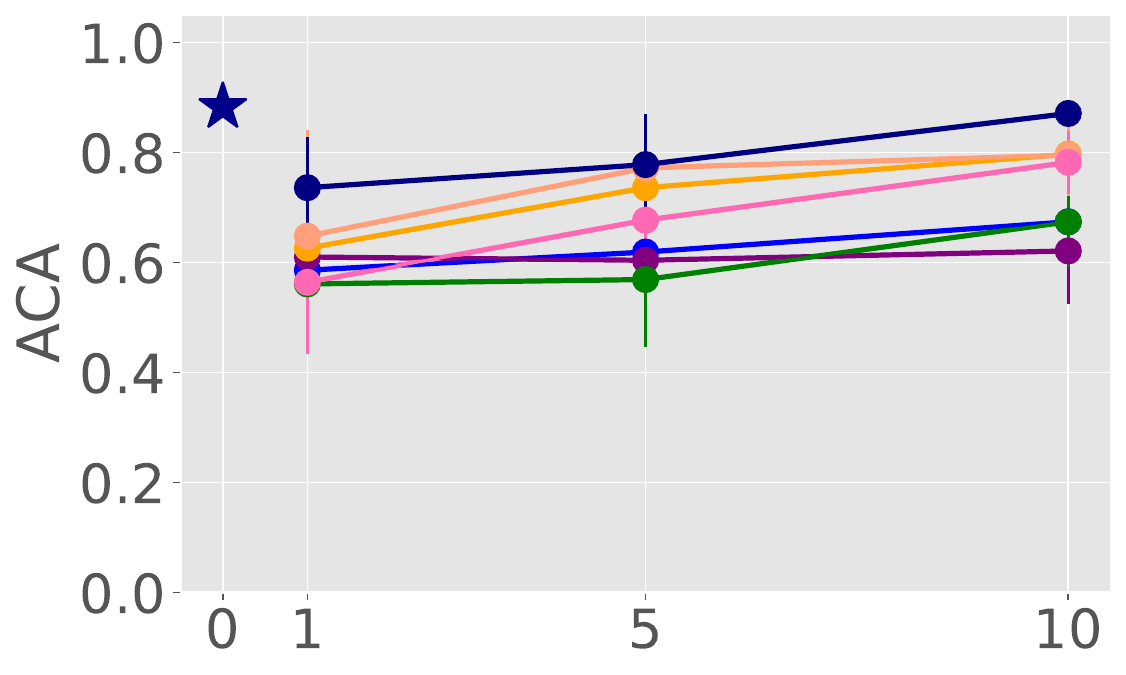}}
          \subfloat{\includegraphics[height=0.31\linewidth]{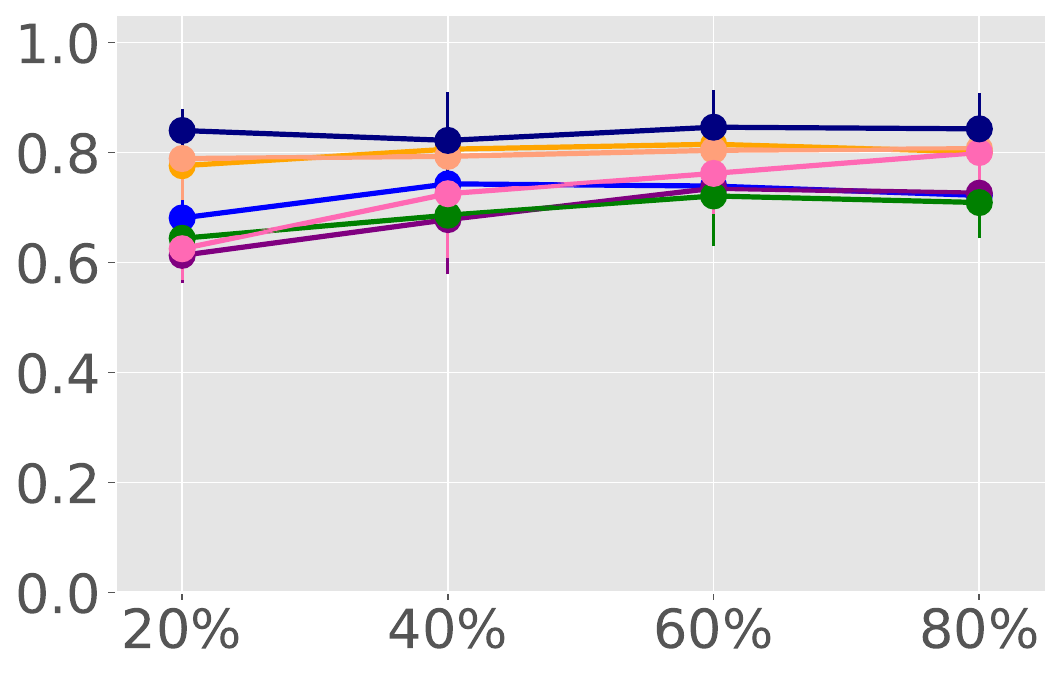}}

         \scriptsize{REFUGE - \textit{Domain shift - Glaucoma}}

          \subfloat{\includegraphics[height=0.31\linewidth]{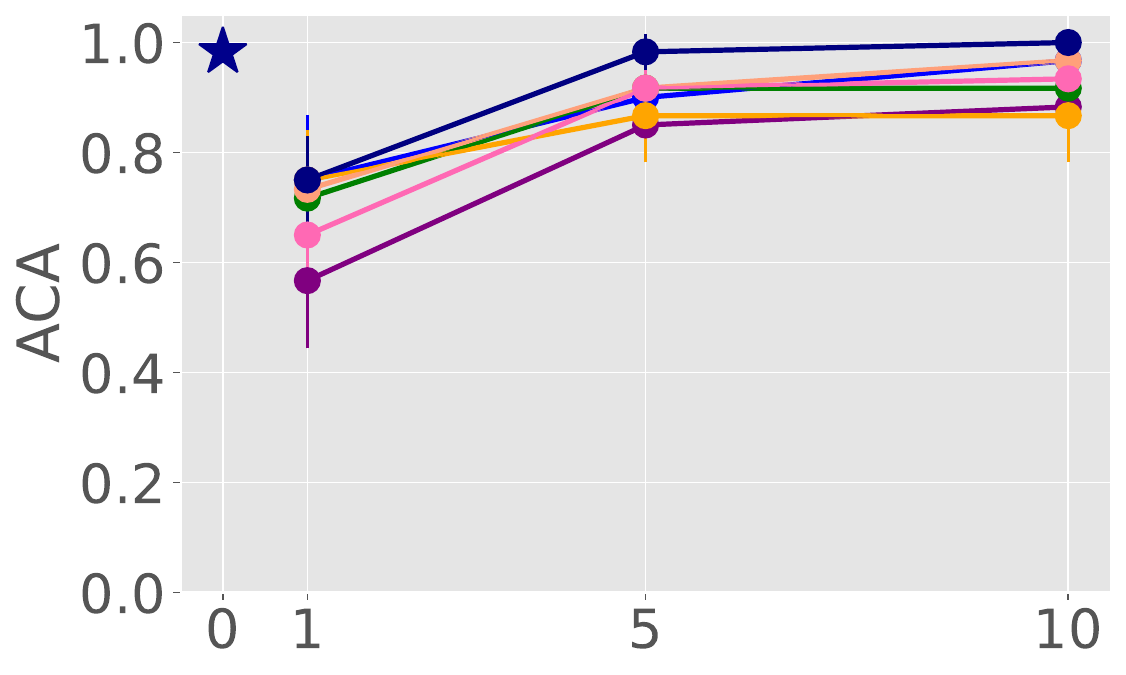}}
          \subfloat{\includegraphics[height=0.31\linewidth]{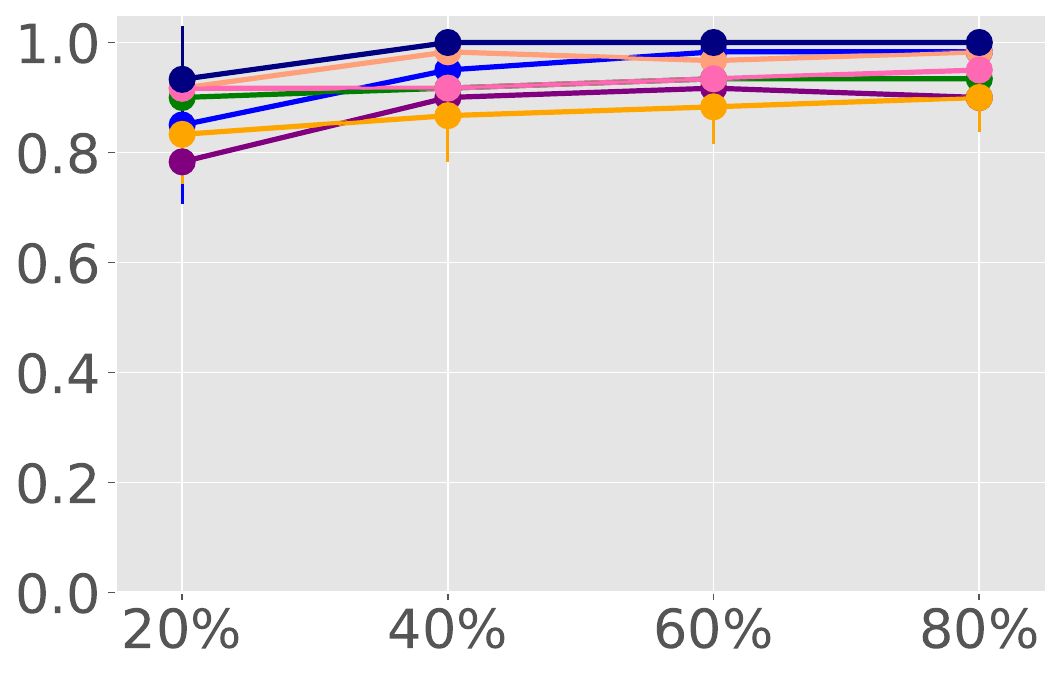}}

         \scriptsize{20x3 - \textit{Unseen categories - N, RP, MH}}
        
          \subfloat{\includegraphics[height=0.31\linewidth]{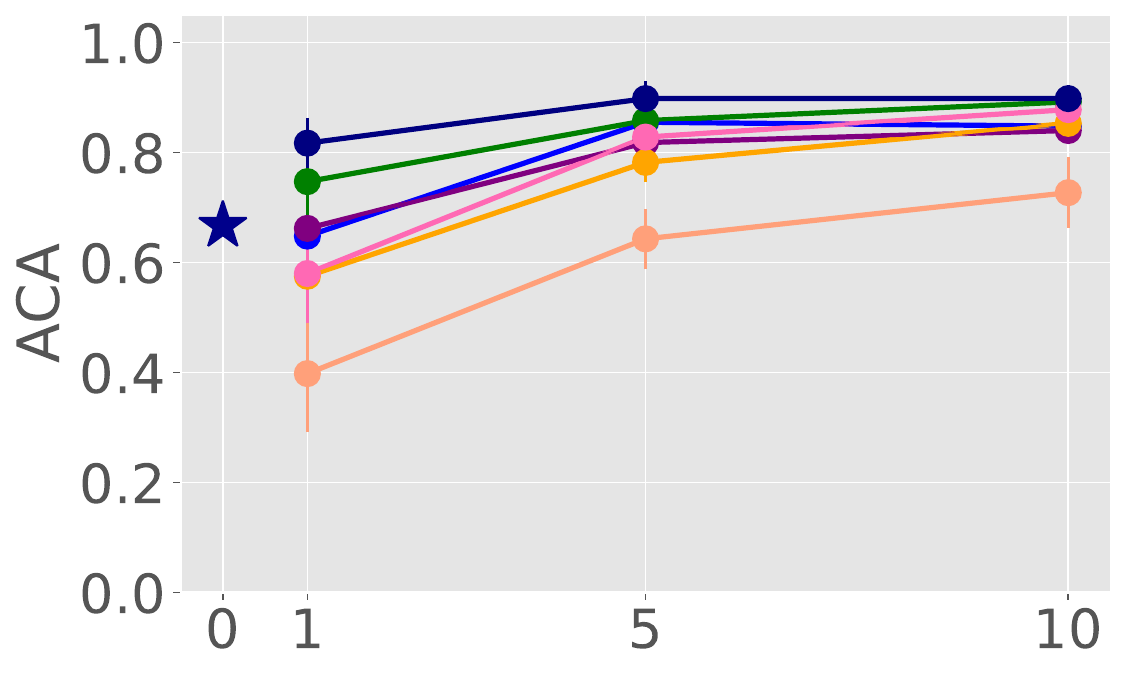}}
          \subfloat{\includegraphics[height=0.31\linewidth]{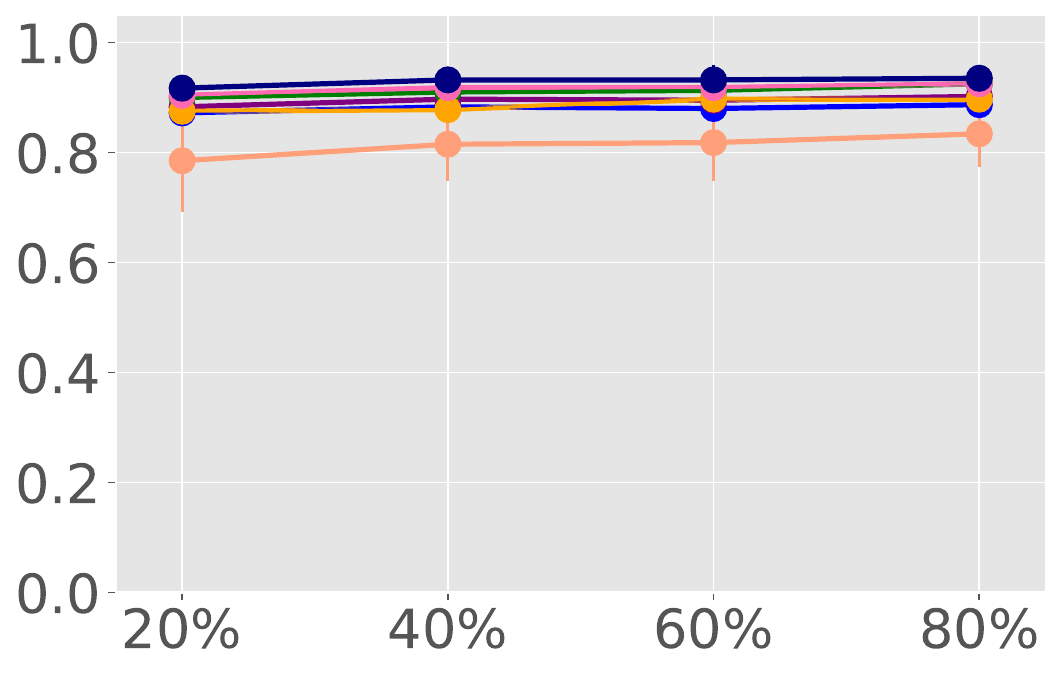}}

          \scriptsize{ODIR200x3 - \textit{Unseen categories - N, CAT, MYA}}

          \subfloat{\includegraphics[height=0.34\linewidth]{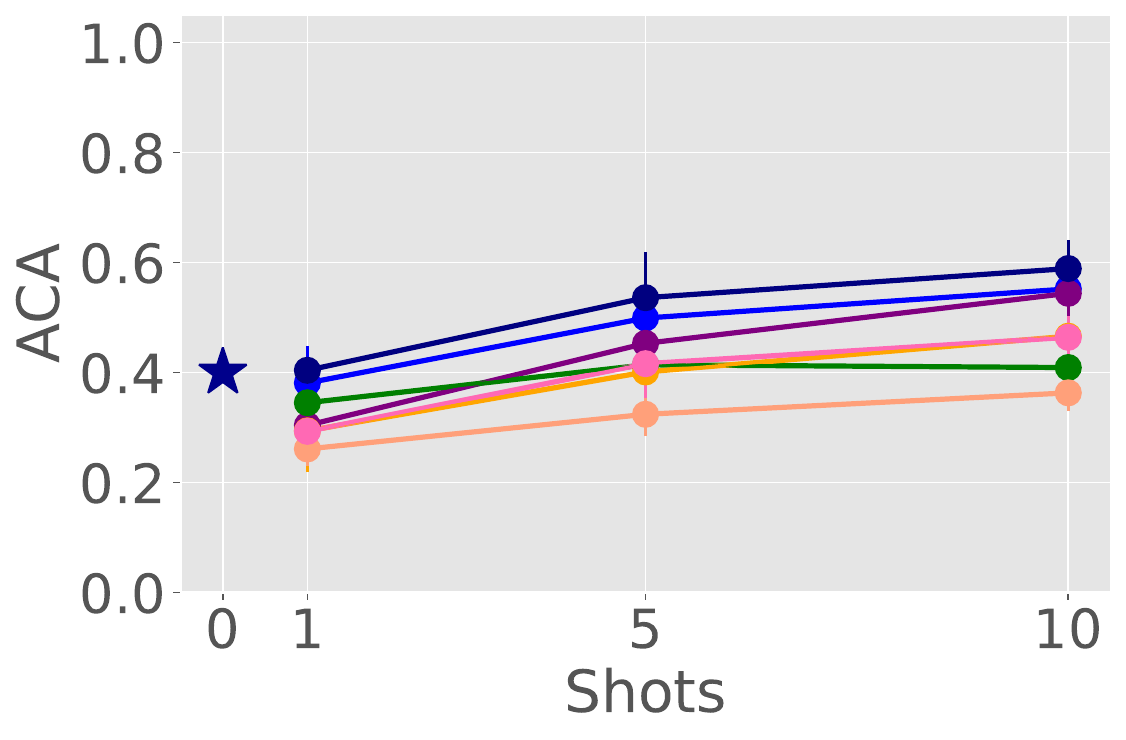}}
          \subfloat{\includegraphics[height=0.34\linewidth]{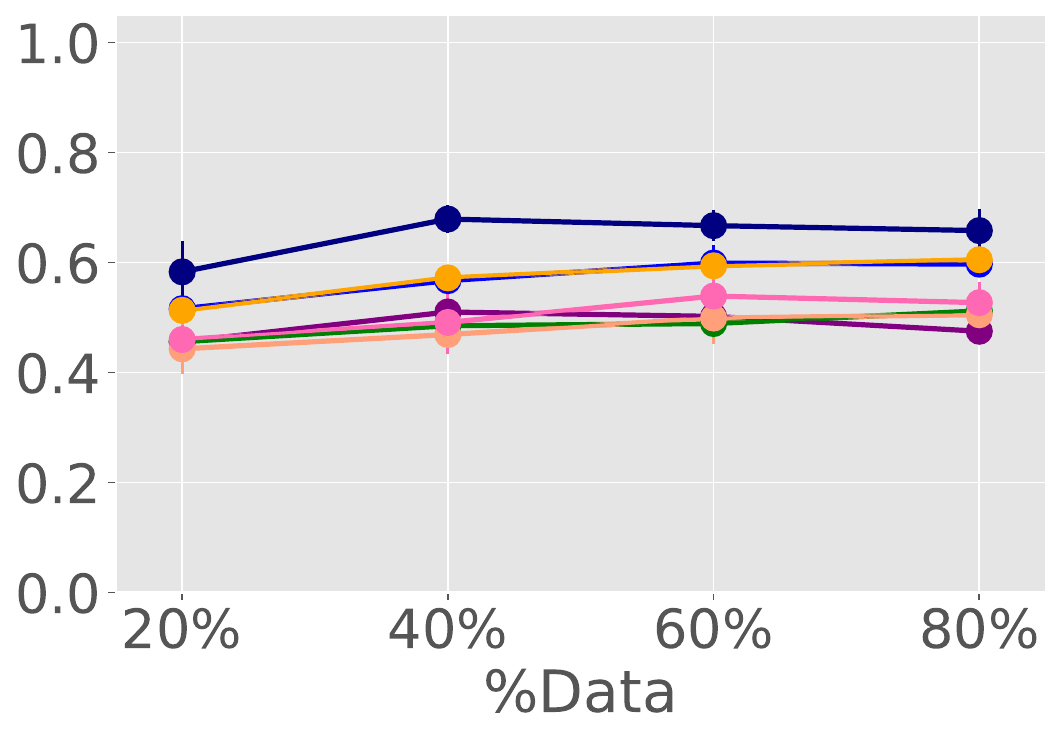}}

          \scriptsize{MMAC - \textit{Unseen categories - MM grading}}

        \caption{\textbf{Transferability.} Results of transferring the feature representations of the pre-trained models to downstream domains and tasks in the low-data (\textit{left column}) and large-data (\textit{right column}) regimes. The results were obtained by adjusting a linear-probe classifier. The metric presented is the average accuracy, averaged across 5 cross-validation folds. ZS: zero-shot (\textit{i.e.}, prompt-based).}
        \label{fig:transferability}
    \end{center}
\vspace{-6mm}
\end{figure}

\begin{figure*}[h!]
    \begin{center}

          \subfloat{\includegraphics[width=0.55\linewidth]{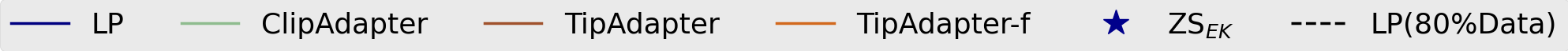}}
          \vspace{-2mm}
          \subfloat{\includegraphics[height=0.15\linewidth]{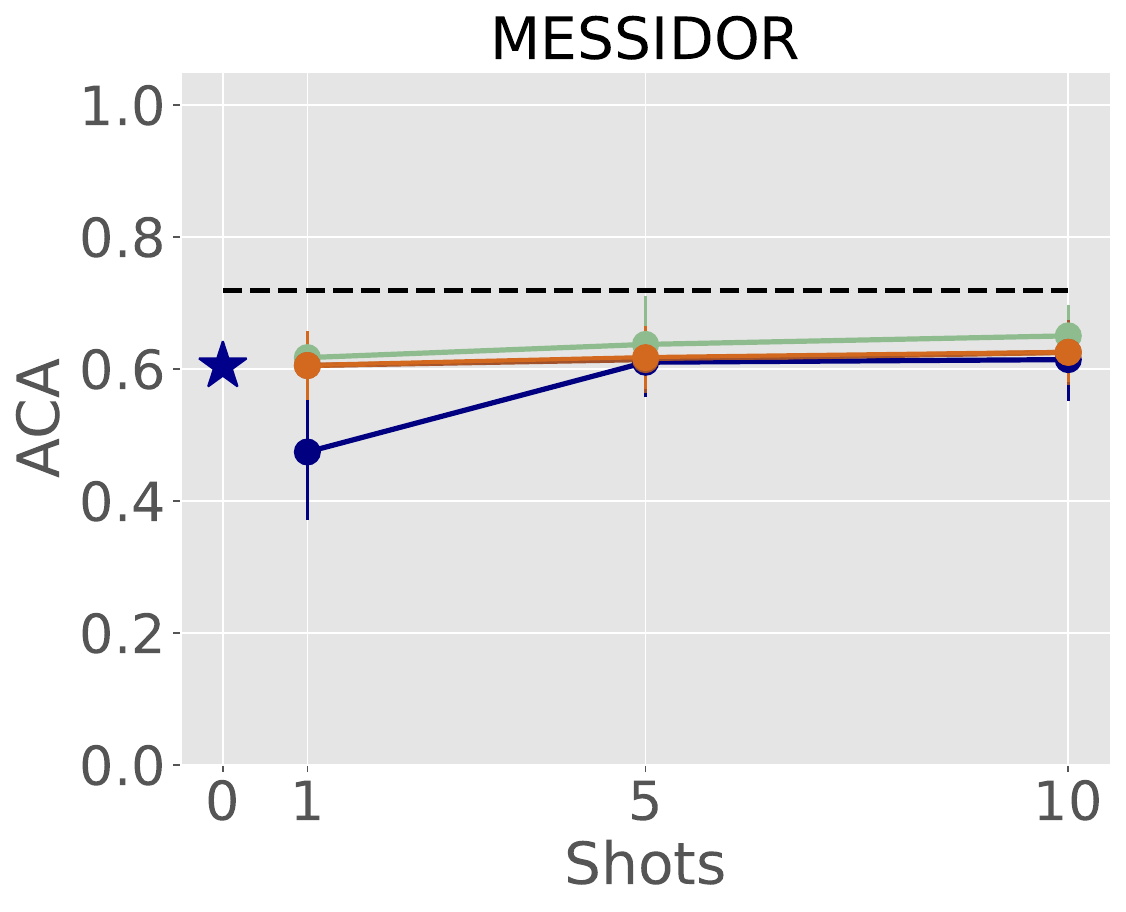}}
          \subfloat{\includegraphics[height=0.15\linewidth]{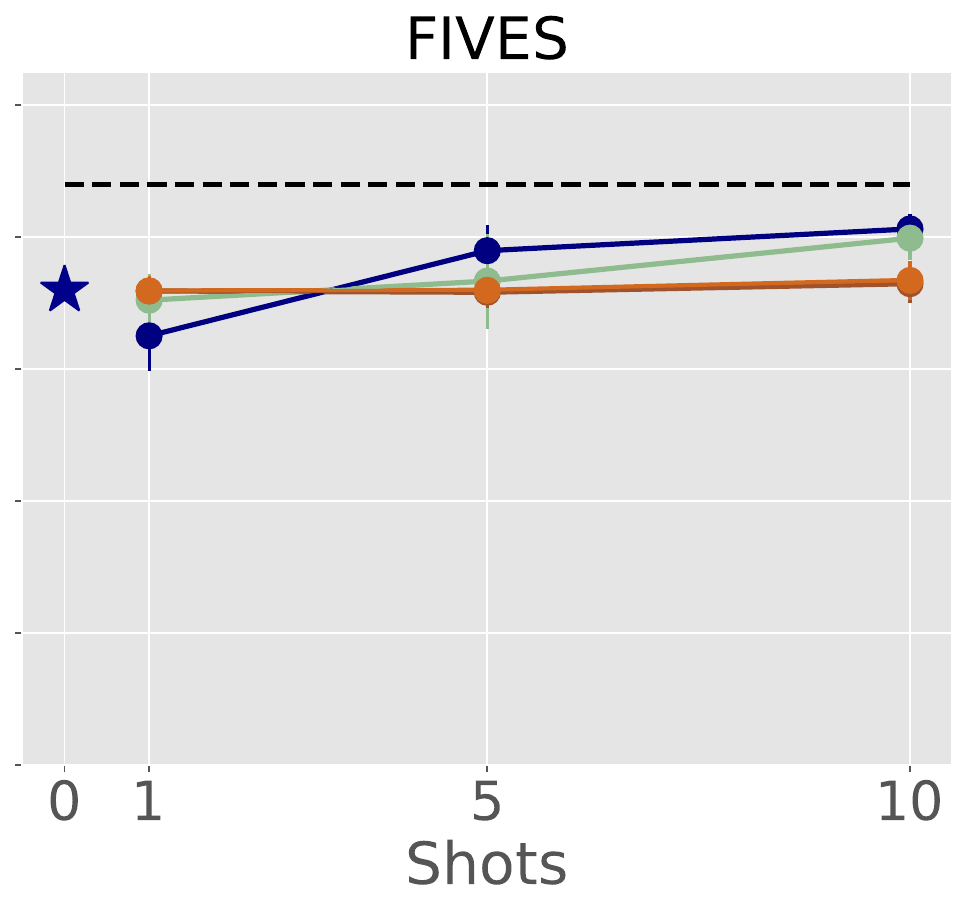}}
          \subfloat{\includegraphics[height=0.15\linewidth]{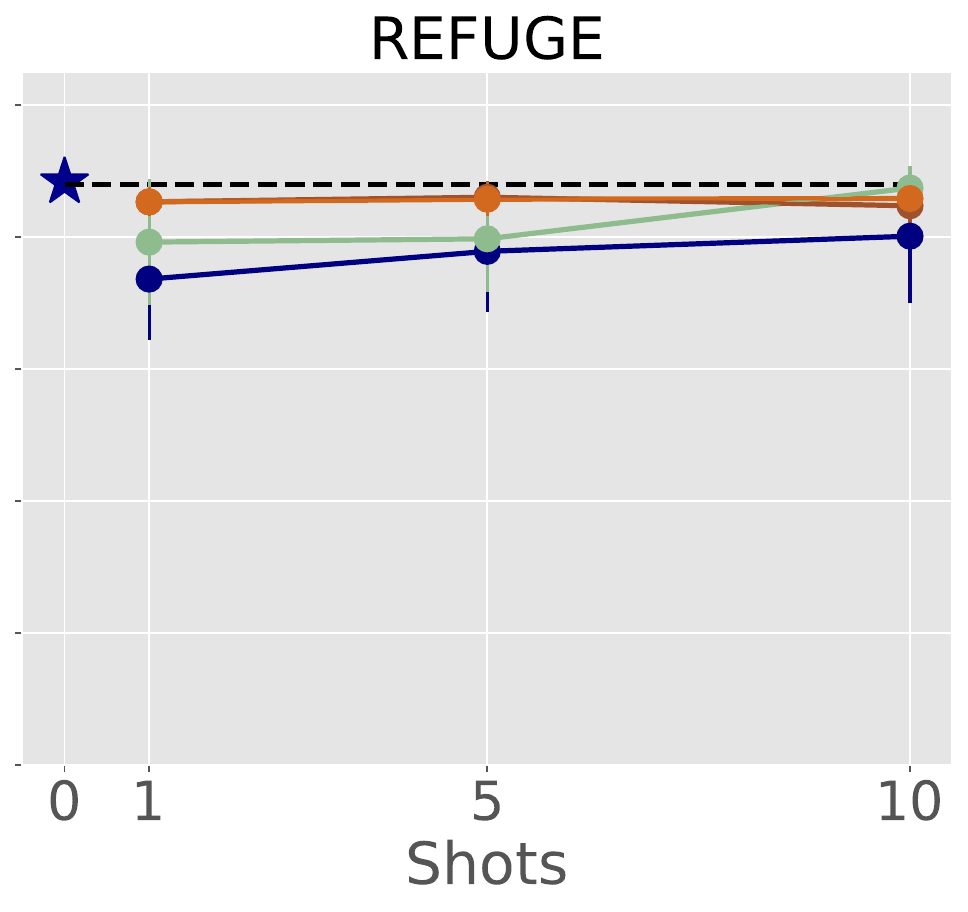}}
          \subfloat{\includegraphics[height=0.15\linewidth]{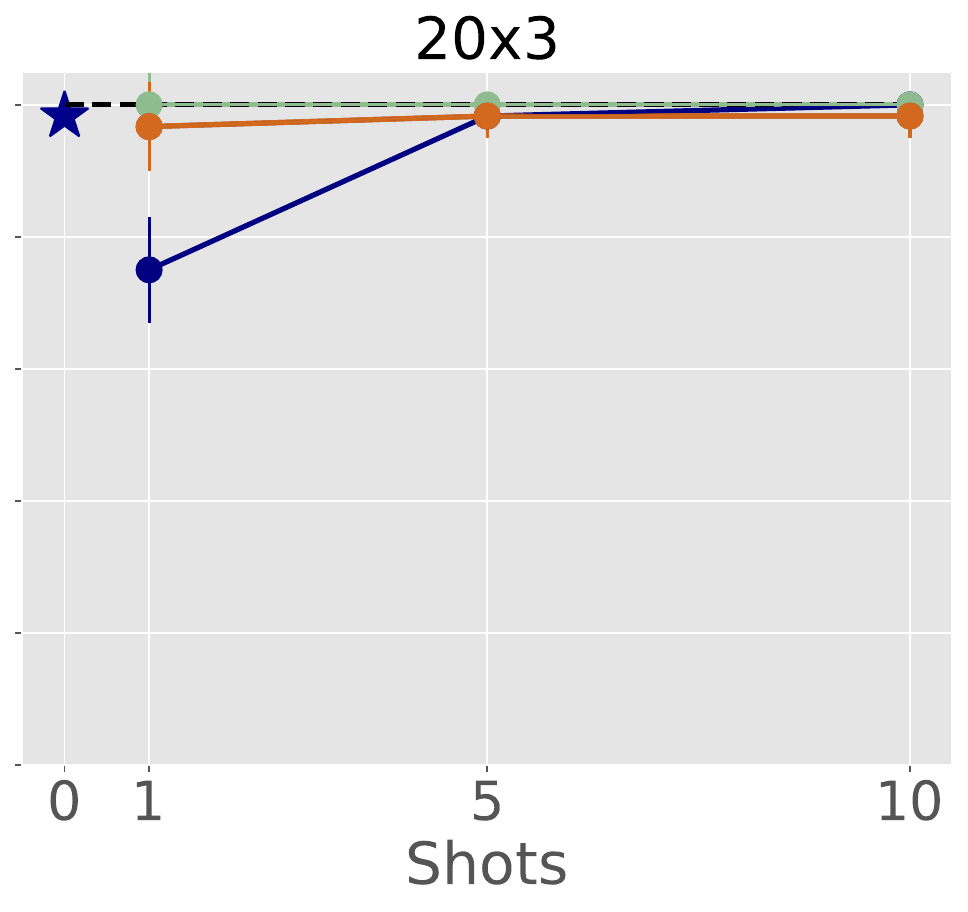}}
          \subfloat{\includegraphics[height=0.15\linewidth]{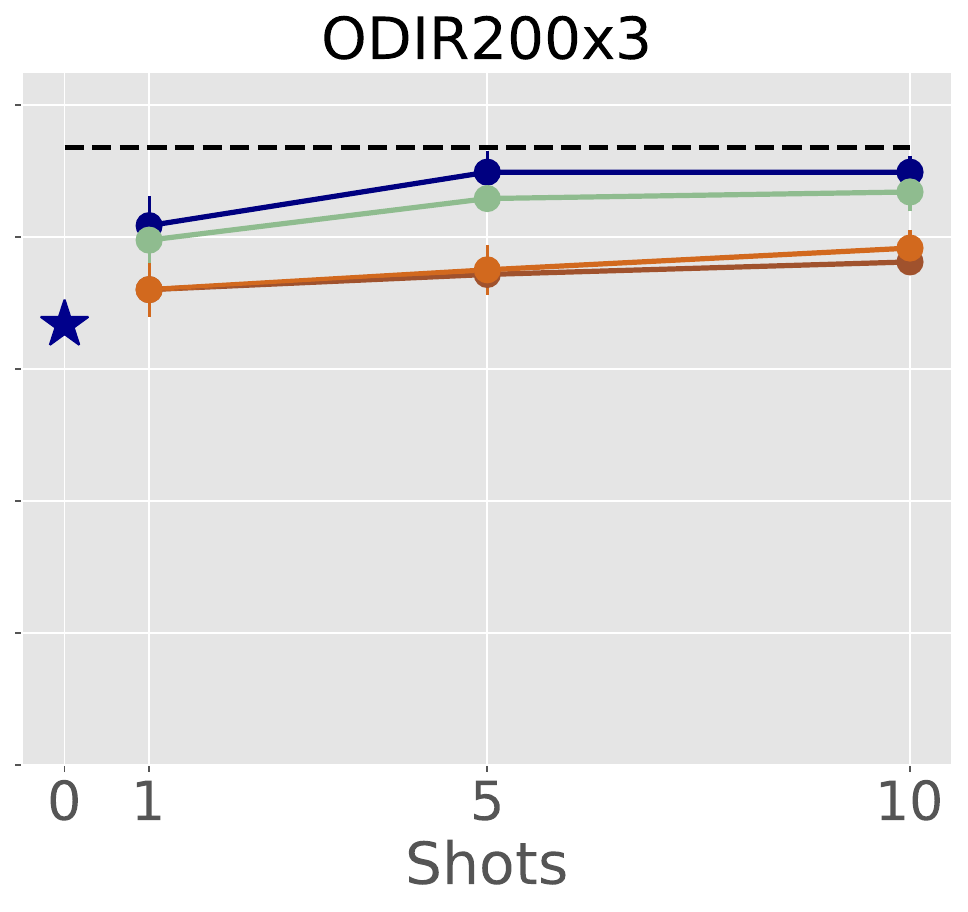}}
          \subfloat{\includegraphics[height=0.15\linewidth]{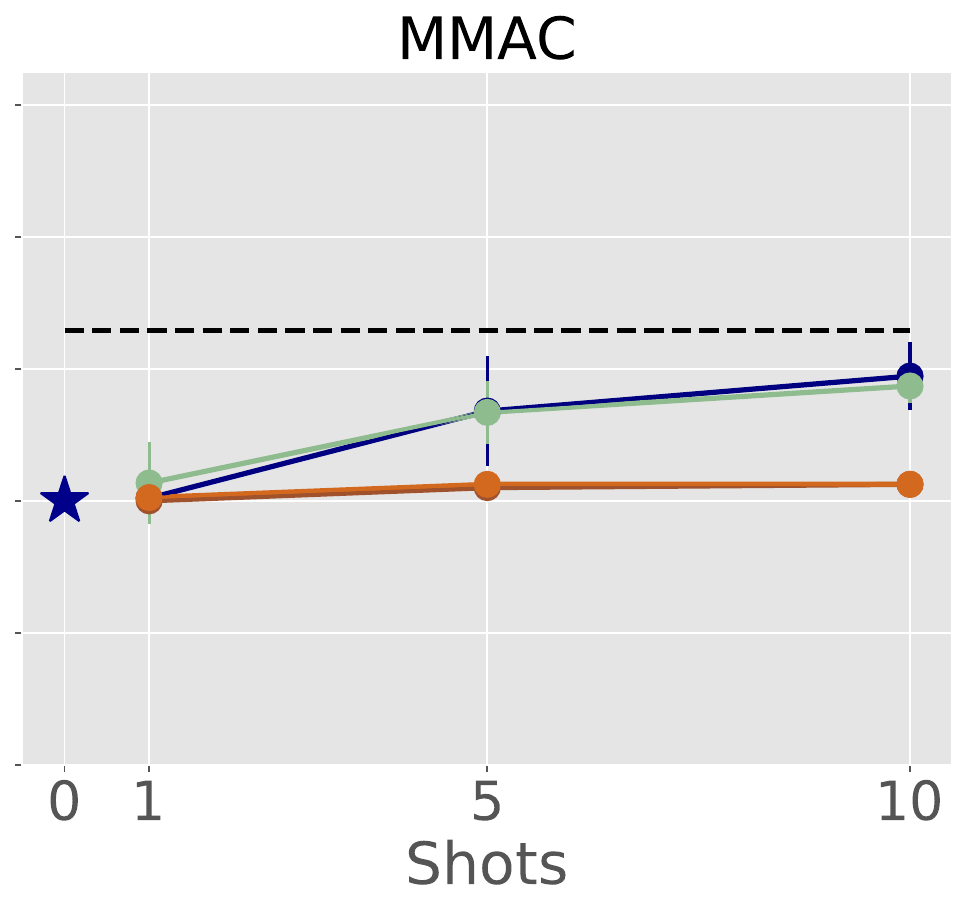}}
        
        \caption{\textbf{Vision-language few-shot Adapters.} The results of different Adapters in the few-shot setting. The metric presented is the average accuracy, averaged across 5 cross-validation folds. ZS: zero-shot (\textit{i.e.}, prompt-based classification with domain-knowledge prompts).}
        \label{fig:adapters}
    \end{center}
\vspace{-5mm}
\end{figure*}

The results for transferability through image features adaptation on the evaluation datasets are presented in Figure \ref{fig:transferability}. 

\textit{\textbf{Task-specific models (TSMs)}.} The obtained results unveil the limitations of task-specific models regarding transferability. While they perform well for the task they have been trained on, with enough pre-training data (see Figure \ref{fig:transferability}: TSM$_{DR}$ on MESSIDOR or TSM$_{Glaucoma}$ on REFUGE), their performance degrades in more challenging scenarios. Concretely, this occurs when \textit{i)}: the pre-training dataset is relatively smaller (see Figure \ref{fig:transferability}: TSM$_{Diseases}$ on FIVES), and \textit{ii)}: the models are adapted to unseen tasks (see Figure \ref{fig:transferability}: TSM$_{Glaucoma}$ on MESSIDOR, or TSM$_{DR}$ on 20x3). 

\textit{\textbf{Dataset-specific models (Fine-tuning)}.} Also, from Figure \ref{fig:transferability}, one may observe the limitations of fully-training dataset-specific models. This strategy performs poorly in the low-data regime (see Figure \ref{fig:transferability}: all datasets). Even in the larger data regime, the available samples might not be enough to reach the performance obtained by FLAIR (see Figure \ref{fig:transferability}: Fine-tuning on MESSIDOR). In addition, they are computationally expensive, since they require tuning the whole model. 

\textit{\textbf{Proposed foundation model (FLAIR)}.} The main takeaways on the performance of LP adaptation using FLAIR pre-training on the assembly dataset are: \textit{i)} It performs on par with the best task-specific models under domain shift (see Figure \ref{fig:transferability}: MESSIDOR dataset); \textit{ii)} It outperforms by a large margin these models on tasks under-represented in the assembly dataset (see Figure \ref{fig:transferability}: FIVES dataset); \textit{iii)} On unseen categories, it outperforms adaptation of the fully fine-tuned dataset-specific models in the low data regime (see Figure \ref{fig:transferability}: 20x3, ODIR200x3, and MMAC datasets); and \textit{iv)} In many cases, it also outperforms these fully-tuned models in the large data regime (see Figure \ref{fig:transferability}: MESSIDOR, REFUGE and 20x3 datasets), which alings with relevant recent literature on vision-language pre-training for radiology imaging \citep{Tiu2022}. 

\subsubsection{Vision-language Adapters}
\label{sec:visionLanguage}

Recent emergent literature in computer vision has investigated strategies, often referred to as Adapters, to fine-tune vision-language models in low-data (few-shot) regimes for the target tasks, e.g., Clip-Adapter \citep{clipAdapter} and Tip-Adapter \citep{tipAdapter}. These strategies typically integrate the knowledge driven from the pre-trained language encoder along with the vision features and use additional layers in the networks. Still, the utility of these adapters remains largely unexplored in the medical domain. Figure \ref{fig:adapters} depicts the results obtained by different vision-language Adapters using our pre-trained FLAIR foundation model and expert-knowledge prompts, across the different tasks. The results point to the powerful capabilities of zero-shot classification in different scenarios. In most of the cases, zero-shot inference, enhanced with domain-expert knowledge prompts, outperforms adaptation using $k\leq5$ shots (see Figure \ref{fig:adapters} MESSIDOR, FIVES, REFUGE, 20x3). As for the vision-language Adapters \citep{tipAdapter,clipAdapter}, these do not seem to provide consistent improvements, neither over zero-shot classification (when $k\leq5$) nor over basic Linear Probing (when $k=10$).

\subsubsection{Ablation experiments}
\label{ablation_experiments}

In this section, we present ablation experiments that motivate different decisions in the design of the proposed framework. 

\paragraph{\textbf{What features to use for knowledge transfer}} Vision-language pre-training models align the image-encoder features, $\theta_f(\cdot)$, to the text representations via a projection, $\theta_p(\cdot)$, along with a mapping to the unit hyper-sphere using an l2 normalization. Regarding the transferability of the pre-trained visual features to downstream domains and tasks via linear probing (LP), the standard feature-representation choice in prior literature is often based on both projection and normalization \citep{Radford2021, clipAdapter, tipAdapter}. In the following ablation experiment, we evaluate the feature transferability for the different evaluation datasets using the following three options: vision, projected, and projected-and-normalized features. We evaluated the three options under both the low and large-data regimes, using $k=10$ and $80\%$ of the dataset for training. 

\begin{figure}[h!]
    \begin{center}

         \subfloat{\includegraphics[width=0.6\linewidth]{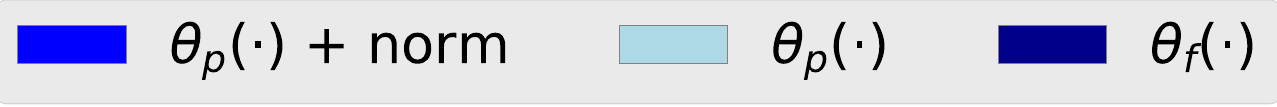}}  
          
          \subfloat{\includegraphics[width=0.50\linewidth]{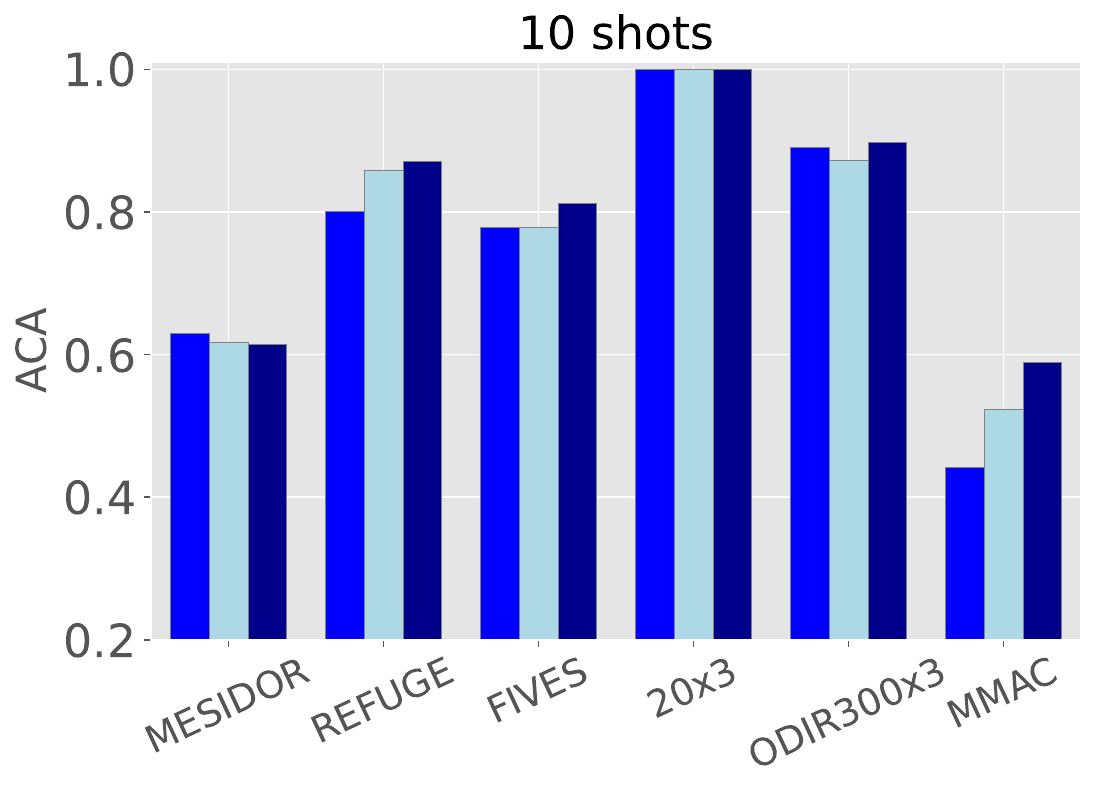}}
          \subfloat{\includegraphics[width=0.50\linewidth]{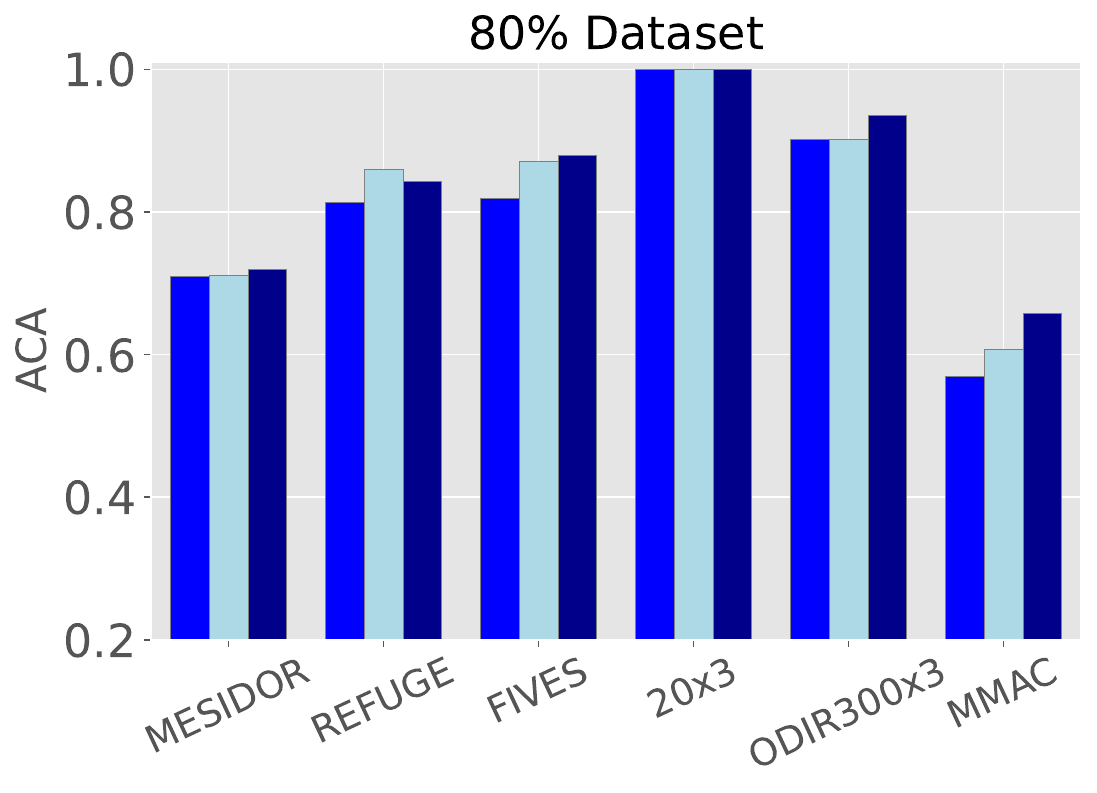}}

        \caption{\textbf{Study of the transferred features for adaptation.} Evaluation of the performance of the linear-probe transferability of the features extracted from the vision encoder, $\theta_f(\cdot)$, the inter-modality projection head, $\theta_p(\cdot)$, and its hypersphere normalization,  $\theta_p(\cdot)$ + norm. The metric presented is the average accuracy, averaged across 5 cross-validation folds. The results are presented for the low-data (10 shots) and large-data ($80\%$ of the whole dataset) regimes.}
        \label{fig:projection}
    \end{center}
\end{figure}

Figure \ref{fig:projection} depicts the results, which show performance improvements across most of the tasks when using visual representation $\theta_f(\cdot)$ for transferability, in comparison to using projected features $\theta_p(\cdot)$ or projected-and-normalized features $\theta_p(\cdot)+\mbox{norm}$. Motivated by these observations, we selected original feature representation $\theta_f(\cdot)$ for the transferability experiments in this work.

\paragraph{\textbf{Generalization of linear-probe adaptation under domain shifts}}

The \textit{pre-train-and-adapt} strategy using image-language models and computationally efficient linear-probe adaptation has shown promising performances on downstream computer-vision tasks. In the following, we aim at conducting a more comprehensive evaluation of this linear-probe strategy, in order to assess the capacity of the adaption stage in response to new changes in a target domain (\textit{i.e.}, there are domain shifts after adaptation). To do so, we employ the supplementary evaluation subsets; see Table \ref{datasets_validation_supplemental}. In particular, we evaluate the performance of the linear probe, which has been fine-tuned on a source domain, in a novel target domain. More concretely, the adaptation is performed as follows using two datasets A and B: The model is fine-tuned on A and tested on B, and vice-versa. Again, two feature representations are evaluated for transferability: features extracted from the vision encoder, $\theta_f(\cdot)$, and features based on the inter-modality projection head, $\theta_p(\cdot)$. We juxtapose the performance of the linear probe to fine-tuning all the model trainable parameters on the source data (\textit{i.e.}, using a standard supervised-learning setting, but with parameter initialization using either FLAIR or Imagenet model), as well as to the zero-shot performance. The experiments are carried out in the large-data regime, to evaluate the best-case scenario, in which the available data is not a limiting factor.

\begin{figure}[h!]
    \begin{center}

         \subfloat{\includegraphics[width=0.6\linewidth]{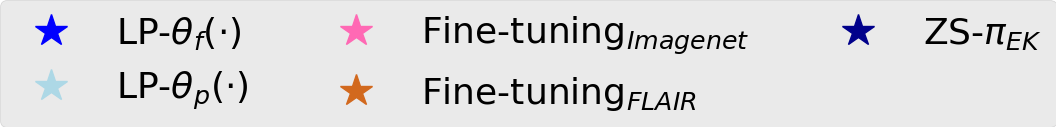}}

         \subfloat{\includegraphics[height=0.35\linewidth]{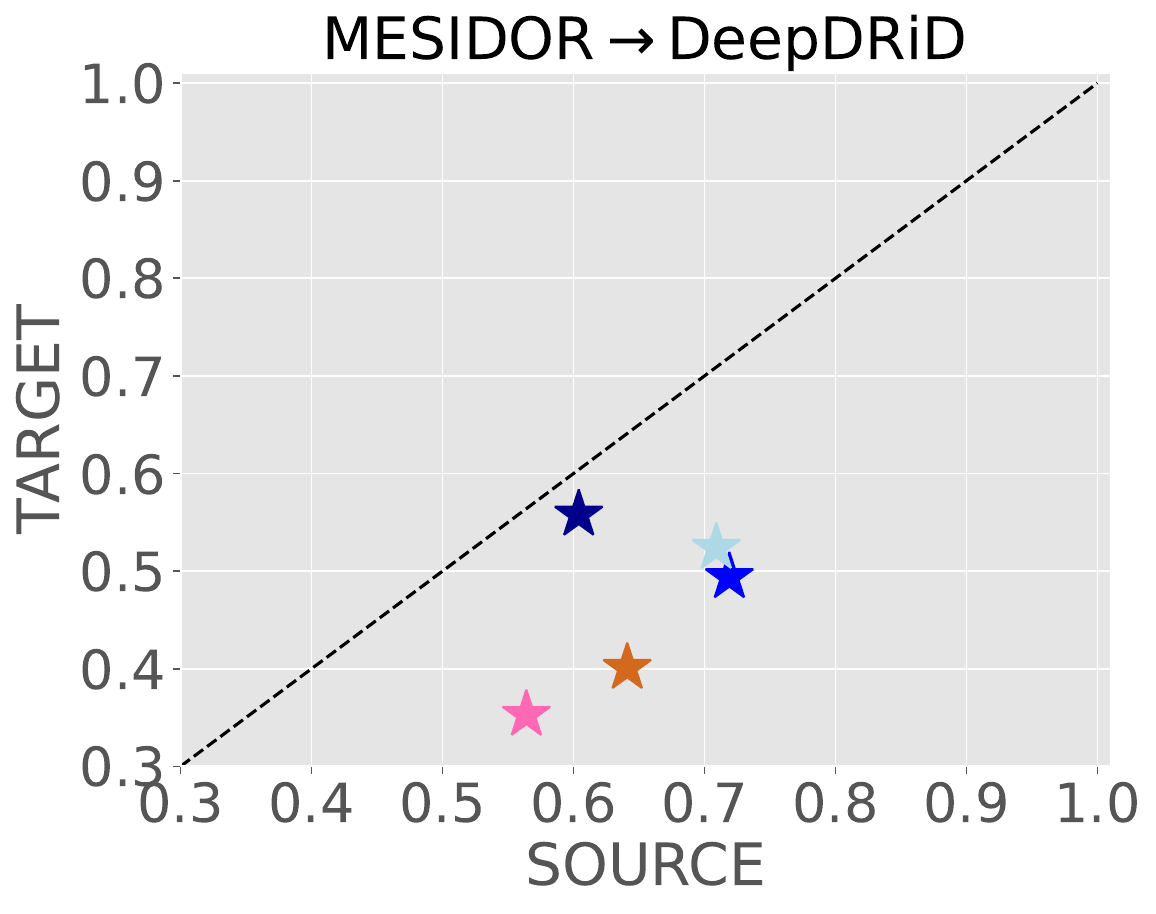}}
         \subfloat{\includegraphics[height=0.35\linewidth]{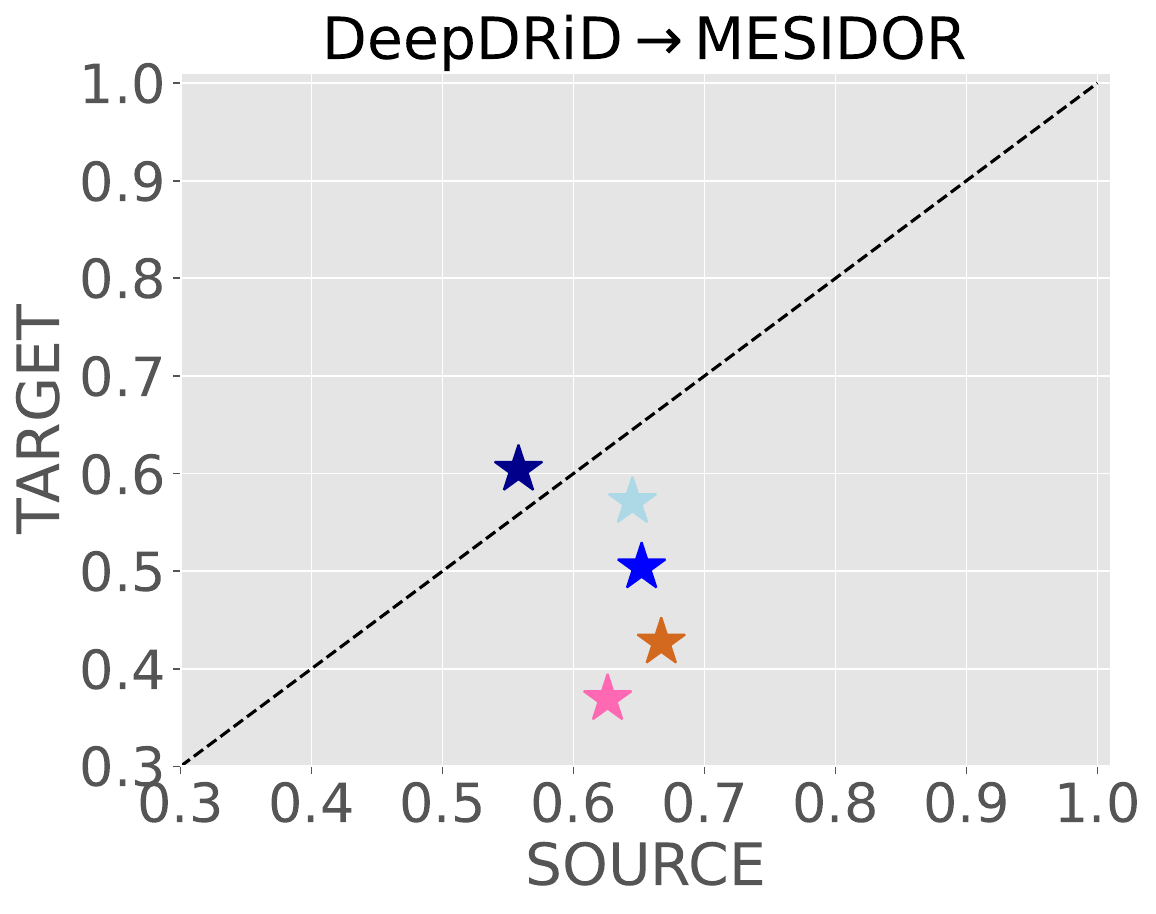}}

         \footnotesize{\textit{Domain shift - DR grading}}

         \subfloat{\includegraphics[height=0.35\linewidth]{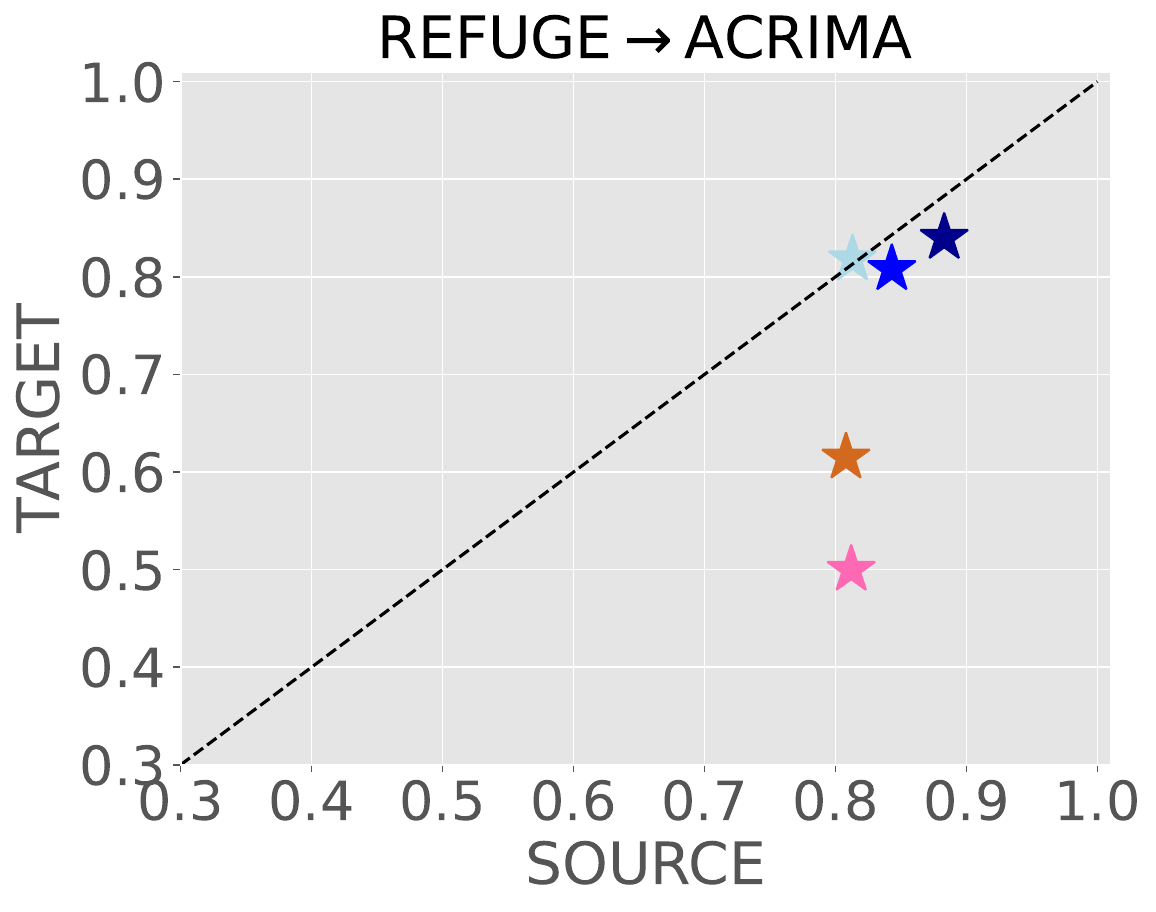}}
          \subfloat{\includegraphics[height=0.35\linewidth]{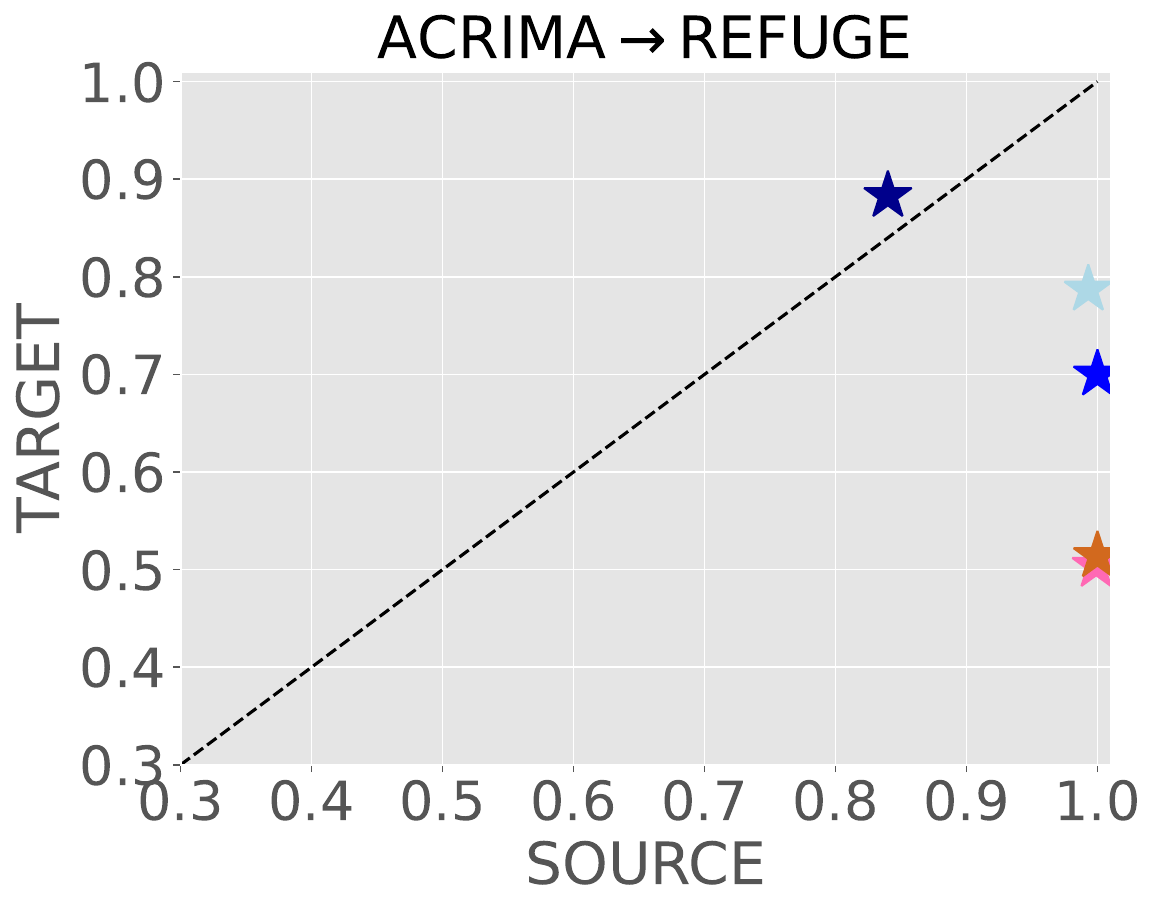}}

         \footnotesize{\textit{Domain shift - Glaucoma}}

          \subfloat{\includegraphics[height=0.35\linewidth]{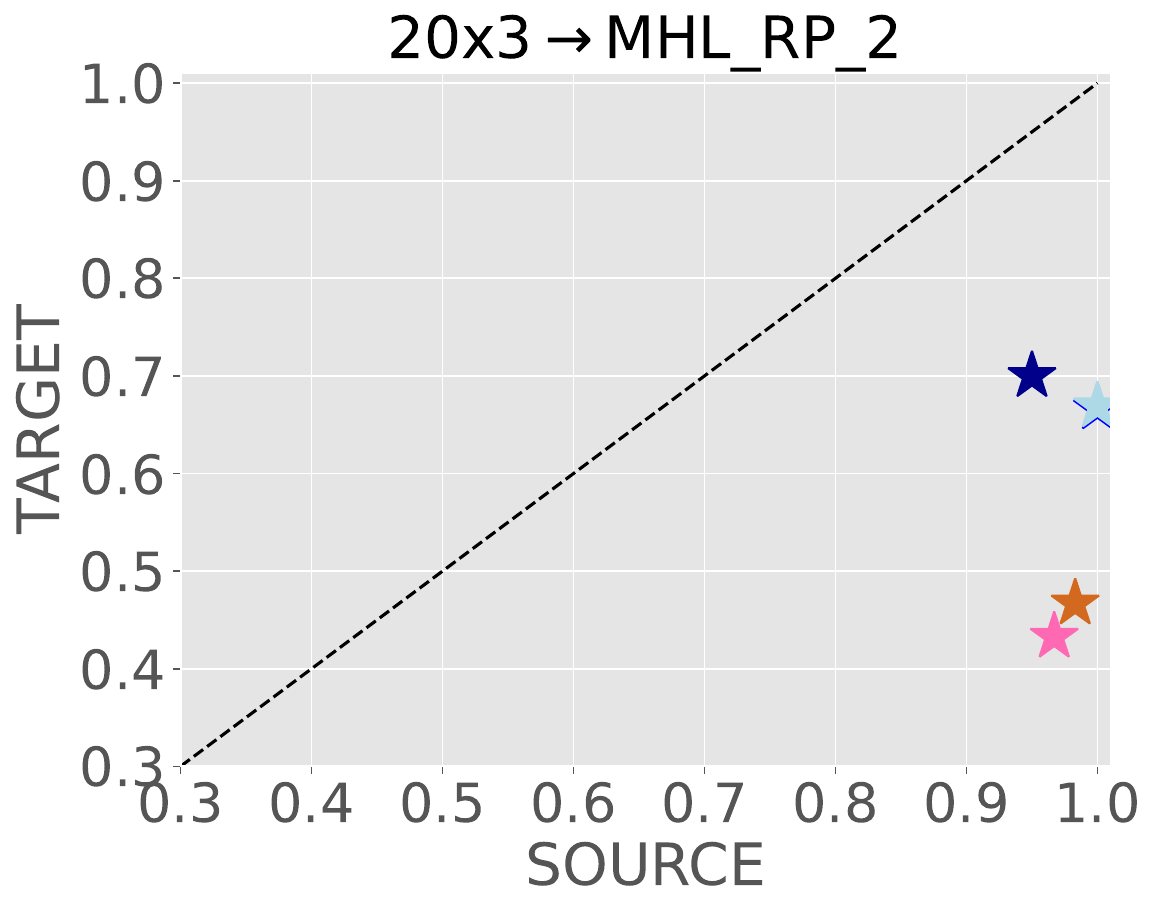}}
          \subfloat{\includegraphics[height=0.35\linewidth]{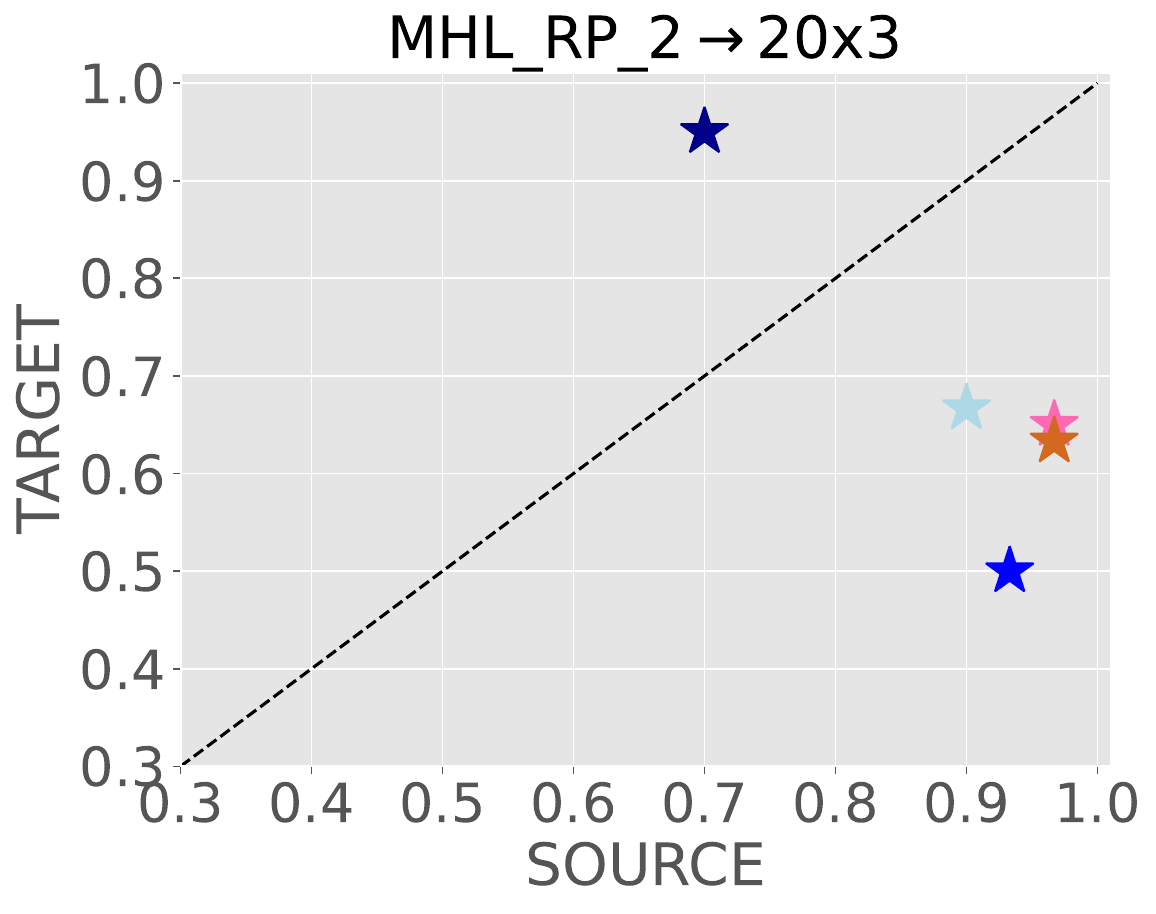}}

         \footnotesize{\textit{Unseen categories - N, RP, MHL}}

          \subfloat{\includegraphics[height=0.35\linewidth]{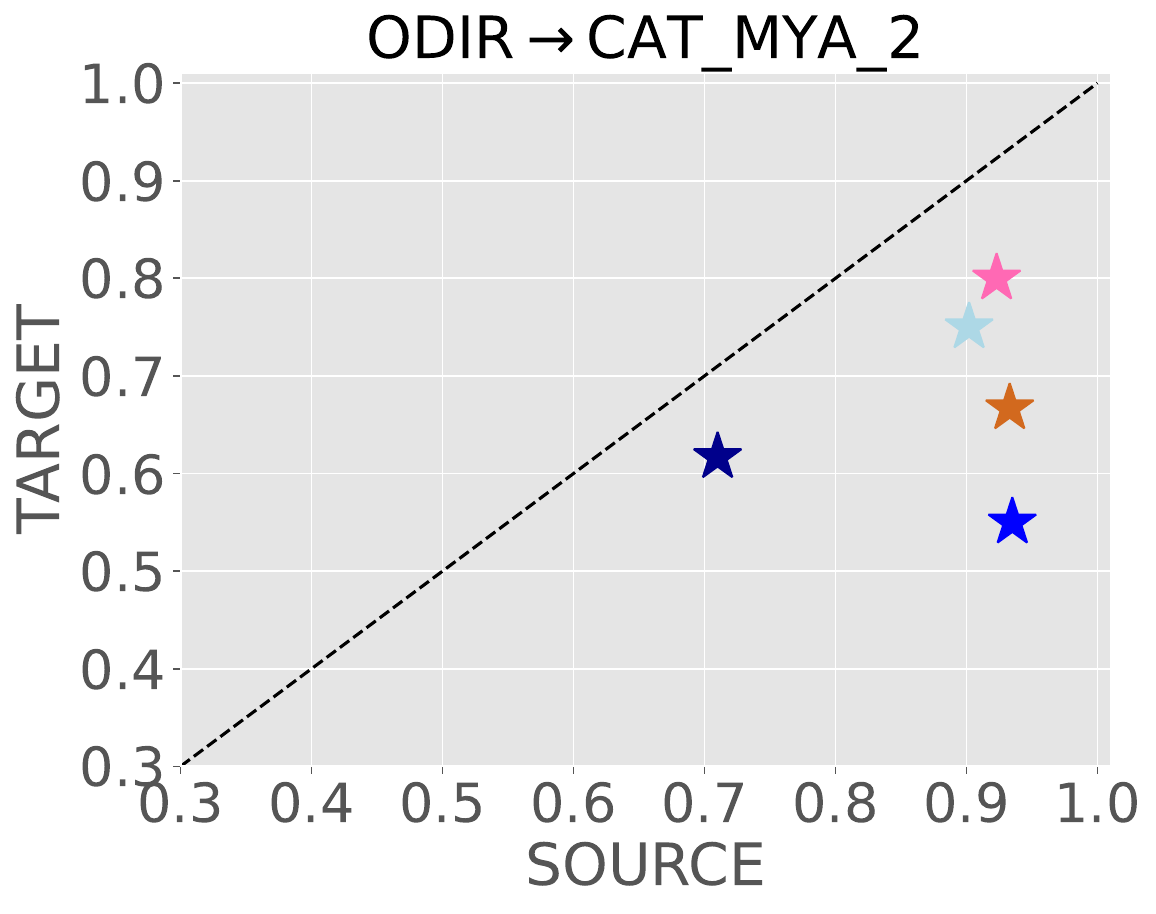}}
          \subfloat{\includegraphics[height=0.35\linewidth]{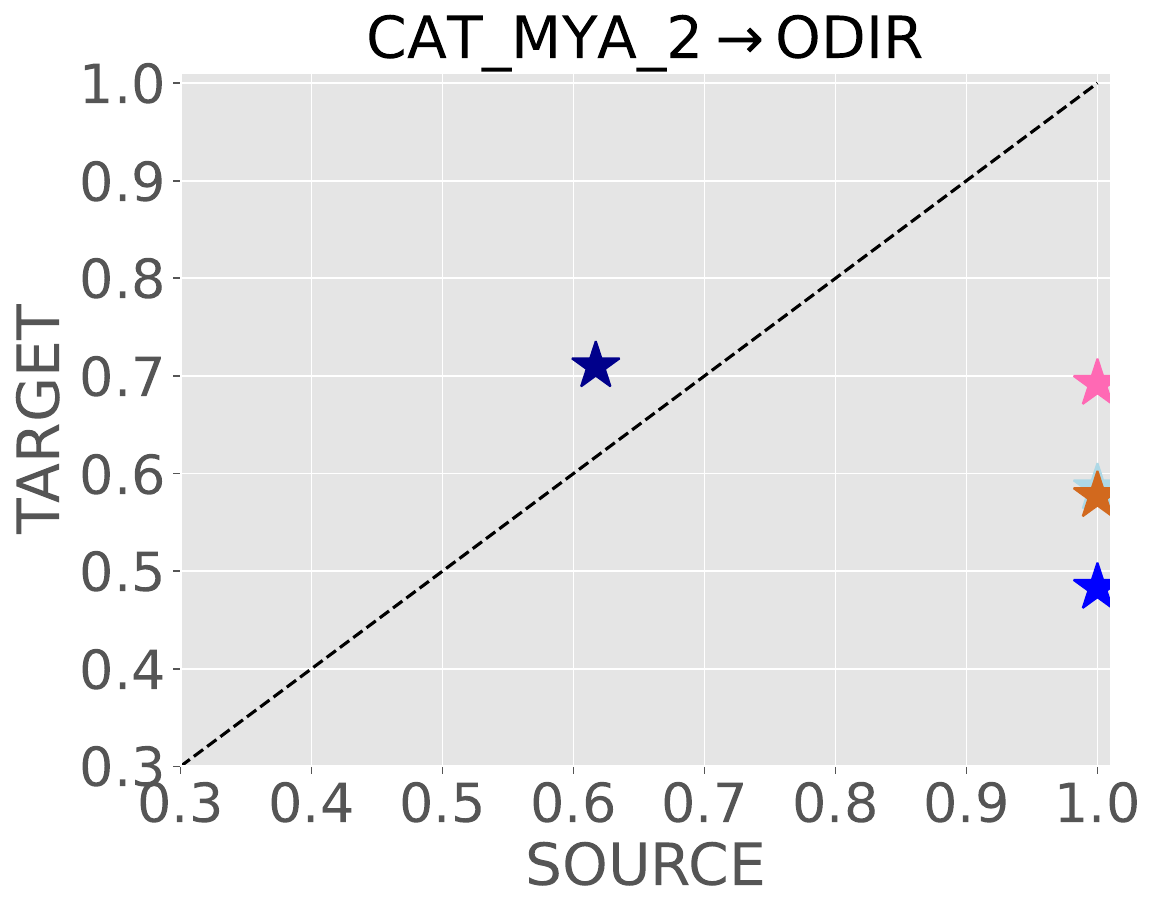}}

         \footnotesize{\textit{Unseen categories - N, CAT, MYA}}
         
        \caption{\textbf{Generalization of the transferred features after the adaptation stage} Evaluation of the linear-probe generalization with respect to the transferred features. Adapters are tuned on the source domain, and the performance is evaluated on another dataset with the same categories. The metric presented is the average accuracy, averaged across 5 cross-validation folds. The results were obtained under the large-data regime, with $80\%$ of the training data.}
        \label{fig:adap_gen_projection}
    \end{center}
\end{figure}

Figure \ref{fig:adap_gen_projection} depicts the results from these experiments, which point to the following takeaways: The Fine-tuned, dataset-specific models, which update all the trainable parameters using the source data, reach good performance on the source domain. However, they struggle to generalize under domain shifts; see Figure \ref{fig:adap_gen_projection}, \textit{first}, and \textit{second rows}. Linear probing (LP) from the foundation model mitigates this difficulty in several cases; see Figure \ref{fig:adap_gen_projection}, \textit{first, second}, and \textit{third rows}. These observations emphasize the generalization capabilities of the foundation model and are in line with recent observations in the computer vision community \citep{Kumar22}. This is especially the case when adapting the feature representation of the multi-modal projection, $\theta_p(\cdot)$. As echoed earlier in Figure \ref{fig:projection}, the vision-encoder features, $\theta_f(\cdot)$, seem more specialized for the source domain and, hence, yielded the best overall performance. However, the experimental results in Figure \ref{fig:adap_gen_projection} suggest that the performances yielded by these features could be affected under domain shifts; see Figure \ref{fig:adap_gen_projection}, \textit{all}. Thus, what representation to use for adaptation might depend on the expected data variability and the consistency between the source and target domains over time. It is worth mentioning that, when a large domain shift is expected, the prompt-based (zero-shot) classification might result in a robust solution. In many scenarios, the performance of LP on a target domain is below the prompt-based classification. In the case of known categories by the foundation model, prompt-driven classification in a zero-shot fashion achieves even more promising results for different domains; see Figure \ref{fig:adap_gen_projection}, \textit{first} and \textit{second rows}. We provide additional results in Section \ref{sec:supTranf}, which illustrate the robustness of text-driven transferability to domain shifts and are in line with recent observations in the computer-vision literature \citep{wortsman2021robust, Goyal23}. Of course, and as echoed in Figure \ref{fig:adapters}, this robustness of zero-shot classification to domain variability comes at the price of a source-domain performance that is lower than LP.

\subsection{Comparison with existing retina foundation models}

Recent works have explored the potential of foundation models for color fundus retina imaging, by leveraging large unlabeled datasets and self-supervised learning objectives. Firstly, REMEDIS \citep{remedis} followed SimCLR \citep{Chen2020} to exploit a large, private dataset of  2.2M images, using different ResNet backbones. Concurrently to our work, RETFound \citep{zhou2023foundation} scaled up the network size using a ViT-B/16 \citep{vit} as the backbone for training a Masked Autoencoder (MAE) \citep{mae} with a private dataset of nearly 800K images. In the following, we compare the transferability of their pre-trained representations to our proposed expert-knowledge-driven foundation model, FLAIR, using Linear Probing. We employed the RETFound encoder with its open-access available weights\footnote{\url{https://github.com/rmaphoh/RETFound_MAE}}. Regarding REMEDIS, its pre-trained weights for fundus imaging are not publicly available. However, since we have explored SimCLR pre-training in this work (see Section \ref{sec:lp}), we resort to this model to approximate the comparison.

\begin{table}[h!]
\setlength{\tabcolsep}{3pt}
\centering
\caption{\textbf{Comparison with SoTA retina foundation models.} Linear-probing transferability of the proposed vision-language foundation model, FLAIR, compared to the self-supervised pre-trained model RETFound, based on Masked Autoencoders (MAE), and other unsupervised strategies. The metric presented is ACA. The best average results are highlighted in bold.}
\label{retfound_lp}
\scriptsize
\begin{tabular}{lccccccc}
\hline
\multicolumn{1}{c}{Method} & \multicolumn{7}{c}{Datasets} \\ \cline{2-8}
                                             & MESS & FIVES & REFUGE & 20x3 & ODIR   & MMAC$^{\ref{note_mmac}}$ & \textbf{Avg.}      \\ \hline 
\multicolumn{8}{l}{\textit{RETFound setting - i.e. ViT-B/16 + private dataset.}} \\ \hdashline 
MAE                                          & 0.457 & 0.765 & 0.747 & 0.950 & 0.887 & 0.512 & 0.719      \\ \hline
\multicolumn{8}{l}{\textit{Ours - i.e. RN50 + open-acces dataset assembly.}} \\ \hdashline 
ImageNet                                     & 0.424 & 0.741 & 0.733 & 0.983 & 0.887 & 0.597 & 0.727  \\
SimCLR                                       & 0.422 & 0.726 & 0.701 & 0.900 & 0.902 & 0.475 & 0.687  \\
\cellcolor{Gray}FLAIR    & \cellcolor{Gray}0.719 & \cellcolor{Gray}0.879 & \cellcolor{Gray}0.843 & \cellcolor{Gray}1.000  & \cellcolor{Gray}0.935  & \cellcolor{Gray}0.658 & \cellcolor{Gray}\textbf{0.839} \\ \hline

\end{tabular}
\end{table}

\begin{figure}[h!]
    \begin{center}
    
         \hspace{-2mm}\subfloat{\includegraphics[width=1\linewidth]{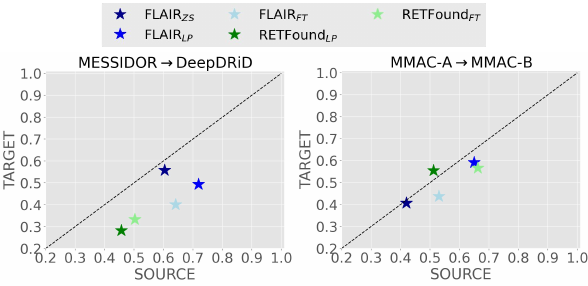}}  

        \caption{\textbf{Fine-tuning self-supervised foundation models $^{\ref{note_mmac}}$.} Effect of fine-tuning (FT) self-supervised pre-trained models, such as RETFound, compared with the proposed vision-language model, FLAIR. Experiments are carried out for the large data regime. Adapted models are evaluated on the source domain and out-of-distribution (target) domains. ZS: zero-shot; LP: Linear Probe.}
        \label{fig:retfounf_ft}
    \end{center}
    \vspace{-2mm}
\end{figure}

Results in Table \ref{retfound_lp} (see note in \footnote{\label{note_mmac}The results on MMAC dataset in Table \ref{retfound_lp} and Figure \ref{fig:retfounf_ft} differ slightly from the ones in the published version of the manuscript. In the published version, we wrongly reported results from the cross-validation set, instead of using the MMAC challenge's validation subset for testing.}) showcase the limitations of self-supervised pre-training for the efficient transferability of its learned representations. Albeit pre-trained on fewer data and using a smaller network architecture, FLAIR performs significantly better than RETFound (+12$\%$). Interestingly, Linear Probing from the basic ImageNet pre-trained RN50 model also outperforms RETFound and SimCLR pre-training. This suggests that self-supervised pre-trained models for color fundus images require more severe fine-tuning for proper adaptation. Note that similar trends have been recently exposed in other medical image modalities, such as histology \citep{ft_histo}. It is fair to acknowledge that authors from REMEDIS and RETFound explicitly recommend fully fine-tuning their pre-trained models for proper adaptation. Nevertheless, such a strategy might undermine the potential of foundation models in medical imaging, neglecting the data- and resource-efficient adaptation to challenging clinical contexts. In addition, fine-tuning might distort the rich feature representations learned during the foundation model pre-training \citep{wortsman2021robust} and fail to properly generalize to unseen domains. As a sample, we extend the studies on domain generalization (see Section \ref{ablation_experiments}, Fig. \ref{fig:adap_gen_projection}) for adapting RETFound to diabetic retinopathy and myopic maculopathy grading, on the favorable large data regime. Results in Fig. \ref{fig:retfounf_ft} demonstrate, that even in this scenario, fine-tuning larger-scale self-supervised pre-trained models does not neccesarily offer better generalization than Linear Probing expert-knowledge supervised pre-trained models.

Thus, even though self-supervised learning has served as a popular lifeguard for pre-training modality-specific models in the absence of labeled data, our observations indicate that leveraging expert knowledge is a more promising venue in the context of color fundus images. It is worth mentioning that these observations are aligned with recent relevant analysis of supervised pre-training in computer vision \citep{mlpproj,rethinking,noreasonnosup}, which indicate the benefits of supervised pre-trained models.


\section{Discussion}
\label{sec:conclusions}

We introduced FLAIR, a novel vision-language foundation model for universal pathology detection and classification in retinal fundus images. Encoding expert's domain knowledge in the form of text-prompt supervision, FLAIR is trained on an assembly of $38$ publicly available, mostly categorical datasets, containing up to $101$ different target categories. By leveraging domain knowledge, we mitigated the scarcity of text-based supervision in retinal fundus imaging datasets, opening a promising avenue toward vision-language pre-training in this domain. Specifically, we enhanced the categorical information in the datasets by text-based encoding of the major features of the pathologies as well as the hierarchies and relationships between them. Such valuable expert knowledge could be extracted from the relevant clinical literature and community standards.

We have empirically evaluated FLAIR capabilities for generalization and transferability, under scenarios with domain shifts and for unseen diseases. The proposed model shows a strong generalization, driven by domain knowledge prompts for zero-shot prediction. This is especially the case when tested on novel pathologies, in which basic categorical text prompts are uninformative. Additionally, with a lightweight, linear-probe adaptation, FLAIR outperforms fully fine-tuned, dataset-specific models on the target domains and tasks. The difference is even more pronounced under the low-data (few-shot) regime. 

We also conducted comprehensive ablation studies, which show the substantial effect of integrating domain knowledge during both vision-language pre-training and zero-shot predictions. Our FLAIR model, along with domain-knowledge prompts for zero-shot prediction, outperformed significantly Contrastive Language-Image Pre-training (CLIP) on general computer vision data. Furthermore, it bypasses by margins a version of CLIP trained on the same retinal-imaging data but with naive categorical text prompts. Our results point to the potential of embedding domain-specific, expert knowledge in building vision-language foundation models targeted at different sub-domains of medical imaging, even beyond the fundus application domain tackled in this study. We showed that, even in the absence of large datasets with text-based supervision, samples with categorical labels could still be exploited to train powerful vision-language representations, by encoding expert's domain knowledge into text supervision. The potential of this strategy is even more evident when FLAIR is compared with larger-scale, recently proposed retina foundation models, such as RETFound, which do not exploit any supervisory signal. Our experiments indicate the importance of expert supervision, ahead of scaling the database used, or network parameters. 

Finally, we have provided in-depth experiments to deepen our understanding of the potential and limitations of the proposed methodology. Firstly, it is important to note that FLAIR benefits from assembling a dataset comprising multiple data sources, which potentially involves heterogeneous annotation protocols and grading systems, as well as intra- and inter-annotator variability. Secondly, we have evaluated the zero-shot and adaptation performance on supplementary datasets, through different ablation experiments. This has revealed certain limitations of the current vision-language pre-training paradigm in the medical field. In particular, the zero-shot generalization is sensitive to the designed prompts when tested on novel categories. While promising, the results still show certain variability, which is aligned with recent observations in other works \citep{Wang2022}. Finally, although the foundation model has shown promising adaptation to new tasks with little resources, it may have difficulties generalizing to out-of-distribution data after adaptation, an observation that has also been stressed in recent vision-language literature \citep{wortsman2021robust, Goyal23}. 

The encountered limitations might inspire further research directions to improve the performances of vision-language foundation models. Developing new reliable tools for language processing on expert domains, such as fundus imaging diagnosis, is an appealing future direction, which may improve the robustness of the text encoders. Finally, integrating novel Adapters, able to generalize well to out-of-distribution data, might improve the performances of this emerging pre-train-and-adapt paradigm. For this challenge, text-driven Adapters might be an interesting research avenue for the future.

\section*{Acknowledgments}

The work of J. Silva-Rodríguez was partially funded by the \textit{Fonds de recherche du Québec (FRQ)} under the Postdoctoral Merit Scholarship for Foreign Students (PBEEE). The work is supported, in part, by PROMPT Quebec, via its PARTNERSHIP-AI program. We also thank Calcul Québec and Compute Canada. 

\bibliographystyle{model2-names.bst}\biboptions{authoryear}
\bibliography{refs}

\clearpage

\setcounter{section}{0}
\renewcommand{\thesection}{\Alph{section}}
\setcounter{table}{0}
\renewcommand{\thetable}{S\arabic{table}}
\setcounter{figure}{0}
\renewcommand{\thefigure}{S\arabic{figure}}

\begin{center}
\textbf{\normalsize Supplementary Materials. \\ A Foundation Language-Image Model of the Retina: Encoding Expert Knowledge in Text Supervision.}
\end{center}

\section{Dataset details}
\label{sec:supMaterials}

This section provides supplementary details regarding the assembly of datasets for the foundation fundus model training. Also, it includes supplementary datasets and partitions used during the evaluation stage, as well as visual descriptions of the categories selected to evaluate the proposed foundation model on novel categories.

\paragraph{\textbf{Categories}} In the following, we provide the categories and corresponding abbreviations used for training and testing the proposed foundation model. No diabetic retinopathy (noDR), mild diabetic retinopathy (mildDR), moderate diabetic retinopathy (modDR), severe diabetic retinopathy (sevDR), proliferative diabetic retinopathy (prolDR), noisy, clean, diabetic macular edema (DME), no referable diabetic macular edema (noDME), hard exudate (hEX), soft exudate (sEX), microaneurysms (MA), haemorrhages (HE), non clinically significant diabetic macular edema (nonCSDME), age-related macular degeneration (ARMD), media haze (MH), drusen (DN), pathologic myopia (MYA), branch retinal vein occlusion (BRVO), tessellation (TSLN), epiretinal membrane (ERM), laser scar (LS), macular scar (MS), central serous retinopathy (CSR), optic disc cupping (ODC), central retinal vein occlusion (CRVO), tortuous vessels (TV), asteroid hyalosis (AH), optic disc pallor (ODP), optic disc edema (ODE), shunt (ST), anterior ischemic optic neuropathy (AION), parafoveal telangiectasia (PT), retinal traction (RT), retinitis (RS), chorioretinitis (CRS), exudate (EX), retinal pigment epithelium changes (RPEC), macular hole (MHL), retinitis pigmentosa (RP), cotton wool spots (CWS), colobomas (CB), optic disc pit maculopathy (ODM), preretinal haemorrhage (PRH), myelinated nerve fibers (MNF), haemorrhagic retinopathy (HR), central retinal artery occlusion (CRAO), tilted disc (TD), cystoid macular edema (CME), post traumatic choroidal rupture (PTCR), choroidal folds (CF), vitreous haemorrhage (VH), macroaneurysm (MCA), vasculitis (VS), branch retinal artery occlusion (BRAO), plaque (PLQ), haemorrhagic pigment epithelial detachment (HPED), collateral (CL), normal (N), large optic cup (LOC), retina detachment (RD), Vogt-Koyanagi syndrome (VKH), maculopathy (M), Glaucoma (G), optic atrophy (OA), severe hypertensive retinopathy (sevHR), disc swelling and elevation (DSE), dragged disk (DD), congenital disk abnormality (CDA), Bietti crystalline dystrophy (BCD), peripheral retinal degeneration and break (PRDB), neoplasm (NP), yellow-white spots flecks (YWSF), fibrosis (F), silicon oil (SO), no proliferative diabetic retinopathy (noProlDR), no glaucoma (noG), cataract (CAT), hypertensive retinopathy (HR), neovascular age-related macular degeneration (neovARMD), geographical age-related macular degeneration (geoARMD), acute central serous retinopathy (acCSR), chronic central serous retinopathy (chCSR), no cataract (noCAT), abnormal optic disc (AOD), abnormal vessels (AV), abnormal macula (AM), macular edema (ME), scar (S), nevus (NE), increased cup disk (ICD), intraretinal microvascular abnormalities (IrMA), red small dots (ReSD), neovascularization (neoV), disease (Dis), superficial haemorraghe (supHE), deep haemorraghe (deepHE), myopic maculopahy (MM).

\begin{figure}[ht!]
\begin{center}
\includegraphics[width=.5\textwidth]{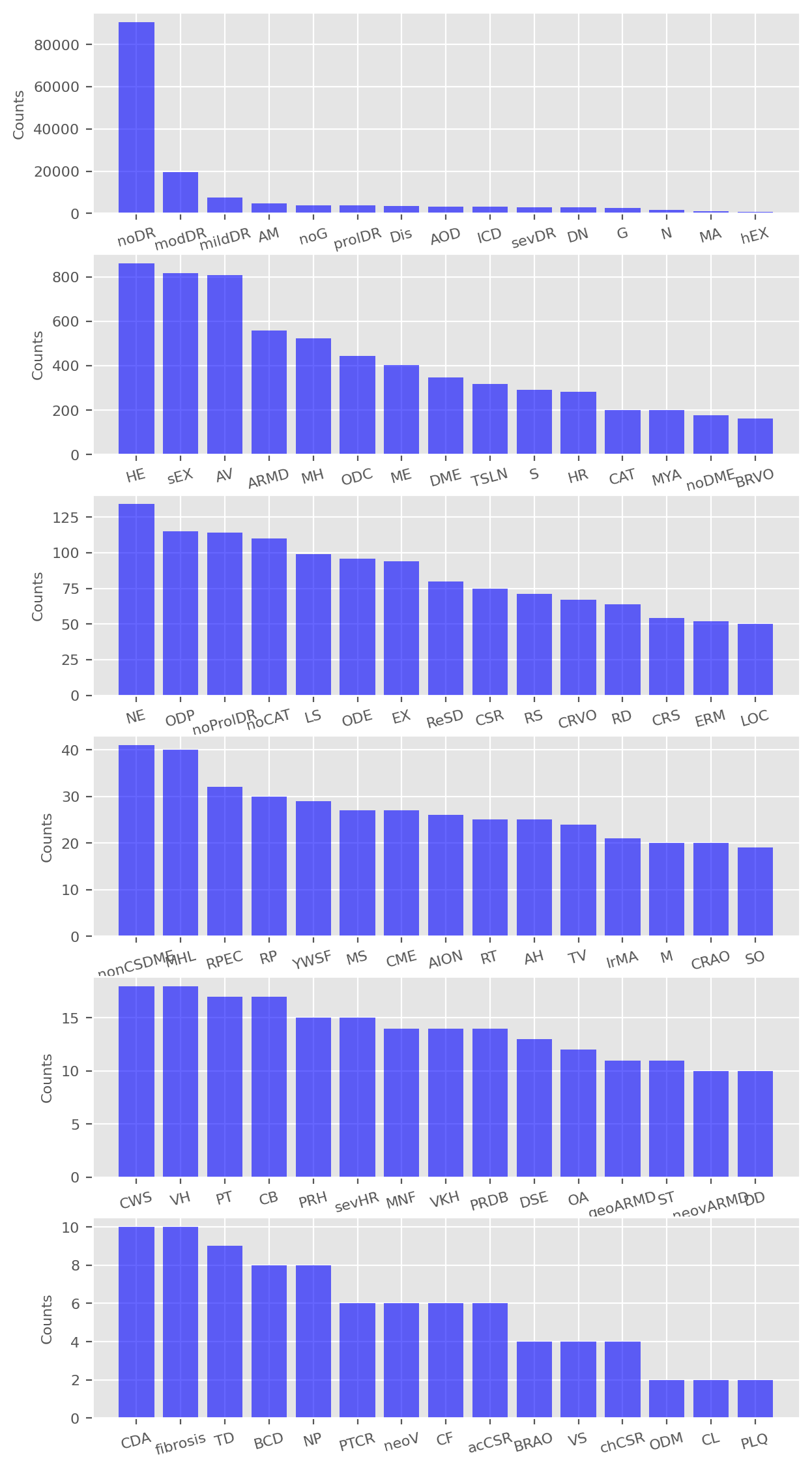}
\caption{\textbf{Distribution of the categories in the dataset assembly}. The plot depicts the number of samples within each category in the assembly dataset, ordered from the most represented class to the least. One may observe a long-tail distribution, which is common when integrating multiple public medical-imaging datasets \citep{Liu2023}.}
\label{fig:label_distribution}
\end{center}
\end{figure}

\paragraph{\textbf{Labels distribution}} We depict in Figure \ref{fig:label_distribution} the label distribution of the assembled dataset, used for training the universal fundus model.

\begin{figure}[ht!]
\begin{center}
\includegraphics[width=.5\textwidth]{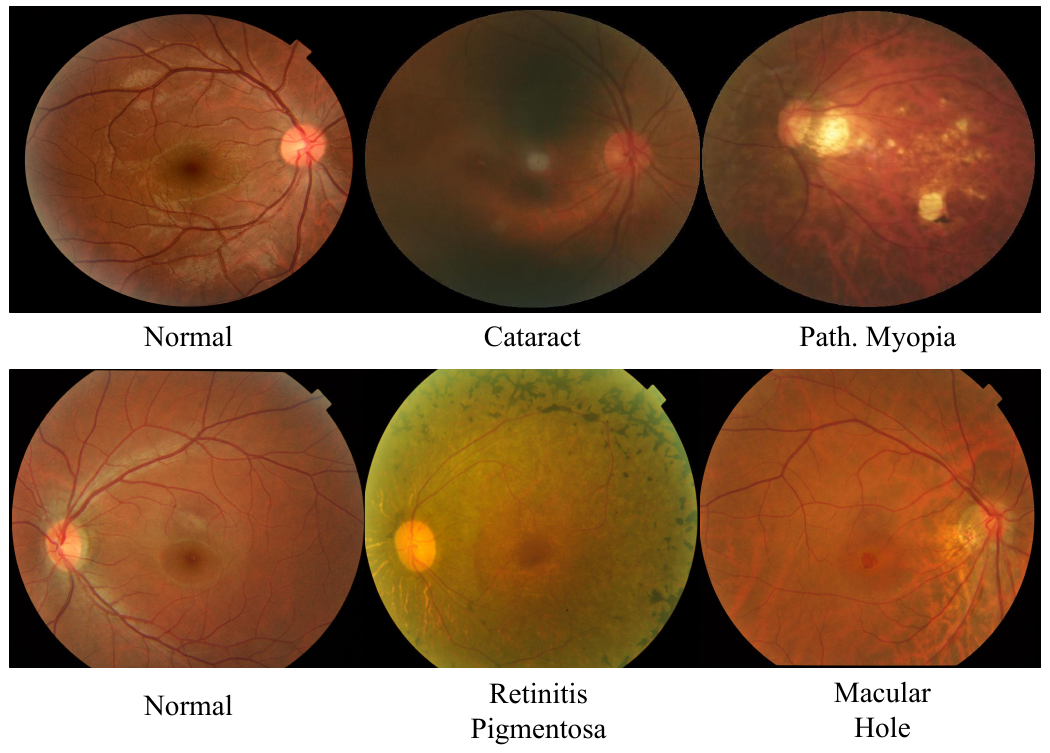}
\caption{\textbf{Novel categories for transferability evaluation.} Visualization of the different diseases used for validating the capabilities of the foundation model for adaptation to novel categories. Samples obtained from the ODIR-5K (\textit{top row}) and 1000x39 (\textit{bottom row}) datasets. Using expert's domain knowledge on retinal fundus image analysis, we define the following findings as descriptors of the different diseases: cataracts are featured `'\textit{opacities in the macular area}” \citep{cataract1, cataract2}, pathological myopia as `'\textit{anomalous disc, macular atrophy, and possible tessellation}” \citep{Medrano19}, retinitis pigmentosa is characterized for `'\textit{pigment deposits are present in the periphery}” \citep{retPig}, and a macular hole might be described as `'\textit{grayish fovea}”/`'\textit{lesion in the macula}” \citep{mhl}.}
\label{fig:incremental_images}
\end{center}
\end{figure}

\paragraph{\textbf{Unseen categories}} Figure \ref{fig:incremental_images} visualizes examples of the target categories selected to explore the transferability of FLAIR to unseen diseases. Also, we present domain-knowledge descriptors generated for prompt-based classification.

\paragraph{\textbf{Supplementary validation partitions}} As stated in the main manuscript, the categories cataract, pathological myopia, retinitis pigmentosa, and macular hole are used for evaluating the transferability of the universal model to downstream tasks. Thus, we proposed two different partitions by retrieving samples from specific datasets of the database assembly. First, ODIR200x3 contains 200 samples for normal fundus images and 200 diagnosed with cataract and pathological myopia, which are sampled from the ODIR-5K dataset. Second, 20x3 contains 20 normal examples and 20 examples of retinitis pigmentosa and macular hole, which are sampled from the 1000x39 dataset. The target diseases were not used during the training of the foundation models. Thus, we create supplementary evaluation datasets by using the discarded samples of the other datasets. First, we create CAT-MYA-2 partition by retrieving samples with cataract from the Cataract dataset and samples with pathological myopia from the BRSET dataset. Since this subset is designed to validate Adapters trained on the ODIR200x3 partition, we select normal samples from the other evaluation subset, 20x3. We followed the same procedure to create the evaluation partition RP-MHL-2, consisting of normal samples from ODIR200x3 and retinitis pigmentosa and macular hole samples from RFMid dataset. We present a summary of supplementary partitions in Table \ref{datasets_validation_supplemental}. It is worth mentioning that the ensuing partitions present a small number of samples, and are used only to evaluate the generalization capability of Adapters trained on new tasks, under the low-data regime. In addition, we also include the ACRIMA and DeepDRiD datasets for the additional validation of the proposed methods for generalization and adaptation to glaucoma detection and DR grading, respectively .

\begin{table}[h!]
\centering
\caption{\textbf{Supplementary datasets for the evaluation of the foundation model}. We set aside additional datasets for Glaucoma and DR grading evaluation, ACRIMA and DeepDRiD, respectively. Also, we combine samples corresponding to the categories selected as unseen for FLAIR training (\textit{i.e.}, RP, MHL, CAT, and MYA), which are discarded during training.
}
\label{datasets_validation_supplemental}
\scriptsize
\begin{tabular}{lrl}
\hline
Dataset                & \multicolumn{1}{l}{\#Images} & Labels                               \\ \hline
\textit{Domain shift}  &                              &                                      \\ \hdashline
DeepDRiD               & 2,256                            & noDR, mildDR, modDR, sevDR, prolDR.  \\
ACRIMA                 & 705                          & G, noG                               \\ \hline
\textit{Unseen categories}   &                              &                                      \\ \hdashline
RP-MHL-2               & 30                           & N, RP, MHL                           \\ 
CAT-MYA-2              & 60                           & N, CAT, MYA                          \\ \hline

\end{tabular}
\end{table}

\section{Detailed and additional results}
\label{sec:sup_results}

This section provides supplementary results to support the capabilities of the proposed universal model. In particular, we first present results on zero-shot classification for novel categories on the supplementary evaluation datasets. Secondly, we introduce the results obtained for the transferability of the pre-trained foundation model to additional datasets.

\subsection{Performance on novel classes}
\label{sec:supZS}

\paragraph{\textbf{Detailed numerical results}} We introduce in Table \ref{prompt_classification_detailed} disentangles results per class for the zero-shot transferability of FLAIR to unseen categories on 20x3 and ODIR200x3 datasets (Table \ref{prompt_classification} in the main manuscript). These results provide a better overview of whether the different vision-language models can differentiate among the target eye diseases.

\begin{table}[h!]
\setlength{\tabcolsep}{2.5pt}
\centering
\caption{\textbf{Disentanlged results for zero-shot generalization to unseen categories (zero-shot classification).} Transferability of the proposed foundation model to new tasks via prompt-based inference. We evaluated three different strategies for generating text prompts: anomaly detection (i.e., “\textit{normal}”/“\textit{disease}”), classification via naive prompt (i.e., the new disease name) and designed prompts (i.e., domain-knowledge descriptors). The metric presented is the accuracy for each category. The proposed method, FLAIR-$\pi_{EK}$, is shadowed, whereas the best results are highlighted in bold. This Table complements Table \ref{prompt_classification} in the main manuscript by providing per-class results.}
\label{prompt_classification_detailed}
\scriptsize
\begin{tabular}{lcccc|cccc}
\hline
Method   & \multicolumn{8}{c}{Dataset}                                  \\ \cline{2-9}
         & \multicolumn{4}{c}{20x3}    & \multicolumn{4}{c}{ODIR200x3}  \\ \cline{2-5} \cline{6-9}
         & N & RP & MHL & Avg. & N & CAT & MYA & Avg. \\ \hline
\multicolumn{8}{l}{\textit{Anomaly Detection Inference (i.e. "normal/disease")}}\\ \hdashline
CLIP                         & 1.000 & \multicolumn{2}{c}{0.200} & 0.600 & 0.770 & \multicolumn{2}{c}{0.412} & 0.591        \\ 
BiomedCLIP                   & 0.950 & \multicolumn{2}{c}{0.125} & 0.538 & 0.800 & \multicolumn{2}{c}{0.770} & \textbf{0.785}        \\
FLAIR-$\pi_{\textit{naive}}$                        & 0.900 & \multicolumn{2}{c}{0.200} & 0.550 & 1.000 & \multicolumn{2}{c}{0.102} & 0.551        \\
\cellcolor{Gray}FLAIR-$\pi_{\textit{EK}}$        & \cellcolor{Gray}0.850 & \multicolumn{2}{c}{\cellcolor{Gray}0.775} & \cellcolor{Gray}\textbf{0.812} & \cellcolor{Gray}0.985 & \multicolumn{2}{c}{\cellcolor{Gray}0.350} & \cellcolor{Gray}0.668        \\ \hline
\multicolumn{8}{l}{\textit{\textit{Inference with Naive Prompts - $\pi_{\textit{naive}}$  (e.g. "cataract")}}}\\ \hdashline
CLIP                        & 0.100 & 1.000 & 0.000 & 0.367 & 0.770 & 0.495 & 0.070 & 0.445        \\ 
BiomedCLIP                  & 0.900 & 0.950 & 0.400 & 0.750 & 0.765 & 0.920 & 0.495 & \textbf{0.727}        \\ 
FLAIR-$\pi_{\textit{naive}}$                        & 0.950 & 0.650 & 0.100 & 0.567 & 0.990 & 0.340 & 0.010 & 0.447        \\ 
\cellcolor{Gray}FLAIR-$\pi_{\textit{EK}}$        & \cellcolor{Gray}0.950 & \cellcolor{Gray}0.600 & \cellcolor{Gray}1.000 & \cellcolor{Gray}\textbf{0.850} & \cellcolor{Gray}0.990 & \cellcolor{Gray}0.455 & \cellcolor{Gray}0.005 & \cellcolor{Gray}0.483         \\ \hline
\multicolumn{8}{l}{\textit{\textit{Inference with Expert Knowledge Prompts - $\pi_{\textit{EK}}$}}}\\
\multicolumn{8}{l}{\textit{\textit{(e.g. "opacity in the macular area")}}}\\ \hdashline
CLIP                          & 1.000 & 0.000 & 0.000 & 0.333 & 0.290 & 0.195 & 0.955 & 0.480        \\
BiomedCLIP                    & 0.400 & 0.800 & 0.650 & 0.617 & 0.125 & 0.695 & 0.930 & 0.583        \\ 
FLAIR-$\pi_{\textit{naive}}$                        & 1.000 & 0.900 & 0.050 & 0.650 & 0.405 & 0.015 & 0.990 & 0.470        \\ 
\cellcolor{Gray}FLAIR-$\pi_{\textit{EK}}$        & \cellcolor{Gray}1.000 & \cellcolor{Gray}0.950 & \cellcolor{Gray}1.000 & \cellcolor{Gray}\textbf{0.983} & \cellcolor{Gray}0.760 & \cellcolor{Gray}0.765 & \cellcolor{Gray}0.475 & \cellcolor{Gray}\textbf{0.667}         \\ \hline

\end{tabular}

\end{table}
\vspace{-1mm}

\paragraph{\textbf{Zero-shot classification on supplementary datasets}} We present in Table \ref{prompt_classification_sup} the results obtained regarding prompt-based classification on unseen categories (\textit{i.e.} no adaptation) for the supplementary datasets RP-MHL-2 and CAT-MYA-2. The obtained results are in line with the ones observed in the main paper. Using domain-knowledge descriptors for prompt augmentation during training outperforms the base CLIP trained on medical images, and using well-designed prompts instead of target-category names for inference provides remarkable improvements, of $\sim 4\%$ and $\sim 11\%$, respectively. Also, in the CAT-MYA-2 dataset, the base CLIP model trained on natural images is able to obtain promising results - in contrast to the other three partitions used for zero-shot evaluation. It should be noted that CLIP is trained on 400M of images and text pairs, including some medical imaging datasets \citep{Radford2021}. 

\begin{table}[h!]
\setlength{\tabcolsep}{2.5pt}
\centering
\caption{\textbf{Zero-shot classification on supplementary datasets.} Transferability of the proposed foundation model to new tasks via prompt-based inference. Three different strategies to generate text prompts are evaluated: anomaly detection (\textit{i.e.}, `'\textit{normal}/\textit{disease}"), classification via naive prompt (\textit{i.e.}, the new disease name) or designed prompts (\textit{i.e.}, domain-knowledge descriptors). The metric presented is the accuracy for each category. The proposed method, FLAIR-$\pi_{EK}$, is shadowed, whereas the best results are highlighted in bold.}
\label{prompt_classification_sup}
\scriptsize
\begin{tabular}{lcccccccc}
\hline
Method   & \multicolumn{8}{c}{Dataset}                                  \\ \cline{2-9}
         & \multicolumn{4}{c}{RP-MHL-2}    & \multicolumn{4}{c}{CAT-MYA-2}  \\ \cline{2-5} \cline{6-9}
         & N & RP & MHL & Avg. & N & CAT & MYA & Avg. \\ \hline
\multicolumn{8}{l}{\textit{Anomaly Detection Inference (\textit{i.e.} "normal/disease")}}\\ \hdashline
CLIP                         & 0.800 & \multicolumn{2}{c}{0.050} & 0.425 & 1.000 & \multicolumn{2}{c}{0.075} & 0.538        \\ 
BiomedCLIP                   & 0.900 & \multicolumn{2}{c}{0.150} & 0.525 & 0.950 & \multicolumn{2}{c}{0.375} & 0.662        \\
FLAIR-$\pi_{\textit{naive}}$                        & 1.000 & \multicolumn{2}{c}{1.000} & \textbf{1.000} & 0.800 & \multicolumn{2}{c}{0.050} & 0.425        \\
\cellcolor{Gray}FLAIR-$\pi_{\textit{EK}}$        & \cellcolor{Gray}0.900 & \multicolumn{2}{c}{\cellcolor{Gray}1.000} & \cellcolor{Gray}0.950 & \cellcolor{Gray}0.850 & \multicolumn{2}{c}{\cellcolor{Gray}0.500} & \cellcolor{Gray}\textbf{0.675}        \\ \hline
\multicolumn{8}{l}{\textit{\textit{Inference with Naive Prompts - $\pi_{\textit{naive}}$  (e.g. "cataract")}}}\\ \hdashline
CLIP                        & 0.000 & 0.600 & 0.100 & 0.233 & 1.000 & 0.750 & 0.050 & \textbf{0.600}        \\ 
BiomedCLIP                  & 0.800 & 0.900 & 0.100 & 0.600 & 0.950 & 0.500 & 0.100 & 0.517        \\ 
FLAIR-$\pi_{\textit{naive}}$                        & 1.000 & 0.000 & 0.700 & 0.567 & 0.450 & 0.750 & 0.000 & 0.400        \\ 
\cellcolor{Gray}FLAIR-$\pi_{\textit{EK}}$        & \cellcolor{Gray}0.900 & \cellcolor{Gray}0.500 & \cellcolor{Gray}0.600 & \cellcolor{Gray}\textbf{0.667} & \cellcolor{Gray}0.950 & \cellcolor{Gray}0.550 & \cellcolor{Gray}0.000 & \cellcolor{Gray}0.500         \\ \hline
\multicolumn{8}{l}{\textit{\textit{Inference with Expert Knowledge Prompts - $\pi_{\textit{EK}}$}}}\\
\multicolumn{8}{l}{\textit{\textit{(e.g. "opacity in the macular area")}}}\\ \hdashline
CLIP                          & 1.000 & 0.000 & 0.000 & 0.333 & 0.900 & 0.250 & 0.950 & \textbf{0.700}        \\ 
BiomedCLIP                  & 0.500 & 0.100 & 0.600 & 0.400 & 0.200 & 0.450 & 1.000 & 0.550        \\ 
FLAIR-$\pi_{\textit{naive}}$                        & 1.000 & 0.400 & 0.000 & 0.467 & 0.850 & 0.800 & 0.050 & 0.567        \\ 
\cellcolor{Gray}FLAIR-$\pi_{\textit{EK}}$        & \cellcolor{Gray}1.000 & \cellcolor{Gray}1.000 & \cellcolor{Gray}0.300 & \cellcolor{Gray}\textbf{0.767} & \cellcolor{Gray}1.000 & \cellcolor{Gray}0.750 & \cellcolor{Gray}0.000 & \cellcolor{Gray}0.583         \\ \hline

\end{tabular}

\end{table}

\subsection{Transferability}
\label{sec:supTranf}

\begin{figure}[h!]
    \begin{center}

           \subfloat{\includegraphics[width=0.88\linewidth]{images/legend.png}} 
        
          \subfloat{\includegraphics[height=0.35\linewidth]{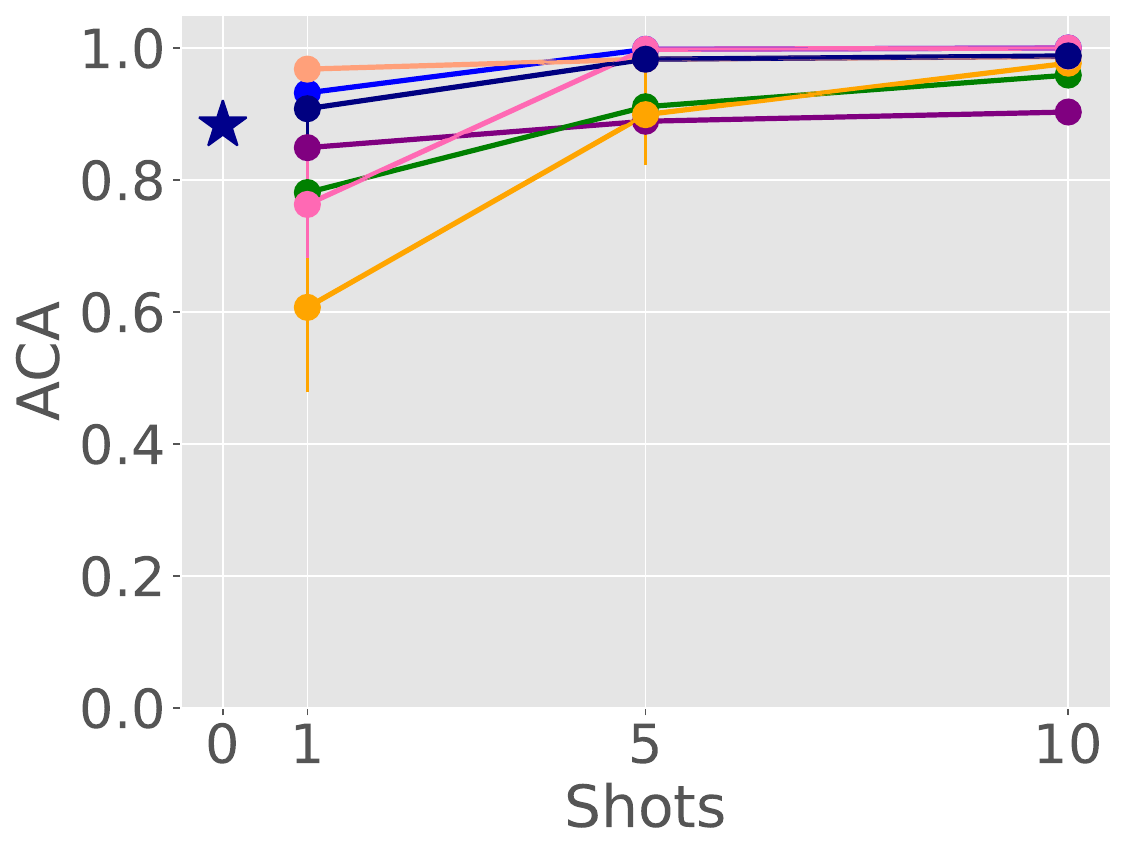}}
          \subfloat{\includegraphics[height=0.35\linewidth]{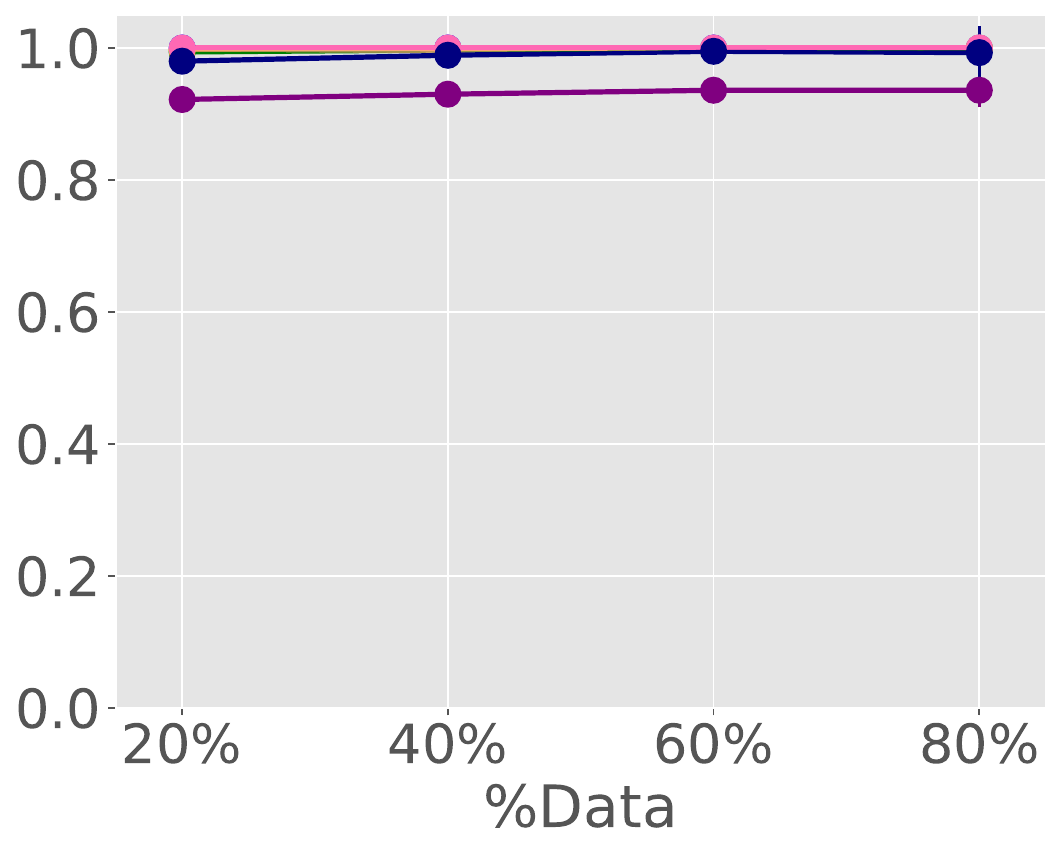}}

         \footnotesize{ACRIMA - \textit{Domain shift - Glaucoma}}
        
        \caption{\textbf{Transferability on ACRIMA dataset.} Results of transferring the feature representations of the pre-trained models to ACRIMA dataset for glaucoma detection in the low-data (\textit{left column}) and large-data (\textit{right column}) regimes. The results were obtained by adjusting a linear-probe classifier. The metric presented is the average accuracy, averaged across 5 cross-validation folds. ZS: zero-shot (\textit{i.e.} prompt-based classification).}
        \label{fig:transferability_acrima}
    \end{center}
\end{figure}

\paragraph{\textbf{Transferability for Glaucoma detection}} Figure \ref{fig:transferability_acrima} shows the transferability results for glaucoma detection on the ACRIMA dataset. In addition, we present in Figure \ref{fig:transferability_acrima} the generalization performance of Adapters using the REFUGE dataset. Concretely, the Adapters are adjusted on one of the datasets, and evaluated on both of them. This process is carried out crosswise, under the low-data regime.

\begin{figure}[h!]
    \begin{center}

         \subfloat{\includegraphics[width=0.6\linewidth]{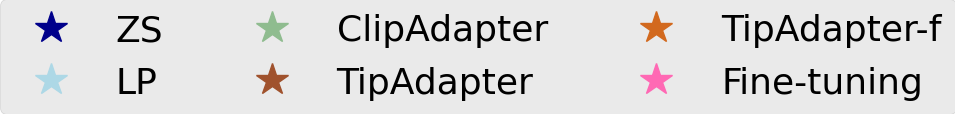}} 
          
          \subfloat{\includegraphics[height=0.35\linewidth]{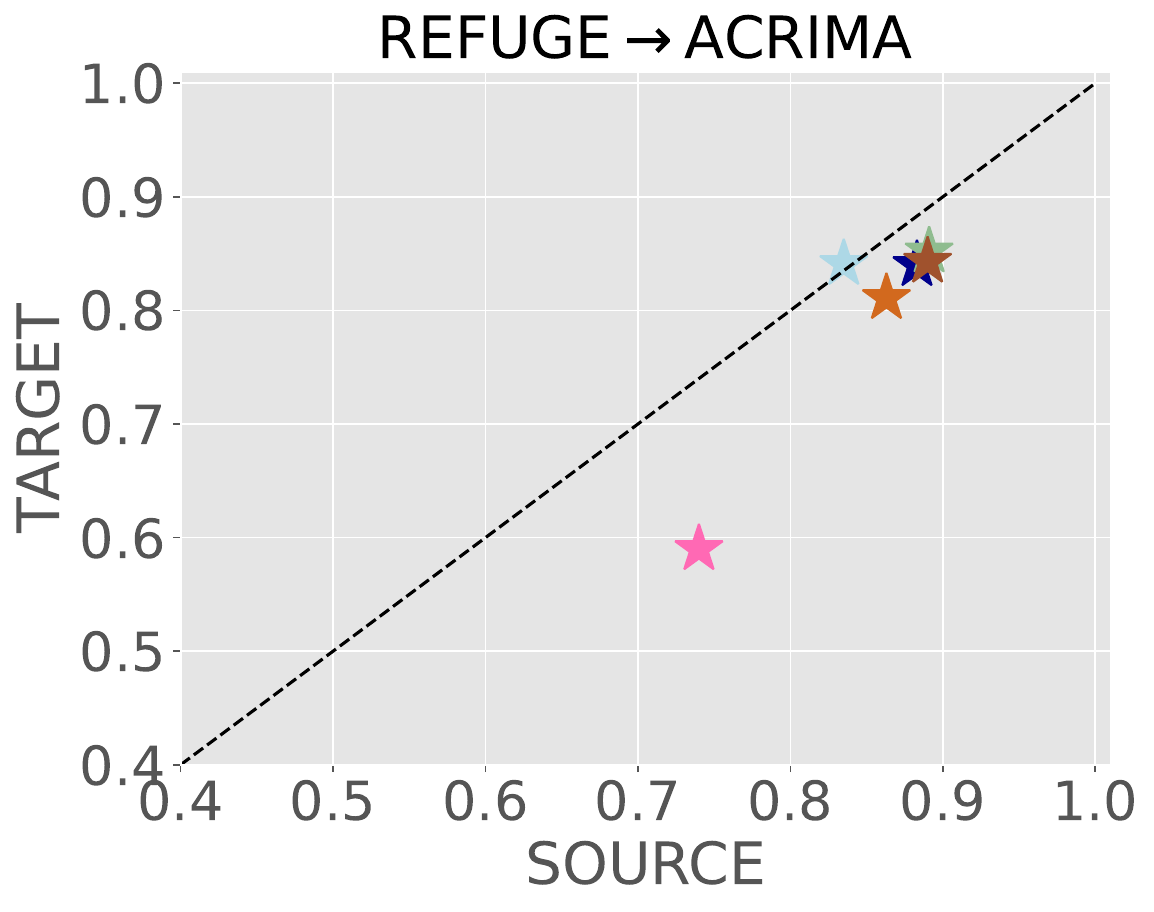}}
          \subfloat{\includegraphics[height=0.35\linewidth]{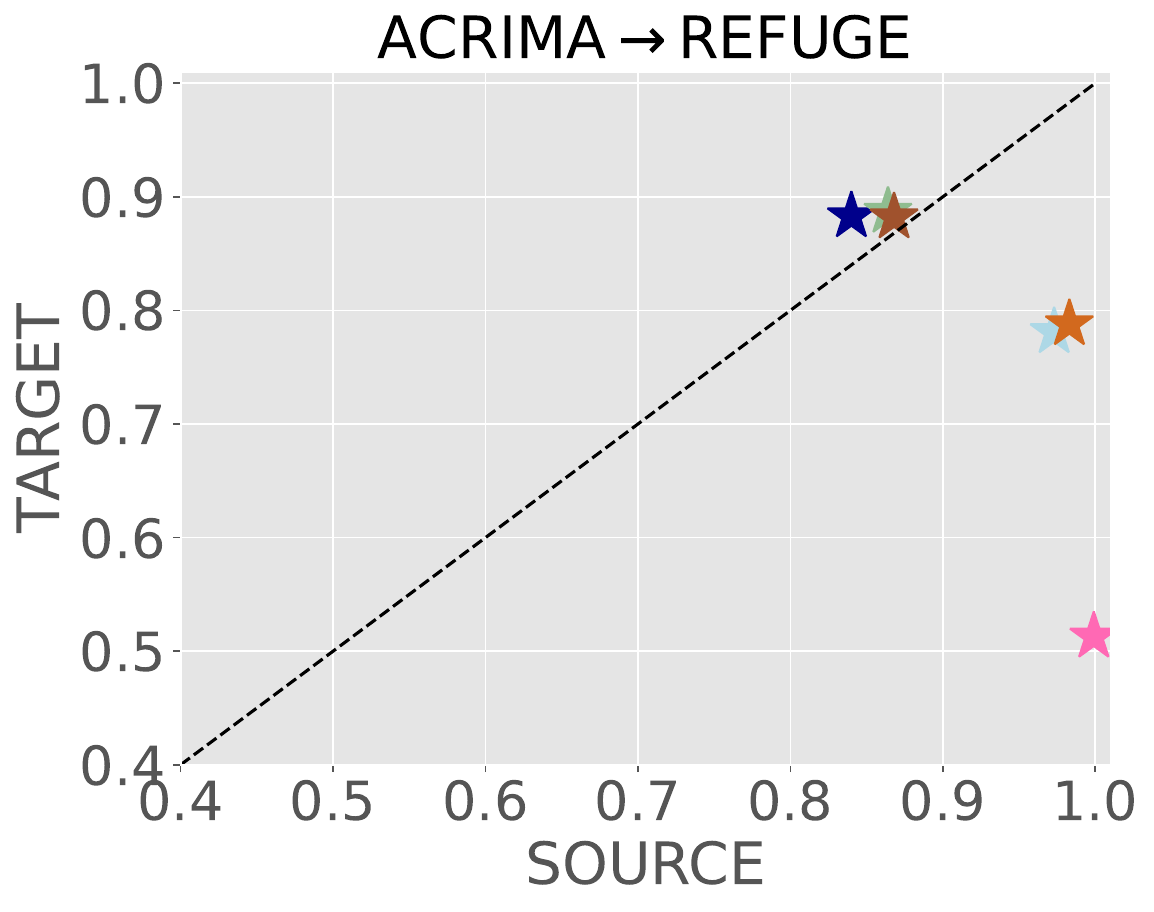}}

        \footnotesize{\textit{Domain shift - Glaucoma}}
        
        \caption{\textbf{Generalization of the Adapters for glaucoma detection.} Evaluation of the generalization capabilities of the few-shot Adapters under domain shifts. The Adapters are tuned on the source domain, and the performance is evaluated on both the source and target datasets. The metric presented is the average accuracy, averaged across 5 cross-validation folds. The results were obtained using $k=10$ shots.}
        \label{fig:adap_gen_sup}
    \end{center}
    \vspace{-5mm}
\end{figure}

Regarding transferability, the obtained results are in line with those observed in the other datasets, which we provided in the main paper. First, the foundation model, pre-trained using language supervision on a large number of tasks, shows better transferability than task-specific models, including the one trained for the glaucoma detection task. Secondly, it performs on par with the fine-tuned, dataset-specific model, while requiring less numbers of shots, with only the linear-probe classifier being tuned, in an efficient way. Still, in this dataset, in contrast to the other evaluation partitions, the fine-tuned counterpart reaches almost a perfect performance while requiring very few shots. Nevertheless, the resultant model shows poor transferability under domain shift when tested on the REFUGE dataset (see Figure \ref{fig:adap_gen_sup}). In contrast, vision-language Adapters over the foundation model tuned on the source domain are able to maintain the performance on the target set. Interestingly, zero-shot classification (\textit{i.e.}, no adaptation) shows the best transferrability between both subsets. It is worth mentioning that ACRIMA combines different centers to obtain glaucoma and non-glaucoma fundus samples, which might introduce a bias when training dataset-specific models.

\begin{figure}[h!]
    \begin{center}

         \subfloat{\includegraphics[width=0.6\linewidth]{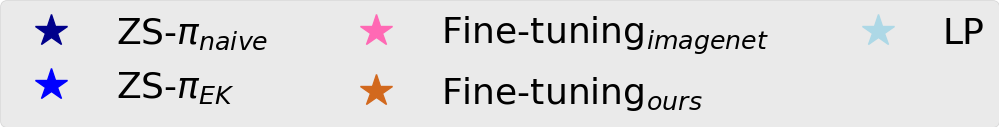}} 
          
          \subfloat{\includegraphics[height=0.35\linewidth]{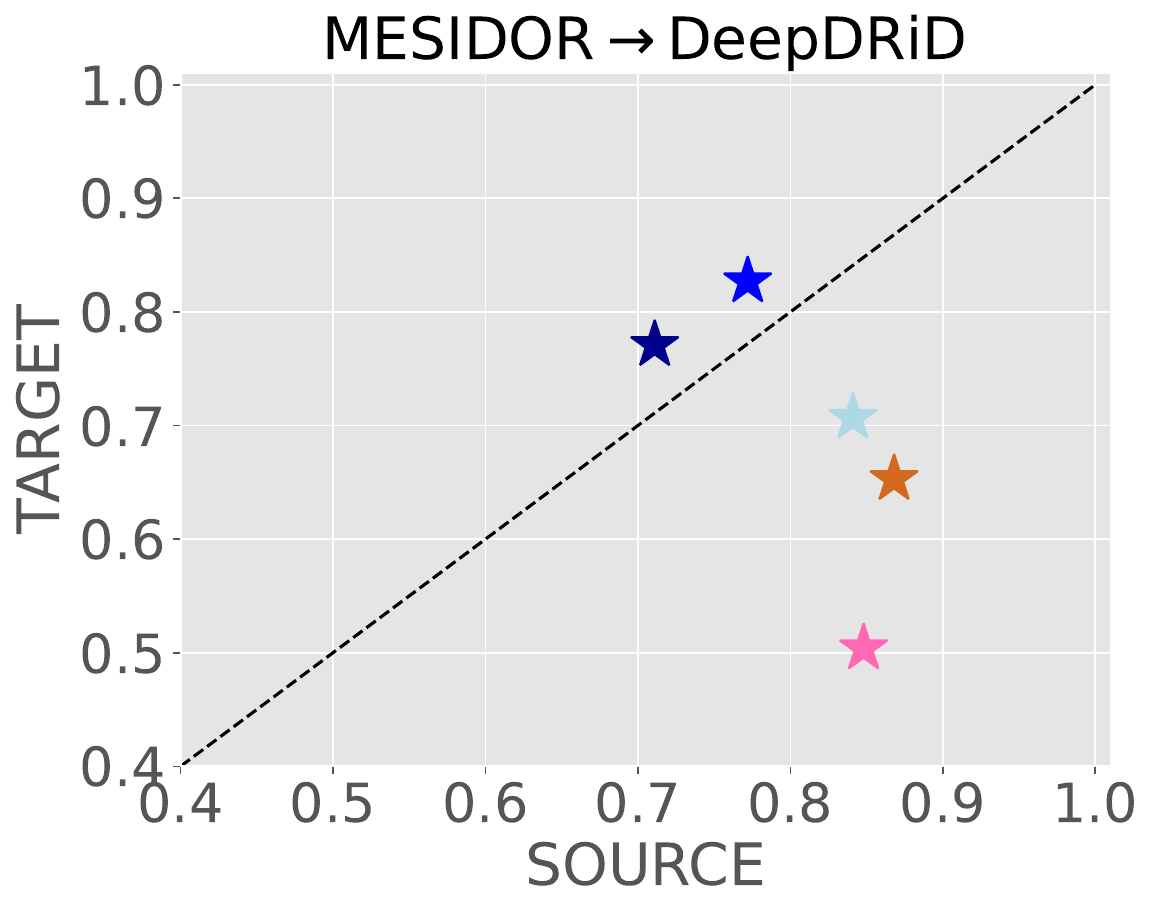}}
          \subfloat{\includegraphics[height=0.35\linewidth]{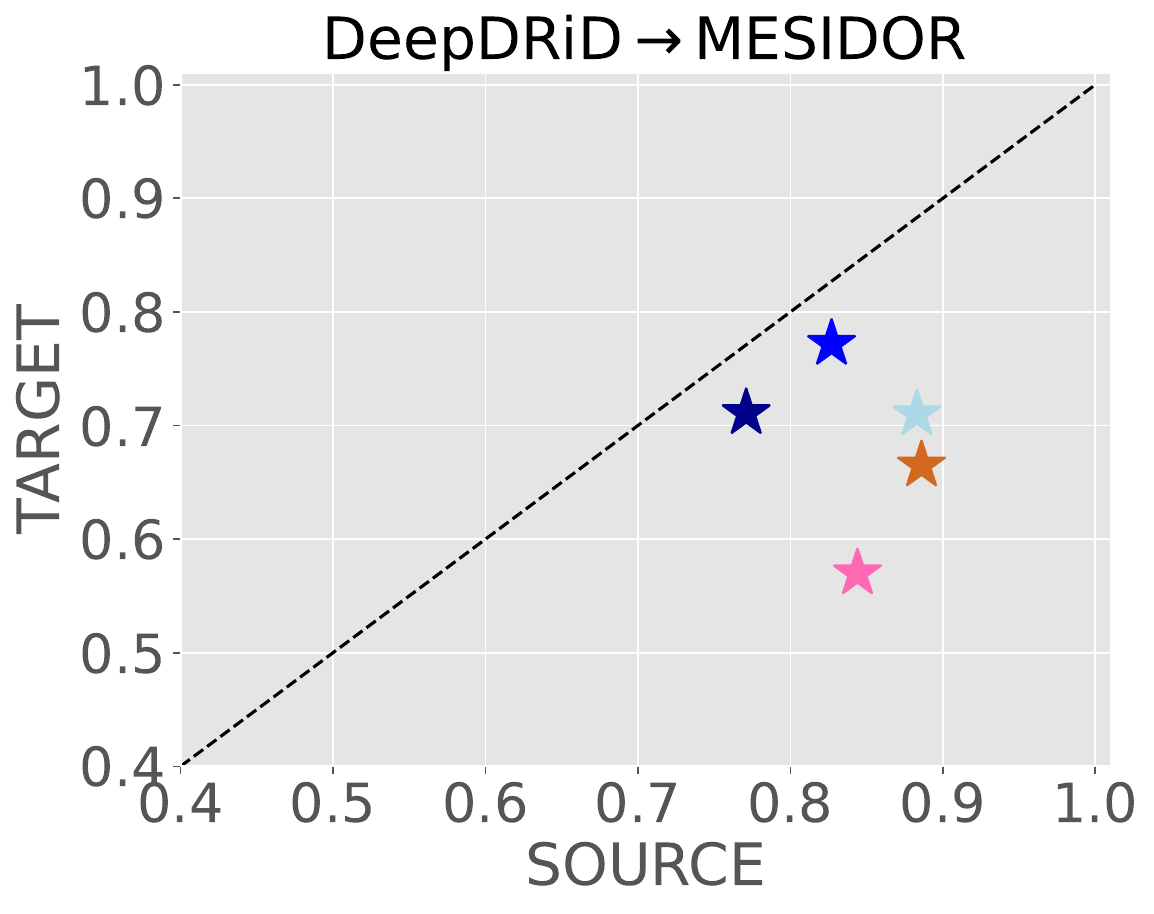}}

          \footnotesize{\textit{Domain shift - DR}}
        
        \caption{\textbf{Adaptation stage generalization for DR grading.} Evaluation of the generalization capabilities of Adapters and dataset-specific models. Different strategies are tuned on the source domain, and the performance is evaluated on both source and target datasets. The metric presented is the quadratic Cohen kappa ($\kappa$). Results were obtained on the large data regime, using the whole train subset from DeepDiD, and $80\%$ of MESIDOR samples for training.}
        \label{fig:dr_generalization}
    \end{center}
\end{figure}

\paragraph{\textbf{Transferability for DR grading}} In the following, we present experiments to study the generalization capabilities of the pre-trained foundation model for DR grading. Concretely, we use the DeepDRiD challenge \citep{Liu2022} dataset for this task. In this section, we evaluate the model performance via the quadratic Cohen kappa, the main figure of merit used for DR grading evaluation. Figure \ref{fig:dr_generalization} presents the generalization performance of different strategies tuned under the large-data regime on DeepDRiD and MESIDOR, evaluated on the source domain, and on the other dataset, used as a target. The evaluated methods are prompt-based classification using the foundation model via category names ($\pi_{\textit{naive}}$), using an ensemble of expert knowledge descriptions for each DR class ($\pi_{\textit{EK}}$), and a linear-probe Adapter. In addition, we train dataset-specific models (\textit{i.e.}, full fine-tuning) using ResNet-50 initialized on ImageNet (Fine-tuning$_{Imagenet}$), and on the foundation fundus model (Fine-tuning$_{ours}$).

\begin{figure}[h!]
    \begin{center}

         \subfloat{\includegraphics[width=0.6\linewidth]{images/legend_adapters.png}}  
          
          \subfloat{\includegraphics[height=0.35\linewidth]{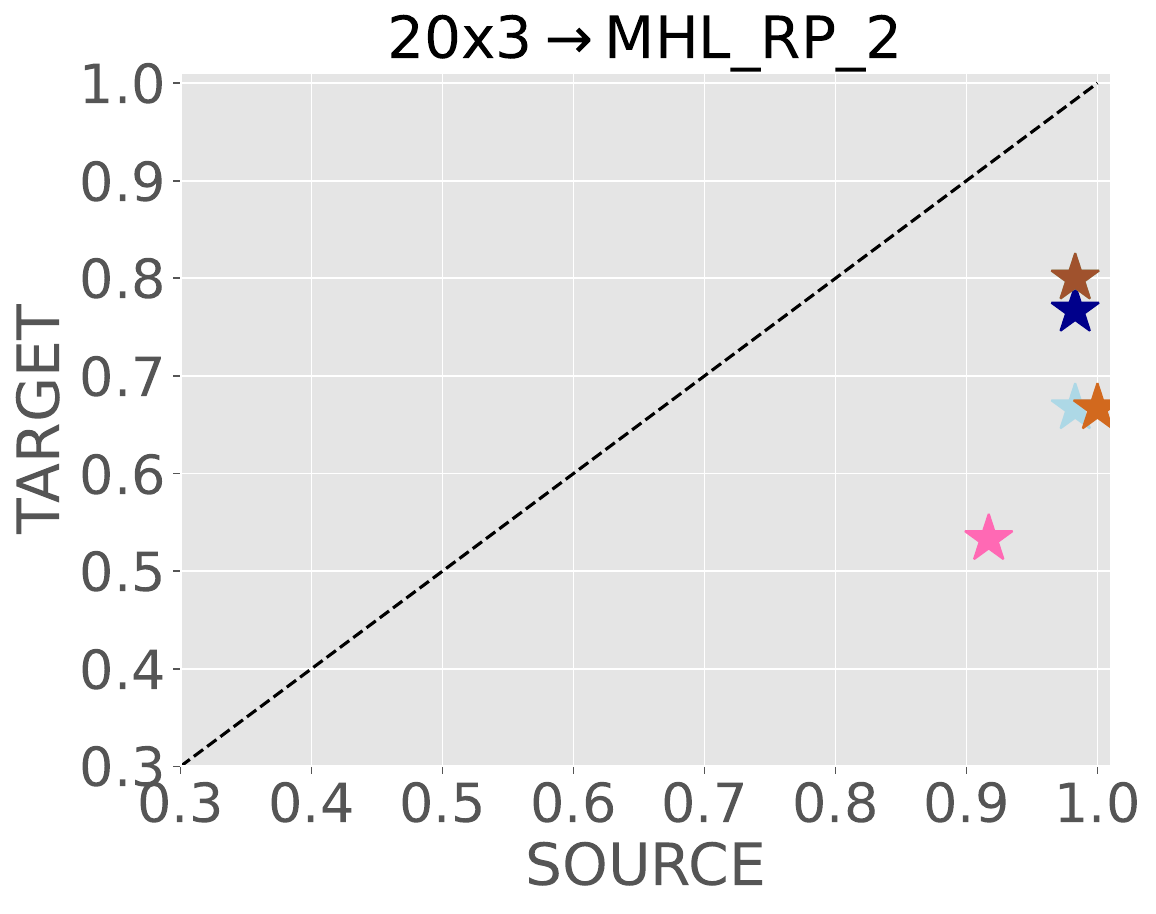}}
          \subfloat{\includegraphics[height=0.35\linewidth]{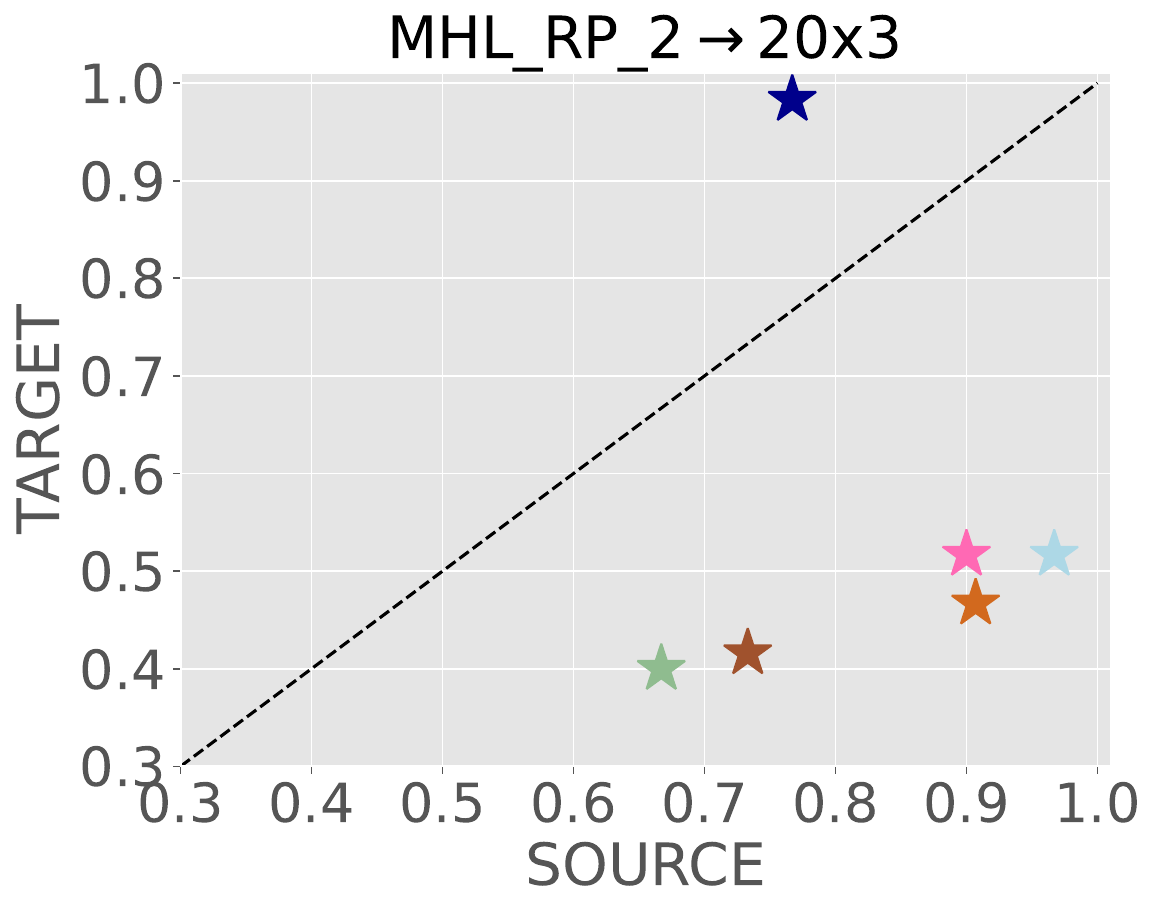}}

         \footnotesize{\textit{Unseen categories - N, RP, MHL}}

          \subfloat{\includegraphics[height=0.35\linewidth]{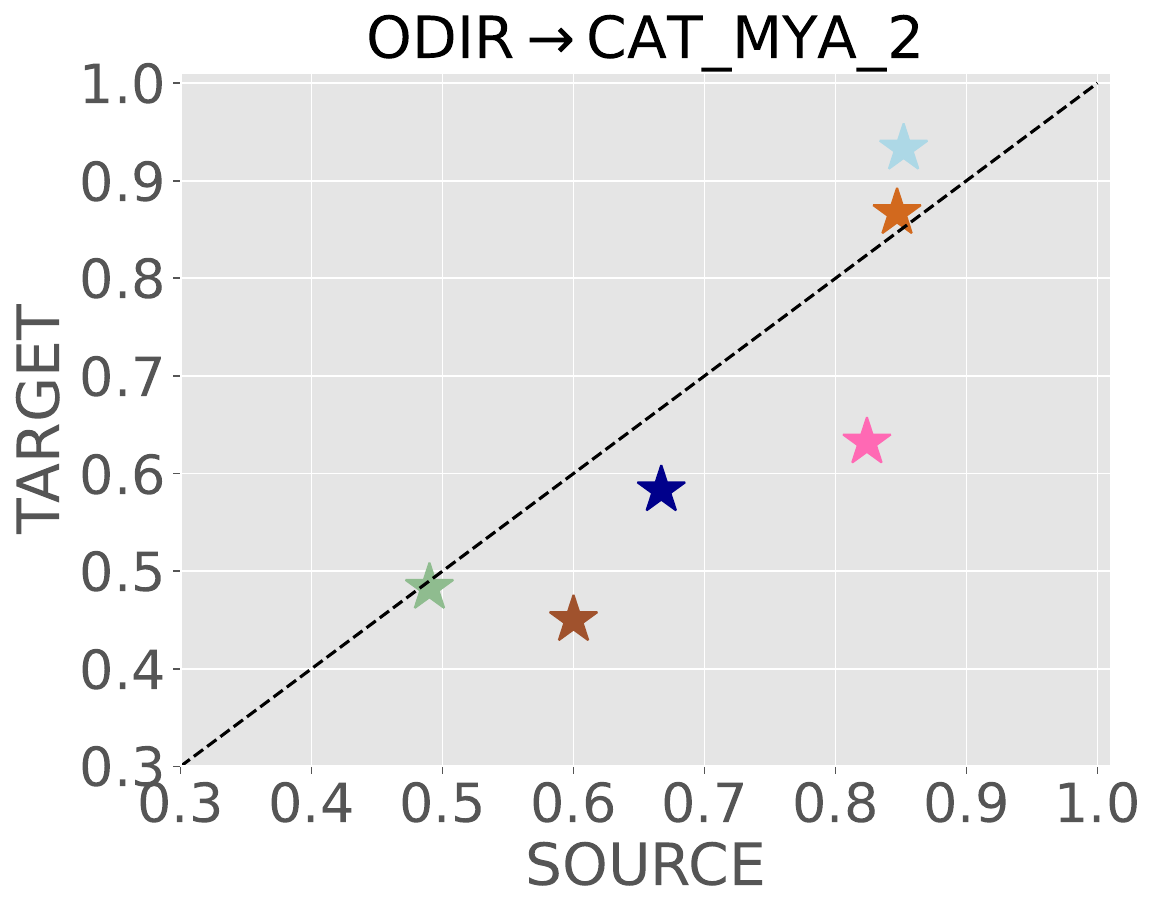}}
          \subfloat{\includegraphics[height=0.35\linewidth]{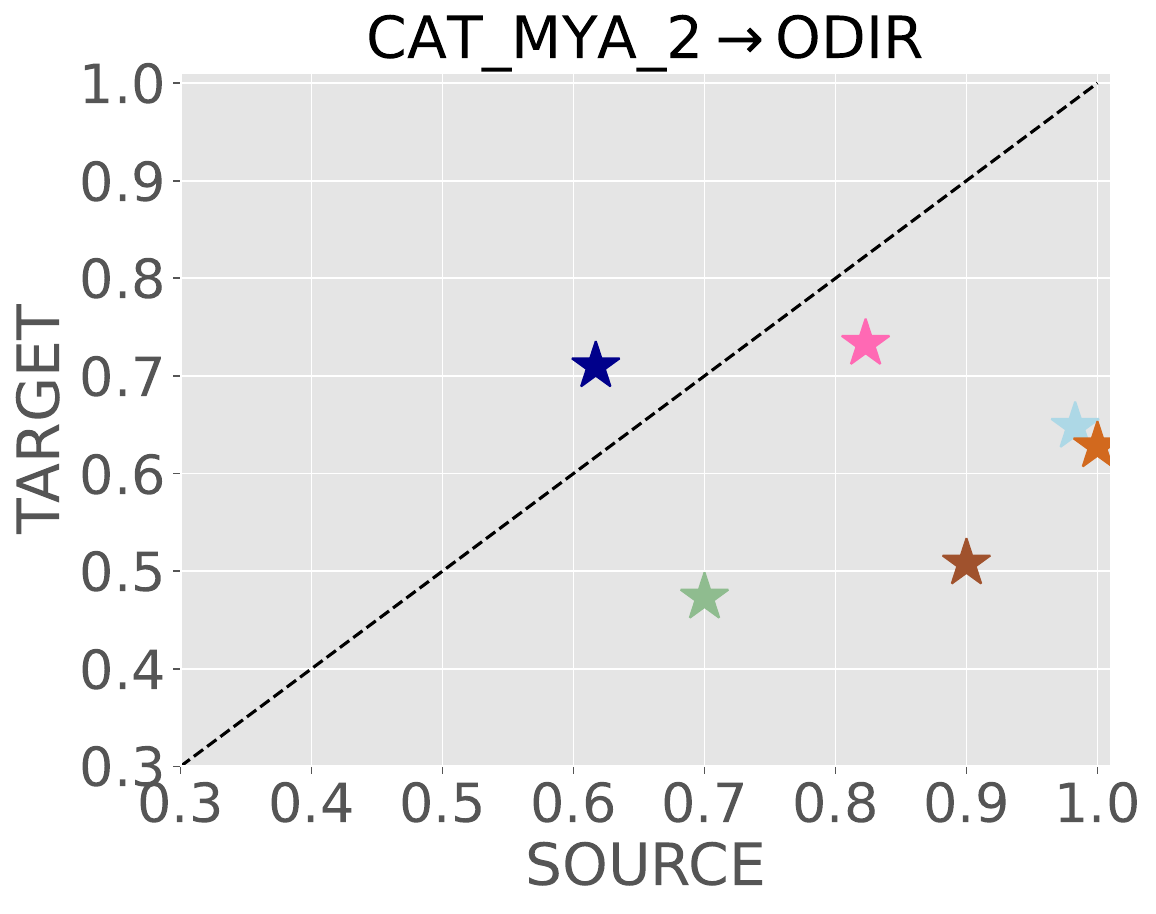}}

         \footnotesize{\textit{Unseen categories - N, CAT, MYA}}

        \caption{\textbf{Adapter generalization.} Evaluation of the generalization capabilities of few-shot Adapters for novel classes under domain shift. Adapters are tuned on the source domain, and the performance is evaluated on another dataset with the same categories. The metric presented is the average accuracy, averaged across 5 cross-validation folds. The results were obtained using $k=5$ shots.}
        \label{fig:adap_gen}
    \end{center}
\end{figure}

The obtained results using full fine-tuning resemble the outstanding performance obtained in the DeepDRiD challenge, in which leaderboard methods reached $[0.90, 0.93]$ quadratic kappa. It is worth mentioning that due to the heuristics for ranking optimization common to these competitions, these methods include additional strategies such as test-time adaptation, model ensemble, or selection of the most appropriate backbone. Thus, they may reach slightly better performance on the source domain, but no cross-domain performance is assessed. Nonetheless, common conclusions can be drawn: using a model with initialized weights on pre-trained fundus analysis tasks yields improved results. However, our results suggest that these source-domain specialized models fail remarkably when faced with a domain shift. In this context, the proposed foundation model presents interesting properties. First, by simply fitting a linear-probe classifier, the model achieves a performance on par with the full-backbone fine-tuning, while not being penalized as much in the scenario of a domain shift. Second, prompt-based classification with no adaptation reaches the best out-of-distribution performance, which improves using an ensemble of domain-knowledge descriptors for both domains.

\paragraph{\textbf{Generalization of the few-shot Adapters on unseen categories}} In the main experiments of this paper, we have empirically presented the limitations of the few-shot vision-language Adapters in the medical context, with respect to domain knowledge prompts (if $k \leq 5$) or linear probes of vision features (if $k \geq 10$); see Section \ref{sec:visionLanguage} for more details. Still, the linear-probe Adapter seems to generalize worse against sequential domain shifts after the tuning stage, compared to zero-shot classification using text prompts (see Section \ref{ablation_experiments}). This motivates additional experiments to validate the effectiveness of vision-language Adapters in this setting. In particular, we evaluate the performance of the Adapters for unseen categories in the foundation model, tuned on a source domain, and evaluated on both, source and target domains. Experiments are carried out in the low-data regime, to reproduce a realistic scenario for vision-language Adapters, where only a handful of labeled samples are available. Results are depicted in Figure \ref{fig:adap_gen}. As already presented, the Adapters struggle to reach the best performance in the source domain. Nevertheless, these Adapters are able to maintain a more robust performance on the target domains (see Figure \ref{fig:adap_gen}, \textit{bottom-right} and \textit{top-right}), or even slight improvements for both, source and target (see Figure \ref{fig:adap_gen}, TipAdapter \textit{top-left}, TipAdapter-f \textit{bottom-left}).

\begin{table*}[h!]
\caption{\textbf{Expert Knowledge descriptions.}}
\label{dk_description}
\scriptsize
\centering
\begin{tabular}{ll}
\hline
\textbf{Category} & \textbf{Domain Knowledge descriptor} \\ \hline
no diabetic retinopathy & "no relevant haemorrhages, microaneurysms or exudates" / "no microaneurysms" / "no referable lesions" \\                \cellcolor{Gray}mild diabetic retinopathy & \cellcolor{Gray}"few microaneurysms" / "few hard exudates" / "few retinal haemorrhages" \\
moderate diabetic retinopathy & "retinal haemorrhages in few quadrants" / "many haemorrhages" / "cotton wool spots" \\ 
\cellcolor{Gray}severe diabetic retinopathy & \cellcolor{Gray}"severe haemorrhages in all four quadrants" / "venous beading" / "intraretinal microvascular abnormalities" \\ 
proliferative diabetic retinopathy & "diabetic retinopathy with neovascularization at the disk" / "neovascularization" \\ 
\cellcolor{Gray}diabetic macular edema & \cellcolor{Gray}"macular edema" / "presence of exudates" / "leakage of fluid within the central macula from microaneurysms" / "presence of \\ \cellcolor{Gray}& \cellcolor{Gray}exudates within the radius of one disc diameter from the macula center" \\ 
no referable diabetic macular edema & "no apparent exudates" \\ 
\cellcolor{Gray}hard exudates & \cellcolor{Gray}"small white or yellowish deposits with sharp margins" / "bright lesion" \\ 
soft exudates & "pale yellow or white areas with ill-defined edges" / "cotton-wool spot" / "small, whitish or grey, cloud-like, linear or serpentine, \\ & slightly elevated lesions with fimbriated edges" \\ 
\cellcolor{Gray}microaneurysms & \cellcolor{Gray}"small red dots" \\ 
haemorrhages & "dense, dark red, sharply outlined lesion" \\ 
\cellcolor{Gray}non clinically significant diabetic macular edema & \cellcolor{Gray}"presence of exudates outside the radius of one disc diameter from the macula center" / "presence of exudates" \\ 
age-related macular degeneration & "many small drusen" / "few medium-sized drusen" / "large drusen" \\ 
\cellcolor{Gray}media haze & \cellcolor{Gray}"vitreous haze" / "pathological opacity" / "the obscuration of fundus details by vitreous cells and protein exudation" \\ 
drusens & "yellow deposits under the retina" / "numerous uniform round yellow-white lesions" \\ 
\cellcolor{Gray}pathologic myopia & \cellcolor{Gray}"tilted disc, peripapillary atrophy, and macular atrophy. There are chorioretinal scars in the inferonasal periphery" / "maculopahy" \\ 
branch retinal vein occlusion & "occlusion of one of the four major branch retinal veins" \\ 
\cellcolor{Gray}tessellation & \cellcolor{Gray}"large choroidal vessels at the posterior fundus" \\ 
epiretinal membrane & "greyish semi-translucent avascular membrane" \\ 
\cellcolor{Gray}laser scar & \cellcolor{Gray}"round or oval, yellowish-white with variable black pigment centrally" / "50 to 200 micron diameter lesions" \\ 
central serous retinopathy & "subretinal fluid involving the fovea" / "leakage" \\ 
\cellcolor{Gray}asteroid hyalosis & \cellcolor{Gray}"multiple sparking, yellow-white, and refractile opacities in the vitreous cavity" / "vitreous opacities" \\ 
optic disc pallor & "pale yellow discoloration that can be segmental or generalized on optic disc" \\  
\cellcolor{Gray}shunt & \cellcolor{Gray}"collateral vessels connecting the choroidal and the retinal vasculature" / "collateral vessels of large caliber and lack of leakage" \\ 
exudates & "small white or yellowish-white deposits with sharp margins" / "bright lesion" \\ 
\cellcolor{Gray}macular hole & \cellcolor{Gray}"a lesion in the macula" / "small gap that opens at the centre of the retina" \\ 
retinitis pigmentosa & "bone spicule-shaped pigment deposits are present in the mid periphery" / "retinal atrophy" "the macula is preserved" / \\ & "peripheral ring of depigmentation" / "arteriolar attenuation and atrophy of the retinal pigmented epithelium" \\
\cellcolor{Gray}cotton wool spots & \cellcolor{Gray}"soft exudates" \\ 
glaucoma & "optic nerve abnormalities" / "abnormal size of the optic cup" / "anomalous size in the optic disc" \\ 
\cellcolor{Gray}severe hypertensive retinopathy & \cellcolor{Gray}"flame-shaped hemorrhages at the disc margin, blurred disc margins" / "congested retinal veins, papilledema, and secondary \\ \cellcolor{Gray} & \cellcolor{Gray} macular exudates" / "arterio-venous crossing changes, macular star and cotton wool spots" \\ 
no proliferative diabetic retinopathy & "diabetic retinopathy with no neovascularization" / "no neovascularization" \\ 
\cellcolor{Gray}hypertensive retinopathy & \cellcolor{Gray}"possible signs of hemorrhage with blot, dot, or flame-shaped" / "possible presence of microaneurysm, cotton-wool spot, or hard \\ \cellcolor{Gray} & \cellcolor{Gray} exudate" / "arteriolar narrowing" / "vascular wall changes" / "optic disk edema" \\ 
intraretinal microvascular abnormalities & "shunt vessels and appear as abnormal branching or dilation of existing blood vessels (capillaries) within the retina" / "deeper \\ & in the retina than neovascularization, has blurrier edges, is more of a burgundy than a red, does not appear on the optic disc" / \\ & "vascular loops confined within the retina" \\ 
\cellcolor{Gray}red small dots & \cellcolor{Gray}"microaneurysms" \\ 
a disease & "no healthy" / "lesions" \\ 
\cellcolor{Gray}normal & \cellcolor{Gray}"healthy" / "no findings" / "no lesion signs" \\ 

myopic maculopathy grade cero  & "healthy macula" \\ 
\cellcolor{Gray}myopic maculopathy grade one   & \cellcolor{Gray}"tessellated fundus" \\ 
myopic maculopathy grade two   & "diffuse chorioretinal atrophy" \\ 
\cellcolor{Gray}myopic maculopathy grade three & \cellcolor{Gray}"patchy chorioretinal atrophy" \\ 
myopic maculopathy grade four  & "macular atrophy" \\

\hline

\end{tabular}
\end{table*}

\end{document}